\documentclass[a4paper,11pt]{article}
\usepackage[top=2.5cm, bottom=2.5cm, left=2.4cm, right=2.4cm]{geometry}

\usepackage[colorlinks = true,
            linkcolor = magenta,
            urlcolor  = blue,
            citecolor = green,
            anchorcolor = blue]{hyperref}

\usepackage{amsmath,amssymb,amsthm}
\usepackage{authblk}  
\usepackage[british]{babel}
\usepackage{bbm}  
\usepackage[sorting=none,doi=false,isbn=false,giveninits=true,style=nature]{biblatex}           
\usepackage{caption}
\usepackage{color}
\usepackage{enumitem}
\usepackage[T1]{fontenc}
\usepackage[pdftex]{graphicx}
\PassOptionsToPackage{hyphens}{url}\usepackage{hyperref}
\usepackage[utf8]{inputenc}
\usepackage{csquotes}  
\usepackage[mono=false]{libertine}
\usepackage{nicefrac}
\usepackage{mathtools}
\usepackage{microtype}      
\usepackage{subcaption}
\usepackage{xcolor}
\usepackage[linguistics]{forest}
\usepackage{multirow}
\usepackage{array, booktabs, makecell}
\newcolumntype{L}{>{$}l<{$}} 
\usepackage{multirow}
\usepackage{tabularx}
\usepackage{wrapfig}

\usepackage[noabbrev,capitalize]{cleveref}
\makeatletter
\newcommand{\printfnsymbol}[1]{%
  \textsuperscript{\@fnsymbol{#1}}%
}

\newcommand{\xhdr}[1]{\textbf{#1}\;}
\newcommand{\odeformer}{ODEFormer}
\newcommand{\odebench}{ODEBench}

\bibliography{refs}

\DeclareUnicodeCharacter{2212}{-}

\newcommand{\R}{\mathbb{R}}

\DeclareMathOperator*{\argmin}{arg\,min}

\usepackage{xcolor}
\definecolor{C0}{HTML}{1f77b4}
\definecolor{C1}{HTML}{ff7f0e}
\definecolor{C2}{HTML}{2ca02c}
\definecolor{C3}{HTML}{d62728}
\definecolor{C4}{HTML}{9467bd}
\definecolor{C5}{HTML}{8c564b}
\definecolor{C6}{HTML}{e377c2}
\definecolor{C7}{HTML}{7f7f7f}
\definecolor{C8}{HTML}{bcbd22}
\definecolor{C9}{HTML}{17becf}

\usepackage[normalem]{ulem}

\newcommand{\din}{D}

\newcommand{\demb}{d_\text{emb}}

\newcommand{\citet}[1]{\textcite{#1}}
\newcommand{\citep}[1]{\parencite{#1}}

\title{\odeformer{}:\\ Symbolic Regression of Dynamical Systems with Transformers}

\author[1]{St\'ephane d'Ascoli$^*$}
\author[2,3]{S{\"o}ren Becker$^*$}
\author[1]{Alexander Mathis}
\author[1,4]{Philippe Schwaller}
\author[2,3]{Niki Kilbertus}
\affil[1]{EPFL}
\affil[2]{Helmholtz AI, Helmholtz Center Munich}
\affil[3]{TU M{\"u}nchen}
\affil[4]{NCCR Catalysis}
\date{}

\makeatother

\begin{document}

\maketitle
\def\thefootnote{*}\footnotetext{Equal contribution} 
\def\thefootnote{\arabic{footnote}}
\vspace{-1cm}
\begin{abstract}
\noindent We introduce \odeformer{}, the first transformer able to infer multidimensional ordinary differential equation (ODE) systems in symbolic form from the observation of a single solution trajectory. We perform extensive evaluations on two datasets: (i) the existing `Strogatz' dataset featuring two-dimensional systems; (ii) \odebench{}, a collection of one- to four-dimensional systems that we carefully curated from the literature to provide a more holistic benchmark. \odeformer{} consistently outperforms existing methods while displaying substantially improved robustness to noisy and irregularly sampled observations, as well as faster inference. We release our code, model and benchmark dataset publicly.

\end{abstract}

\section{Introduction}\label{sec:intro}

Recent triumphs of machine learning (ML) spark growing enthusiasm for accelerating scientific discovery~\citep{davies2021advancing, jumper2021highly,degrave2022magnetic}.
In particular, inferring dynamical laws governing observational data is an extremely challenging task that is anticipated to benefit substantially from modern ML methods.
Modeling dynamical systems for forecasting, control, and system identification has been studied by various communities within ML.
Successful modern approaches are primarily based on advances in deep learning, such as neural ordinary differential equation (NODE) (see \citet{chen2018neural} and many extensions thereof).
However, these models typically lack interpretability due to their black-box nature, which has inspired extensive research on explainable ML of overparameterized models \citep{tjoa2020survey,dwivedi2023explainable}.
An alternative approach is to directly infer human-readable representations of dynamical laws.

\begin{figure*}
    \centering
    \includegraphics[width=\columnwidth]{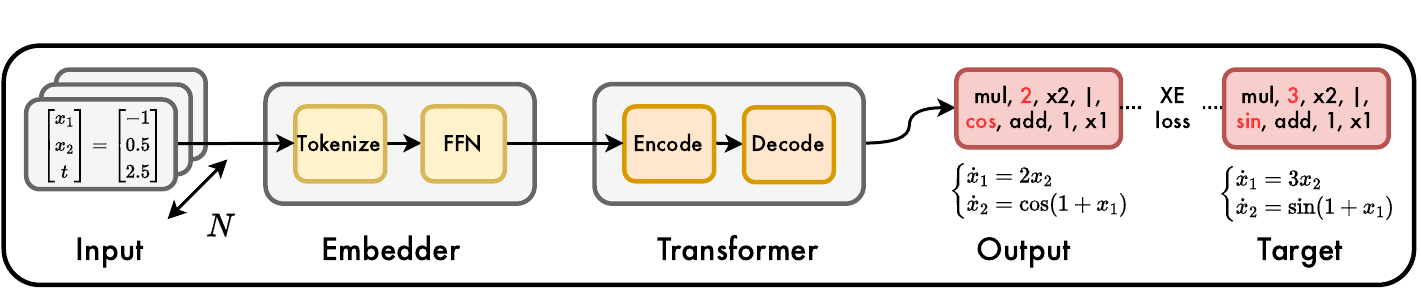}
    \caption{\textbf{Sketch of our method to train \odeformer{}}. We generate random ODE systems, integrate a solution trajectory on a grid of $N$ points $x\in \R^{\din}$, and train \odeformer{} to directly output the ODE system in symbolic form, supervising the predicted expression via cross-entropy loss.}\label{fig:sketch}
\end{figure*}

This is the main goal of symbolic regression (SR) techniques, which make predictions in the form of explicit symbolic mathematical expressions directly from observations.
Recent advances in SR make it a promising alternative to infer natural laws from data and have catalyzed initial successes in accelerating scientific discoveries~\citep{Archiga2021AcceleratingUO,Udrescu2021SymbolicPD,butter2021back}.
So far, SR has most commonly been used to infer a function $g(x)$ from paired observations $(x, g(x))$ -- we call this \emph{functional SR}.
However, in many fields of science, understanding a system involves deciphering its dynamics, i.e., inferring a function $f(x)$ governing its evolution via an ODE $\dot{x} = f(x)$ -- we call this setting \emph{dynamical SR}.
The task is then to uncover $f$ from an observed solution trajectory $(t, x(t))$, where observations of $x(t)$ may be noisy and times $t$ may be irregularly sampled.
\begin{figure}[t]
    \centering
    \vspace{-6mm}
    \begin{subfigure}[b]{.24\linewidth}
    \includegraphics[width=\linewidth]{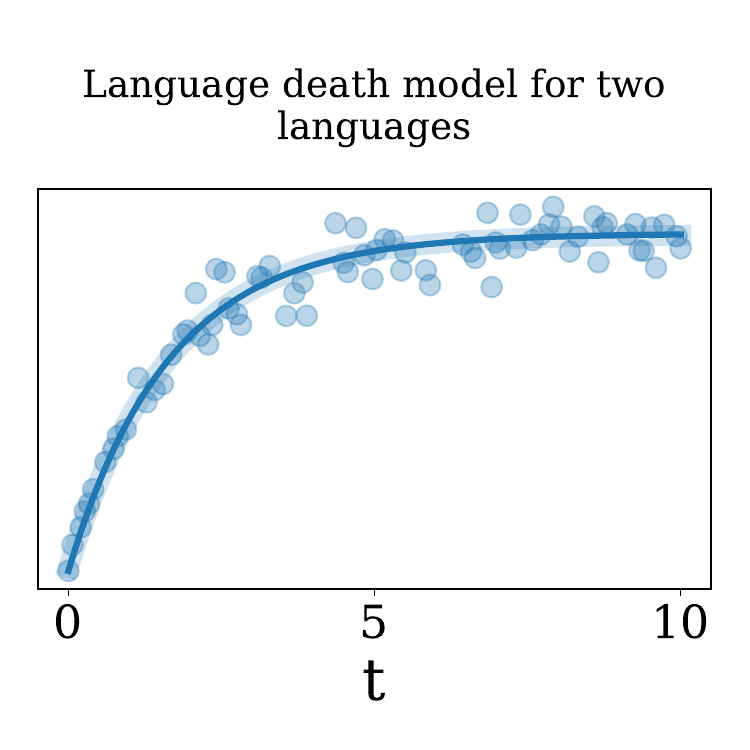}
    \end{subfigure}
    \begin{subfigure}[b]{.24\linewidth}
    \includegraphics[width=\linewidth]{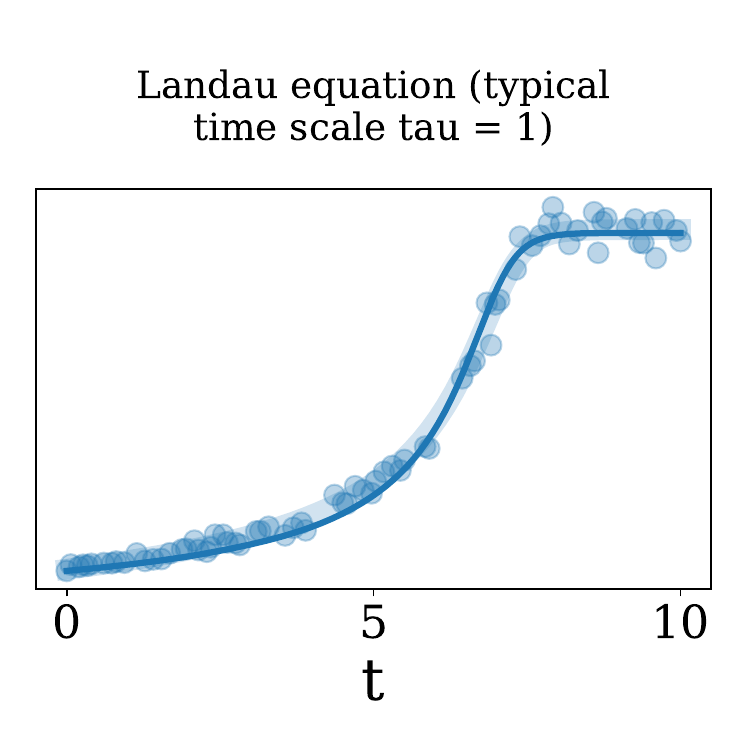}
    \end{subfigure}
    \begin{subfigure}[b]{.24\linewidth}
    \includegraphics[width=\linewidth]{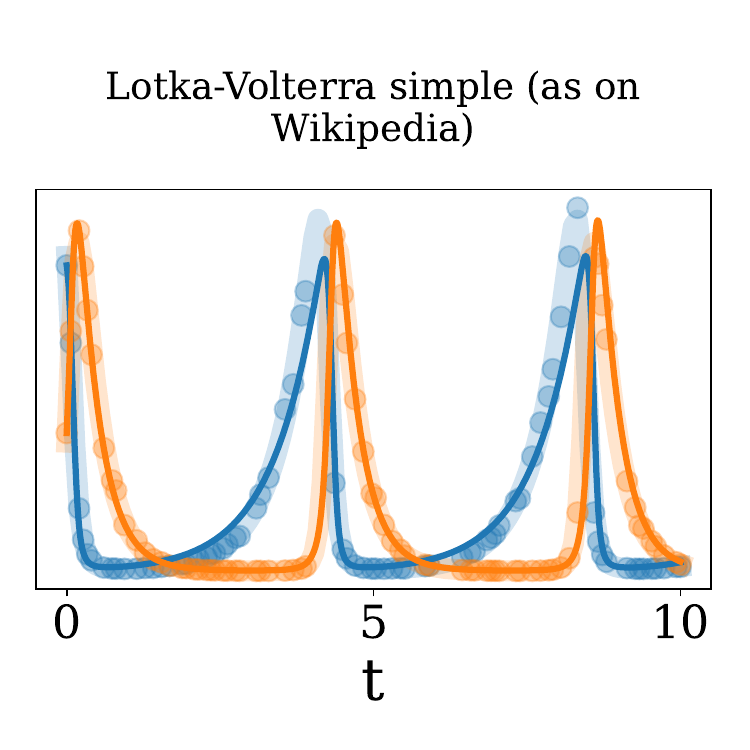}
    \end{subfigure}
    \begin{subfigure}[b]{.24\linewidth}
    \includegraphics[width=\linewidth]{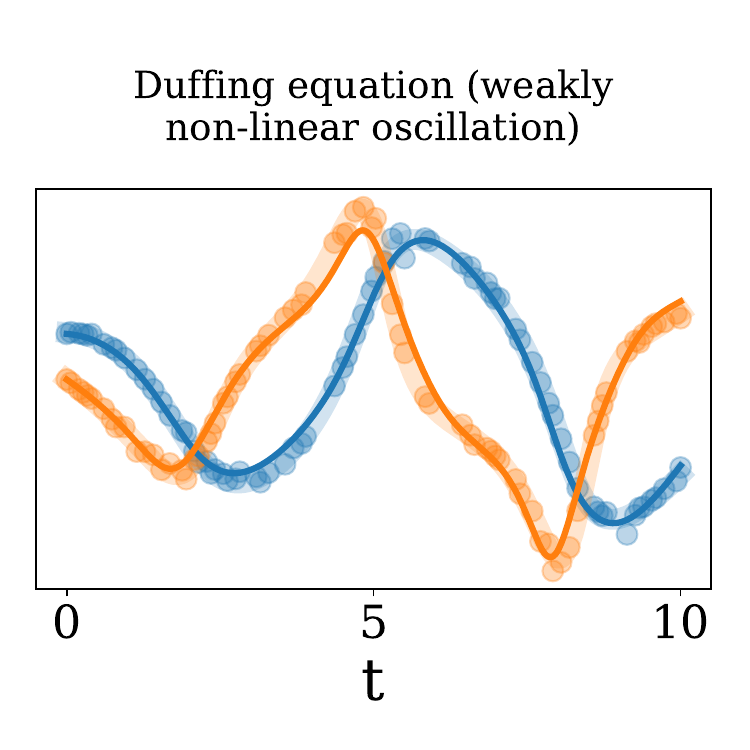}
    \end{subfigure}\\[-5mm]
    \begin{subfigure}[b]{.24\linewidth}
    \includegraphics[width=\linewidth]{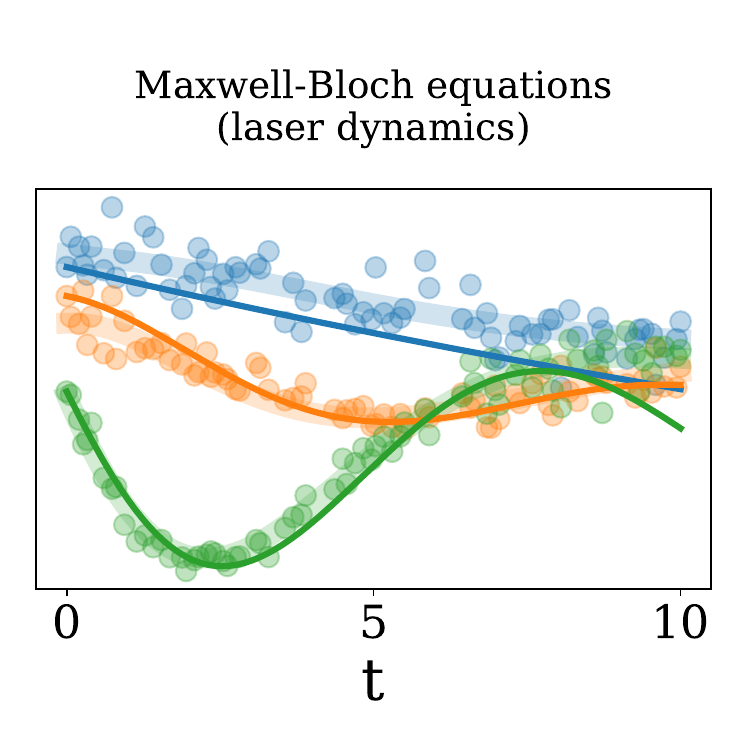}
    \end{subfigure}
    \begin{subfigure}[b]{.24\linewidth}
    \includegraphics[width=\linewidth]{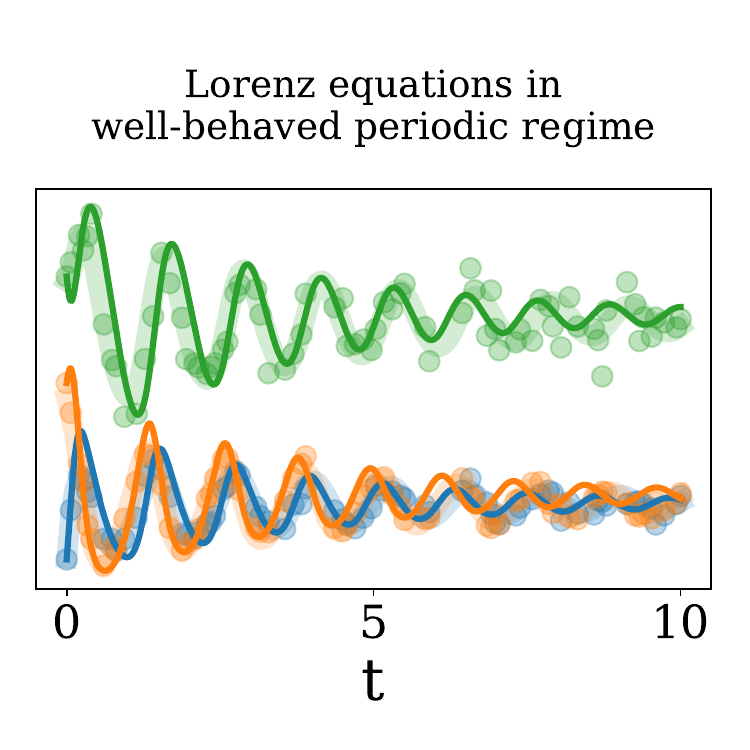}
    \end{subfigure}
    \begin{subfigure}[b]{.24\linewidth}
    \includegraphics[width=\linewidth]{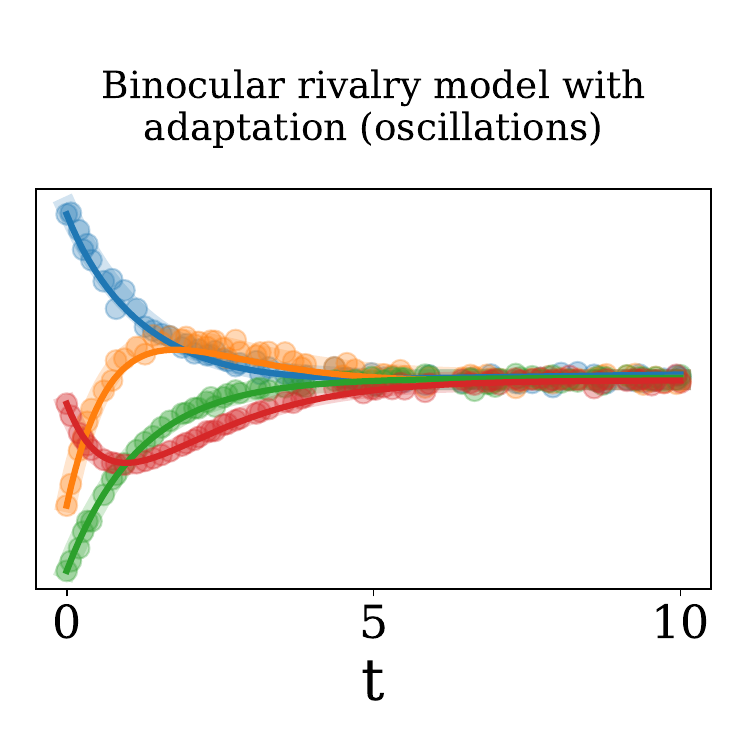}
    \end{subfigure}
    \begin{subfigure}[b]{.24\linewidth}
    \includegraphics[width=\linewidth]{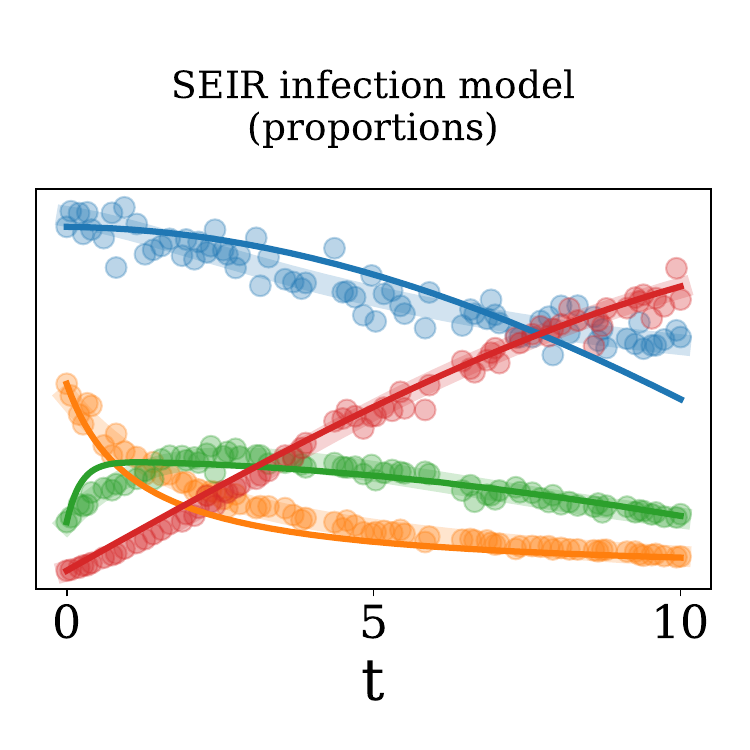}
    \end{subfigure}
    \vspace{-3mm}
    \caption{\textbf{\odeformer{} reconstructs noisy and irregularly sampled trajectories well.} Ground truth trajectories (thick lines) are corrupted (5\% noise) and unevenly sampled (50\% of the equally spaced points are dropped uniformly at random), yielding the observed trajectories (dots) from which \odeformer{} infers a symbolic ODE. 
    Integrating this ODE from the original initial condition (thin lines) shows that it approximates the observations well.
    We display 1, 2, 3 and 4 dimensional ODEs from our dataset \odebench{} -- see \cref{app:odebench} for details and results on the remaining ODEs. 
    }\label{fig:examples}
    \vspace{-4mm}
\end{figure}
\\
\newline
\noindent\xhdr{Contributions.}
In this work, we introduce \odeformer{}, the first Transformer trained to infer dynamical laws in the form of multidimensional ODEs from observational (noisy, irregularly sampled) data.
It relies on large-scale training of a sequence-to-sequence transformer on diverse synthetic data, leading to efficient and scalable inference for unseen trajectories. 
Faced with the lack of benchmarks for dynamical SR (the only existing one, called ``Strogatz dataset'' \citep{la2021contemporary}, contains only seven two-dimensional systems, and is not integrated with sufficient precision \citep{omejc2023probabilistic}), we also introduce \odebench{}, a more extensive dataset of 63 ODEs curated from the literature, which model real-world phenomena in dimensions one to four (see \cref{app:odebench} for details).
On both benchmarks, \odeformer{} achieves higher accuracy than existing methods for irregular, noisy observations while being faster than most competitors.
Our code, models and benchmark dataset are available publicly at
\url{https://github.com/sdascoli/odeformer}.
\\
\newline
\noindent\xhdr{Problem setting and overview.}
We assume observations $\{(t_i, x_i)\}_{i \in [N]}$,\footnote{We use the notation $[N] := \{1, \ldots, N\}$ and $\dot{x}$ for the temporal derivative.} of a solution $x(t)$ of
\begin{equation*}
    \dot{x} = f(x), \quad \text{for some } f: \R^{\din} \to \R^{\din}\:,
\end{equation*}
where $x_i$ can be noisy observations of $x(t_i)$ with irregularly sampled $t_i$.
The task is to infer $f$ in symbolic form from the data $\{(t_i, x_i)\}_{i \in [N]}$.
As illustrated in \cref{fig:sketch}, \odeformer{} is based on large-scale pre-training on ``labelled'' examples that consist of randomly generated symbolic mathematical expressions as the prediction target $f$ and a discrete solution trajectory $\{(t_i, x_i)\}_{i \in [N]}$ as input, which we obtain by integrating $\dot{x} = f(x)$ for a random initial condition.

\section{Related Work}\label{sec:related_work}
\xhdr{Modeling dynamical systems.}
We briefly mention some relevant cornerstones from the long history of modeling dynamical systems from data, each of which has inspired a large body of follow-up work.
Neural ODEs (NODE)~\citep{chen2018neural} parameterize the ODE $f$ by a neural network and train it by backpropagating through ODE solvers (either directly or via adjoint sensitivity methods).
NODEs require no domain knowledge but only represent the dynamics as a generally overparameterized black-box model void of interpretability.
Assuming prior knowledge about the observed data, physics-informed neural networks~\citep{raissi2019physics,karniadakis2021physics} aim to model dynamical systems using neural networks regularized to satisfy a set of physical constraints.
This approach was recently extended to uncover unknown terms in differential equations~\citep{podina2023universal}. 
In this work, we aim to infer interpretable dynamical equations when no domain knowledge is available.
\\
\newline
\noindent\xhdr{Approaches to symbolic regression.}
Symbolic regression aims to find a concise mathematical function that accurately models the data. While it is possible to generate complex analytical expressions that fit the observations, an unnecessarily lengthy function is often impractical. Symbolic regression seeks a balance between fidelity to the data and simplicity of form.
Therefore, predictions are typically also evaluated in terms of some ``complexity'' metric.
Because the symbolic output makes it difficult to formulate differentiable losses, SR has traditionally benefitted comparably little from advances in autodifferentiation and gradient-based optimization frameworks.
The dominant approach has thus been based on evolutionary algorithms such as genetic programming (GP)
\citep{la2016epsilon,la2018learning,kommenda2020genetic,virgolin2021improving,tohme2022gsr,cranmer2023interpretable}, optionally guided by neural networks~\citep{mundhenk2021symbolic,udrescu2020ai,costa2021fast} and recently also employing reinforcement learning \citep{petersen2020deep} -- see~\citep{la2021contemporary,makke2022interpretable} for reviews.
Most of these approaches require a separate optimization for each new observed system, severely limiting scalability.
\\
\newline
\noindent\xhdr{Transformers for symbolic regression.}
With the advent of transformer models~\citep{vaswani2017attention}, efficient learning of sequence-to-sequence tasks for a broad variety of modalities became feasible.
Paired with large-scale pre-training on synthetic data, transformers have been used for symbolic tasks such as integration~\citep{lample2019deep}, formal logic~\citep{hahn2020teaching}, and theorem proving~\citep{polu2020generative}.
Few recent works applied them to functional SR~\citep{biggio2021neural,valipour2021symbolicgpt,kamienny2022end,vastl2022symformer} obtaining comparable results to GP methods, with a key advantage: after one-time pre-training, inference is often orders of magnitude faster since no training is needed for previously unseen systems.
\citet{landajuela2022unified} recently proposed a hybrid system combining and leveraging the advantages of most previous approaches for state of the art performance on functional SR tasks.
\\
\newline
\noindent\xhdr{Dynamical SR.}
In principle, dynamical SR, inferring $f$ from $(t, x(t))$, can be framed as functional SR for $(x(t), \dot{x}(t))$ pairs.
However, when transitioning from functional to dynamical symbolic regression, a key challenge is the absence of regression targets $\dot{x}(t)$ since temporal derivatives are usually not observed directly.
A common remedy is to use numerical approximations of the missing derivatives as surrogate targets instead. This approach has been employed in the GP community \citep{gaucel2014learning,la2016inference} and is also key to the widely used SINDy~\citep{sindy} algorithm which performs sparse linear regression on a manually pre-defined set of basis functions. While SINDy is computationally efficient, its modeling capacity is limited to linear combinations of its basis functions. 
Like in functional SR, neural networks have also been combined with GP for dynamical SR \citep{atkinson2019data}, and the divide-and-conquer strategy by \citet{udrescu2020ai} has also been extended to inferring dynamical systems \citep{weilbach2021inferring}.
Finally, \citet{omejc2023probabilistic} recently introduced ProGED, which performs dynamical SR via random search of candidate equations, constrained by probabilistic
context-free grammars (PCFGs). This approach does not require numerical approximations to temporal derivatives, however, the parameters of the PCFGs need to be carefully tailored to each problem, which assumes prior knowledge about the ground truth.
\\
\newline
\noindent\xhdr{Transformers for dynamical SR.}
In the realm of transformers,
\citet{d2022deep} infer one-dimensional recurrence relations in sequences of numbers, which as discrete maps are closely related to differential equations.
Most related to our work, \citet{becker2023predicting} explore a transformer-based approach to dynamical SR for ODEs.
However, their method is limited to univariate ODEs.
Such systems exhibit extremely limited behavior, where solution trajectories can only either monotonically diverge or monotonically approach a fixed value -- not even inflections let alone oscillations are possible \citep{strogatz:2000}.
In this work, we tackle the important yet unsolved task of efficiently discovering arbitrary non-linear ODE systems in symbolic form directly from data in multiple dimensions, without assuming prior knowledge of the ground truth.
\\
\newline
\noindent\xhdr{Theoretical identifiability.}
Traditionally, the identifiability of ODEs from data has been studied in a case-by-case analysis for pre-defined parametric functions  \citep{aastrom1971system,miao2011identifiability,villaverde2016structural,hamelin2020observability}.
Recent work made progress on identifiability within function classes such as linear (in parameters) autonomous ODE systems \citep{stanhope2014identifiability,duan2020identification,qiu2022identifiability} or, in the scalar case, even non-parametric classes such as analytic, algebraic, continuous, or smooth functions \citep{scholl2023uniqueness}.
As it stands, it is difficult to gain practical insights from these results for our setting as they do not conclude whether non-linear ODEs can practically be uniquely inferred from data -- see \cref{app:identifiability} for details.

\section{Data Generation}\label{sec:generation}
Our method builds on pretraining on a large dataset of ODEs which is assembled as follows.
\\
\newline
\noindent\xhdr{Generating ODEs.}
For a $\din$-dimensional ODE $f$, we independently sample the $\din$ component functions $f_1, \ldots, f_{\din}$ as random unary-binary trees following the method of~\citep{lample2019deep}, where internal nodes represent mathematical operators and leaves represent constants or variables.
In our specific procedure, we first sample the system dimension $\din$ uniformly from $[\din_{\max}]$ for a fixed $\din_{\max} \in \mathbb{N}$ and then perform the following steps for each component function:
\begin{enumerate}[leftmargin=*,noitemsep,topsep=-1mm]
    \item Sample the number of binary operators $b$ uniformly from $[b_{\max}]$ for a fixed $b_{\max} \in \mathbb{N}$.

\item Sample a binary tree with $b$ non-leaf nodes, following the procedure of ~\citet{lample2019deep}. 

\item Decorate each non-leaf node with a binary operator sampled from 
$P(+) = \nicefrac{3}{4}$ and $P(\times) = \nicefrac{1}{4}$.\footnote{Subtractions and divisions are included via multiplication with negative numbers and the unary operator $x \mapsto x^{-1}$ respectively. It has been argued that divisions appear less frequently than additions and multiplications in ``typical'' expressions~\citep{guimera2020bayesian}.}

\item For each leaf in the tree, sample one of the variables $x_i$ for $i \in [\din]$.

\item Sample the number of unary operators $u$ uniformly from $[u_{\max}]$ for a fixed $u_{\max} \in \mathbb{N}$.

\item Iteratively with $u$ repetitions, select a node whose subtree has a depth smaller than 6\footnote{This aims at avoiding deeply nested uninterpretable expressions, which often occur in GP-based SR.} and insert a new node directly above. Populate the new node with a unary operator that is sampled uniformly at random from $\{x\mapsto \sin(x), x\mapsto x^{-1}, x\mapsto x^2\}$.

\item Convert the tree into a mathematical expression via preorder traversal \citep{lample2019deep}.

\item Finally, prepend a coefficient to each term and wrap the argument of any unary operator in an affine transformation $x\to a\cdot x + b$. Coefficients and constants of affine transformations are sampled independently from a log-uniform distribution on $[c_{\min}, c_{\max}]$.

\end{enumerate}
\vspace{2mm}
\noindent Due to random continuous constants (and initial conditions), we almost surely never sample a function twice.
In our experiments, we use $\din_{\max}=6$, $b_{\max} = 5$, $u_{\max} = 3$, $(c_{\min}, c_{\max}) = (0.05, 20)$.
\\
\newline
\noindent\xhdr{Integrating ODEs.}
Once the function $f$ is generated, we sample an initial condition $x_0 \sim \mathcal{N}(0, \gamma \mathbb{I}_D)$ for a fixed $\gamma \in \R_{> 0}$ and $\mathbb{I}_D$ the identity matrix, and integrate the ODE from $t=1$ to $t=T$ using the numerical solver $\texttt{scipy.integrate.solve\_ivp}$ provided by SciPy~\citep{virtanen2020scipy} on a fixed homogeneous grid of $N$ points, where $N$ is sampled uniformly in $\{50, 51, \ldots, 200\}$.
The \texttt{solve\_ivp} function defaults to an adaptive 5th order explicit Runge Kutta method with 4th order error control \citep{dormand1980family} as well as relative and absolute tolerances of $10^{-3}$ and $10^{-6}$ respectively.
When integration fails, i.e., when the solver throws an error, returns unsuccessfully, or takes longer than one second, we simply discard the current example.
\Cref{sec:method} explains in detail that we can fix $T$ and $\gamma$ during training without loss of generality, as we can rescale observations at inference time.
Hence, we fix $\gamma = 1$ and $T = 10$ during training.
\\
\newline
\noindent\xhdr{Filtering data.}
Under the distribution over functions $f$ defined implicitly by our generator, a substantial fraction of sampled ODEs (and initial conditions) leads to solutions where at least one component diverges over time.
Continued divergence over long time spans is typically deemed ``unphysical''.
Among the trajectories that remain finite, we observe that again a substantial fraction swiftly converges towards a fixed point.
Although these solutions may be realistic, their dominance hampers diversity in our dataset. Also, stable constant observations over long times spans are arguably rarely of interest.
Hence, we use the following heuristics to increase diversity of the generated ODEs.
\begin{itemize}[leftmargin=0.5cm,nosep]
    \item If any variable of the solution trajectory exceeds a fixed threshold ($10^2$ in our experiments), we discard the example.
    This amounts to filtering out divergent systems.
    \item If the oscillation of all component functions over the last quarter of the integration range is below a certain threshold ($10^{-3}$ in our experiments), we discard the example with a probability of 90\%.\footnote{The oscillation of a function $h: [a, b] \to \R$ is given by $\sup_{x \in [a,b]} h(x) - \inf_{x \in [a,b]} f(x)$.}
    This filters out a majority of rapidly converging systems.
\end{itemize}
\ \\
\noindent\xhdr{Corrupting data.}
We apply two forms of corruption to the clean solution trajectories:
\begin{itemize}[leftmargin=0.5cm,nosep]
    \item \emph{Noise:} We sample a noise level $\sigma$ uniformly in $[0,0.1]$ and corrupt each observation of each component of the trajectory independently multiplicatively with Gaussian noise: $x_j(t_i) \to (1 + \xi) x_j(t_i)$ for $j \in [D], i \in [N]$ and $\xi \sim \mathcal{N}(0, \sigma)$.
    This noise model has been used and argued for in previous works \citep{d2022deep,becker2023predicting}.
    \item \emph{Subsampling:}  For each trajectory, we sample a subsampling ratio $\rho$ uniformly in $[0,0.5]$ and drop a fraction $\rho$ of the points along the trajectory uniformly at random.
    Since the equally spaced original trajectories contained between 50 and 200 points, after subsampling inputs can vary in length between 25 and 200.
\end{itemize}

\section{Model, Training, and Inference}\label{sec:method}

\odeformer{} is an encoder-decoder transformer ~\citep{vaswani2017attention} for end-to-end dynamical SR, illustrated in \cref{fig:sketch}. The model comprises 16 attention heads and an embedding dimension of 512, leading to a total parameter count of 86M. As observed by~\cite{charton2021linear}, we find that optimal performance is achieved in an asymmetric architecture, using 4 layers in the encoder and 16 in the decoder. Since the time component is explicitly included in the inputs, we remove positional embeddings from the encoder. Model optimization follows established procedures, with details given in \cref{app:training_details}.
\\
\newline
\noindent\xhdr{Tokenizing numbers.}
Since numeric input trajectories as well as symbolic target sequences may contain floating point values, we need an efficient encoding scheme that allows the infinite number of floats to be sufficiently well conserved by a fixed-size vocabulary.
Following \citet{charton2021linear}, each number is rounded to four significant digits and disassembled into three components: sign, mantissa and exponent, each of which is represented by its own token. This tokenization scheme condenses the vocabulary size to represent floating point values to just 10203 tokens (\texttt{+}, \texttt{-}, \texttt{0}, ..., \texttt{9999}, \texttt{E-100}, ..., \texttt{E100})  and works well in practice despite the inherent loss of precision.
We also experimented with three alternative representations:
(i) two-token encoding, where the sign and mantissa are merged together,
(ii) one-token encoding where sign, mantissa and exponent are all merged together,
(iii) a two-hot encoding inspired by \citet{schrittwieser2020mastering} and used by \citet{becker2023predicting}, which interpolates linearly between fixed, pre-set values to represent continuous values.
These representations have the advantage of decreasing sequence length, and (iii) has the added benefit of increased numerical precision for the inputs.
Since all three alternatives led to worse overall performance, we used the three token representation (sign, mantissa, exponent).
\\
\newline
\noindent\xhdr{Embedding numerical trajectories.}
The above tokenization scheme scales the length of numerical input sequences by a factor of three. 
Points $(t_i,x_i) \in \R^{\din+1}$ of the trajectory of a $\din$ dimensional ODE system are hence mapped to a token sequence of dimension $\R^{(\din+1) \times 3}$. 
We feed the token sequence of each dimension separately to an embedding layer and concatenate the result to obtain a representation in $\R^{((\din+1) \times 3) \times \demb}$. 
Before handing this representation to the encoder it is reduced such that each original input point corresponds to a single embedding vector ~\citep{kamienny2022end}, effectively scaling the input sequence back to its original length. 
For this, potentially vacant input dimensions are padded up to $D_{\max}$ before the resulting $3 \times (D_{\max} + 1) \times \demb$-dimensional vector is processed by a 2-layer fully-connected feed-forward network (FFN) with Sigmoid-weighted linear unit (SiLU) activations~\citep{elfwing2018sigmoid}, which projects down to dimension $\demb$. This embedding process allows a single trained model to flexibly handle input trajectories of variable lengths as well as ODE systems of different dimensionalities.
\\
\newline
\noindent\xhdr{Encoding symbolic functions.}
To encode mathematical expressions, the vocabulary of the decoder includes specific tokens for all operators and variables, in addition to the tokens used to represent floating point values. Importantly, the decoder is trained on expressions in prefix notations to relieve the model from predicting parentheses ~\citep{lample2019deep}.
With these choices, the target sequence for an exemplary ODE $f(x)=\cos(2.4242 x)$ corresponds to following sequence of six tokens $\texttt{[cos\ mul\ +\ 2424\ E-3\ x]}$.
For $\din$-dimensional systems, we simply concatenate the encodings of the $\din$ component functions, separated by a special token ``$\texttt{|}$''.
With this simple method, the sequence length scales linearly with the dimensionality of the system, i.e., the number of variables.
While this is unproblematic for small dimensions such as $\din \le \din_{\max} = 6$, it may impair the scalability of our approach.\footnote{A possible alternative would be to treat the decoding of each component function as a separate problem, adding a specifier to the $\texttt{BOS}$ (beginning of sequence) token to identify which component is to be decoded.}
As the encoder is only concerned with numeric input trajectories, its vocabulary only includes tokens for numbers.
\\
\newline
\noindent\xhdr{Rescaling.}
During training, the model only observes initial conditions from a standard normal distribution, and the integration range is fixed to $[1,10]$.
To accommodate for different scales of initial conditions and time ranges during inference, we apply the affine transformations $t \to \tilde{t} = a t + b$ to rescale the observed time range to $[1,10]$ and $x_i(t) \to \tilde{x}_i(t) = \tfrac{x_i(t)}{x_i(t_0)}$ to rescale initial values to unity. 
The prediction $\tilde{f}$ that \odeformer{} provides on inputs $(\tilde{t}, \tilde{x})$ are then transformed as $f_i = \tfrac{dx_i}{dt} = \tfrac{1}{ax_i(t_0)}\tfrac{d\tilde{x}_i}{d\tilde{t}} = \tfrac{1}{a x_i(t_0)}\tilde{f_i}$ to recover original units.
\\
\newline
\noindent\xhdr{Decoding strategy.}
At inference, we use beam sampling \citep{van2008beam} to decode candidate equations, and select the candidate with highest reconstruction $R^2$ score.\footnote{Beam search tends to produce candidates which all have the same skeleton, and only differ by small variations of the constants, leading to a lack of diversity. Beam sampling ensures that randomness is added at each step of decoding leading to a more diverse set of candidate expressions.}
The beam temperature is an important parameter to control diversity -- as the beam size increases, it typically becomes useful to also increase the temperature to maintain diversity.
Unless stated otherwise, we perform our experiments with a beam size of 50 and a temperature of 0.1.
\\
\newline
\noindent\xhdr{Optional parameter optimization.}
Most SR methods break the task into two subroutines (possibly alternating between the two): predicting the optimal ``skeleton'', i.e., equation structure, and fitting the numerical values of the constants.
Just like~\citet{kamienny2022end}, our model is end-to-end, in the sense that it handles both subroutines simultaneously.
However, we also experimented with adding an extra parameter optimization step, as performed in methods such as ProGED~\citep{omejc2023probabilistic}. We describe the details of the parameter optimization procedure in \cref{app:optional_parameter_optimization} and denote this method as ``\odeformer{} (opt)''.

\section{Evaluation}\label{sec:experiments}
\noindent\xhdr{Symbolic vs numerical evaluation.}
When evaluating SR methods, the desired metric is whether the inferred ODE $\hat{f}$ perfectly agrees symbolically with the ground truth expression $f$.
However, such an evaluation is problematic primarily because
(i) the ambiguity in representing mathematical expressions and non-determinism of ``\texttt{simplify}'' operations in modern computer algebra systems render comparisons on the symbolic level difficult, and
(ii) expressions may differ symbolically, while evaluating to essentially the same outputs on all possible inputs (e.g., in the presence of a negligible extra term).
Hence, comparing expressions numerically on a relevant range is more reliable and meaningful.
Even modern computer algebra systems include numerical evaluations in equality checks, which still require choosing a range on which to compare expressions \citep{meurer2017sympy}.
\\
\newline
\noindent\xhdr{Evaluation types.}
For dynamical SR there is a spectrum of reasonable comparisons.
We could simply be interested in finding \emph{some} ODE $\hat{f}$, whose solution approximates the observed trajectory $x(t)$ on the observed time interval, even if $\hat{f}$ still differs from $f$ on (parts of) their domain.
A more ambitious goal closer to full identification is to find a $\hat{f}$ that also approximates the correct trajectories for unobserved initial conditions.
This evaluation, which in our view is most meaningful to assess dynamical SR, is often absent in the literature, e.g., from \citet{omejc2023probabilistic}.
If an inferred $\hat{f}$ yields correct solutions for all initial values and time spans, we would consider it as perfect identification of $f$.
Accordingly, we evaluate the following aspects of performance in our experiments.
\begin{itemize}[leftmargin=0.5cm,nosep]
    \item \emph{Reconstruction:} we compare the (noiseless, dense) ground truth trajectory with the trajectory obtained by integrating the predicted ODE from the same initial condition and on the same interval $[1,T]$ as the ground truth.
    \item \emph{Generalization:} we integrate both the ground truth and the inferred ODE for a new, different initial condition over the same interval $[1,T]$ and compare the obtained trajectories.
\end{itemize}
\ \\
\noindent\xhdr{Metrics.}
We consider the following performance metrics.
\begin{itemize}[leftmargin=0.5cm,nosep]
    \item \emph{Accuracy:} A classical performance metric for regression tasks is the coefficient of determination, defined as $R^2 = 1-\frac{\sum_i\left(y_i-\hat{y}_i\right)^2}{\sum_i\left(y_i-\bar{y}\right)^2} \in (-\infty, 1]$. Since $R^2$ is unbounded from below, average $R^2$-scores across multiple predictions may be severely biased by even a single particularly poor outlier. We therefore instead report the percentage of predictions for which the $R^2$ exceeds a threshold of 0.9 in the main text and show the distribution of scores in \cref{app:results}.
    \item \emph{Complexity:} we define the complexity of a symbolic expression to be the total number of operators, variables and constants. We acknowledge that this is a crude measure, for example, it would assign a lower complexity to $\exp(\tan(x))$ (complexity=3) than to $1+2x$ (complexity=5).
    Yet, there is no agreed-upon meaning of the term ``complexity'' for a symbolic expression in this context so that
    simply counting ``constituents'' is common in the literature, see e.g. the discussion in the benchmark by \citet{la2021contemporary}.
    \item \emph{Inference time:} time to produce a prediction.
\end{itemize}
\ \\
\noindent\xhdr{Corruptions.}
For all datasets, we also compare models on their robustness to two types of corruption: (i) we add noise to the observations via $x_j(t_i) \to (1 + \xi) x_j(t_i)$ for $j \in [D], i \in [N]$ and $\xi \sim \mathcal{N}(0, \sigma)$; (ii) we drop a fraction $\rho$ of the observations along the trajectory uniformly at random.
We report results for various noise levels $\sigma$ and subsampling fractions $\rho$.

\begin{figure}[htb]
    \centering
    \includegraphics[width=\linewidth]{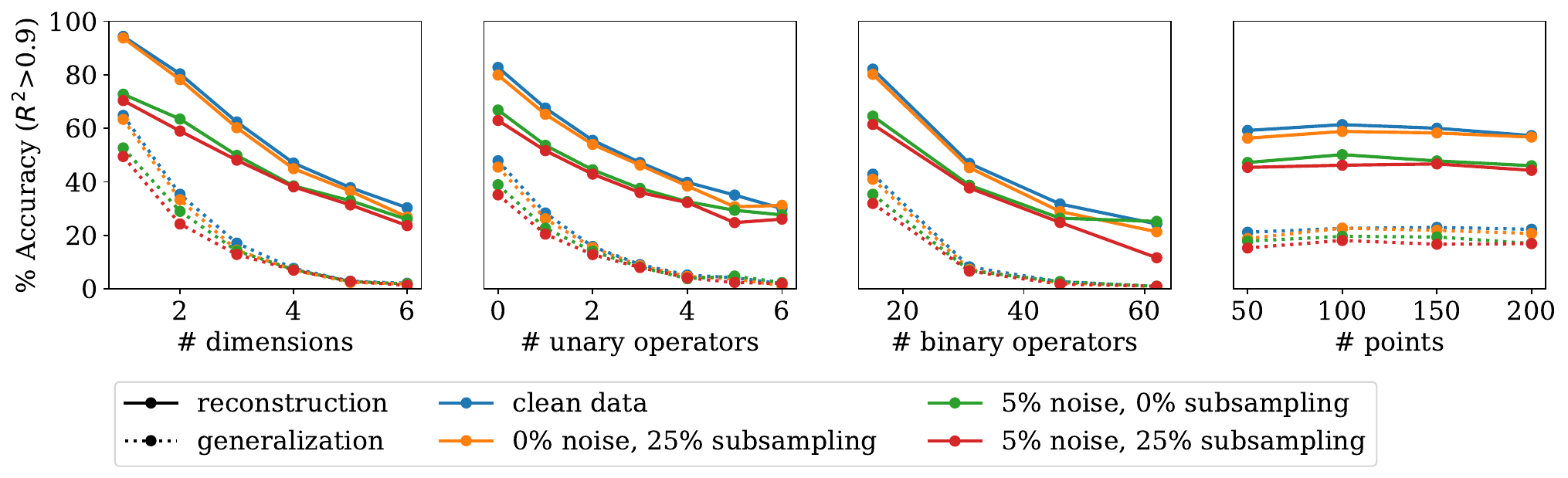}%
    \caption{\textbf{Ablation study on synthetic data.} We vary four parameters governing the difficulty of an example (from left to right): the dimension of the system, the total number of unary and binary operators, and the number of points in the trajectory. In each panel, we average the results over a dataset of 10,000 examples. In all cases, we use a beam size of 50.}\label{fig:synthetic}
    \vspace{-2mm}
\end{figure}

\noindent\xhdr{Results on synthetic data.}
We first assess how the performance of \odeformer{} is affected by the dimensionality of the ODE system, the number of unary and binary operators, and the number of points in the trajectory.
The ablation results on 10,000 synthetic examples with a beam size of 50 are shown in \cref{fig:synthetic} -- we asses the effect of the beam size in \cref{app:beamsize}. We make three observations:
\begin{itemize}[noitemsep, leftmargin=0.5cm,topsep=-1mm]
    \item Performance degrades with the first three parameters as expected, but \odeformer{} is surprisingly insensitive to the number of points in the trajectory.
    \item Generalization accuracy is substantially lower than reconstruction accuracy as expected, but at least for low-dimensional systems we achieve non-trivial generalization (e.g., 60\% generalization accuracy vs 85\% reconstruction accuracy for 1D).
    \item \odeformer{} copes well with subsampling, but suffers more from noisy trajectories. However, the effect on generalization is smaller than that on reconstruction.
\end{itemize}

\section{Benchmarking Dynamical Symbolic Regression Methods}\label{sec:results}
\xhdr{Strogatz benchmark.}
We first consider the ``Strogatz dataset'', included in the Penn Machine Learning Benchmark (PMLB) database \citep{la2021contemporary}.
It consists of seven ODE systems and has been used as a benchmark by various SR methods, in particular those specialized on dynamical SR \citep{omejc2023probabilistic}.
However, it has several limitations: (i) it is small (only seven unique ODEs, each integrated for 4 different initial conditions), (ii) it only contains 2-dimensional systems, (iii) it is not integrated with sufficient precision \citep{omejc2023probabilistic}, and (iv) its annotations are misleading (e.g., claiming that all systems develop chaos even though none of them does).
\\
\newline
\noindent\xhdr{\odebench{}.}
Faced with these limitations, we introduce  \odebench{}, an extended benchmark curated from ODEs that have been used by \citet{strogatz:2000} to model real-world phenomena as well as well-known systems from Wikipedia.
We fix parameter values to obtain the behavior the models were developed for, and choose two initial conditions for each equation to evaluate generalization.
\odebench{} consists of 63 ODEs (1D: 23, 2D: 28, 3D: 10, 4D: 2), four of which exhibit chaotic behavior.
We publicly release \odebench{} with descriptions, sources of all equations, and well integrated solution trajectories -- more details are in \cref{app:odebench}.
\begin{table*}[b]
\centering
\vspace{-3mm}
\caption{\textbf{Overview of models.} f.d.: finite differences required, ode: method developed for dynamical SR, T: transformer-based, GP: genetic programming, MC: Monte Carlo, reg: regression}
\label{tab:baselines}
\vspace{-2mm}
{\small
\begin{tabular}{cccccc}
  \toprule
    \textbf{name} & \textbf{type} & \textbf{ode} & \textbf{f.d.} & \textbf{description} & \textbf{reference} \\ \midrule
    ODEFormer & T & yes & no & seq.-to-seq. translation& ours\\
    AFP        & GP & no & yes & age-fitness Pareto optimization & \citep{schmidt2011age} \\
    FE-AFP     & GP & no & yes & AFP with co-evolved fitness estimates & \citep{schmidt2011age} \\
    EHC        & GP & no & yes & AFP with epigenetic hillclimbing  & \citep{la2016thesis} \\
    EPLEX      & GP & no & yes & epsilon-lexicase selection & \citep{la2016epsilon} \\
    PySR       & GP & no & yes & AutoML-Zero + simulated annealing & \citep{cranmer2023interpretable} \\
    SINDy      & reg    & yes & yes & sparse linear regression & \citep{sindy} \\
    FFX        & reg    & no & yes & pathwise regularized ElasticNet regression & \citep{mcconaghy2011ffx} \\
    ProGED     & MC   & yes & no & MC on probabilistic context free grammars & \citep{omejc2023probabilistic} \\
    \bottomrule
\end{tabular}}%
\end{table*}
\\
\newline
\noindent\xhdr{Baselines.}
In our experiments, we extensively compare \odeformer{} with (strong representatives of) existing methods described in \cref{tab:baselines}.
For each baseline model, we perform a separate hyperparameter optimization for each run to ensure maximal fairness.
Apart from ProGED and SINDy, all baselines were developed for functional SR. We use them for dynamical SR as described in \cref{sec:related_work}, by computing temporal derivatives $\dot{x}_i(t)$ via finite differences with hyperparameter search on the approximation order and optional use of a Savitzky-Savgol filter for smoothing. For more details, please refer to \cref{app:baselines}. Note that our method is the only one which does not require any hyperparameter tuning or prior knowledge on the set of operators to be used.
\\\newline

\begin{figure}[h]
    \vspace{-5mm}
    \centering
        \hspace{2cm}
        \begin{subfigure}[b]{0.8\linewidth}
        \includegraphics[width=\linewidth]{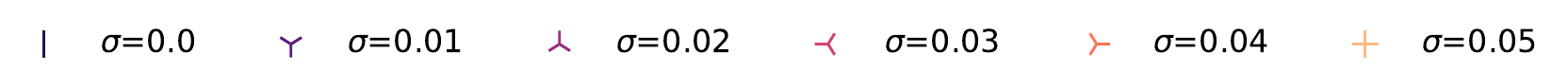}
    \end{subfigure}\\[2mm]
    \begin{subfigure}[b]{1\linewidth}
    \includegraphics[width=\linewidth]{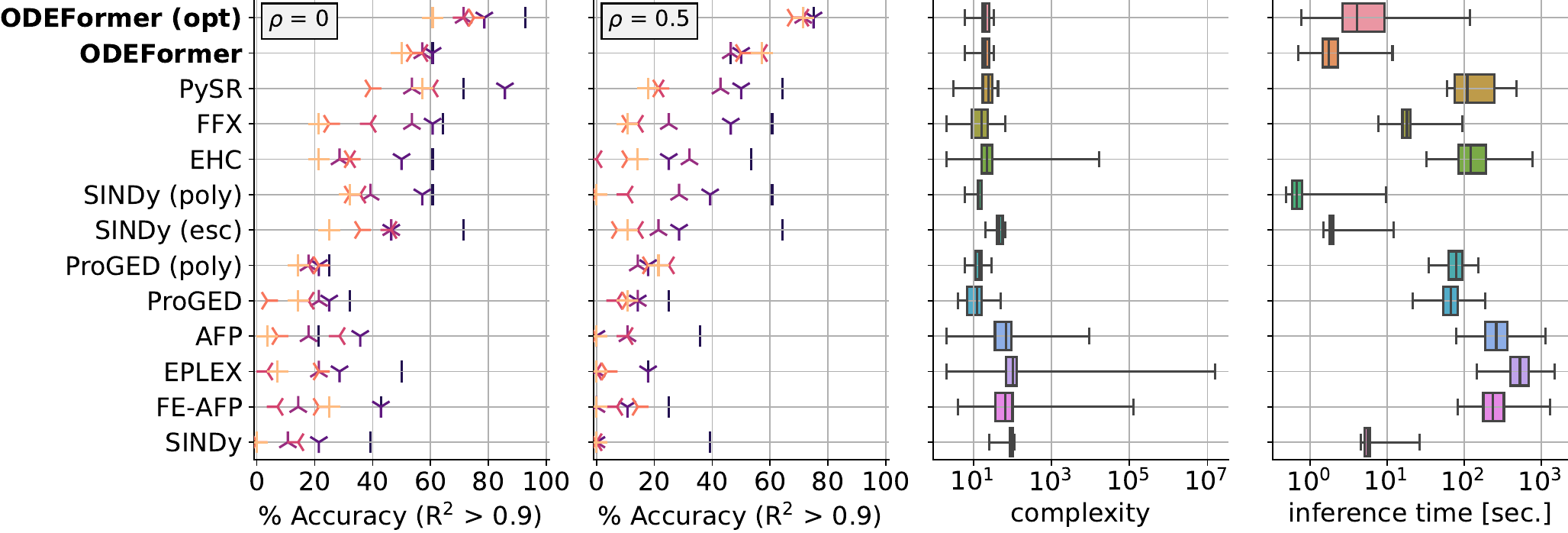}
        \caption{Reconstruction on Strogatz}
    \end{subfigure}\\[2mm]
    \begin{subfigure}[b]{1\linewidth}
        \includegraphics[width=\linewidth]{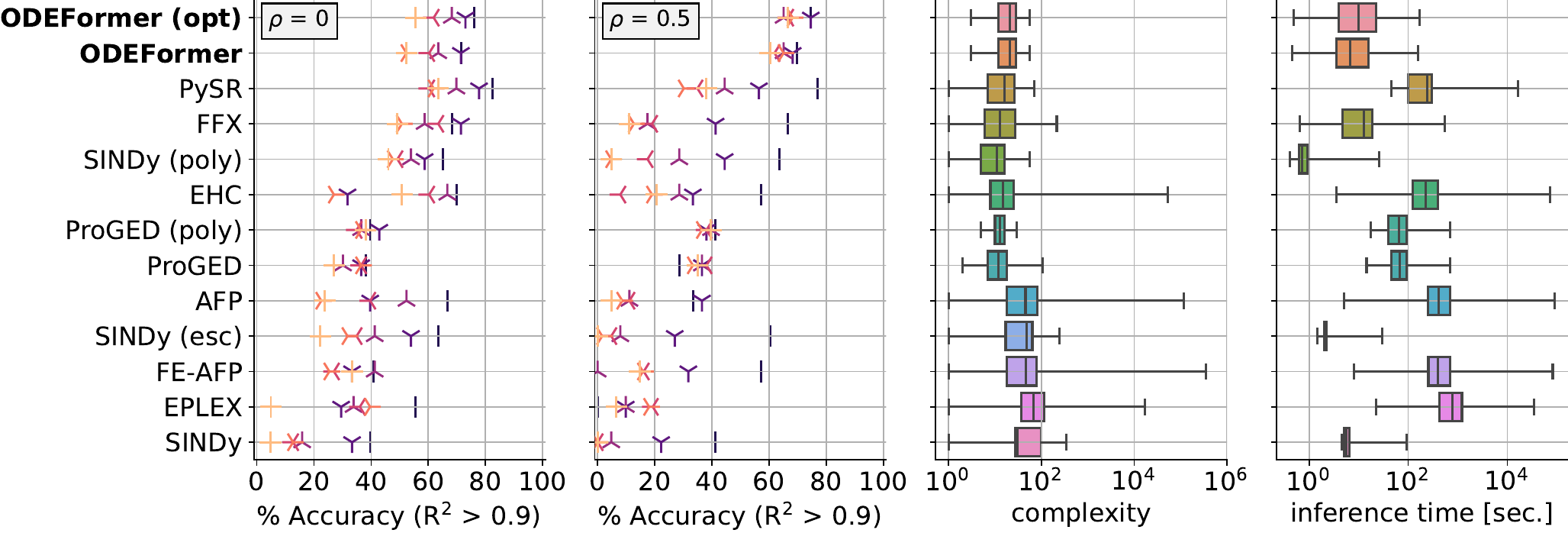}
        \caption{Reconstruction on \odebench{}}
    \end{subfigure}
    \caption{\textbf{Our model achieves state-of-the art performance on both benchmarks considered, while achieving higher robustness to noise and irregular sampling.} We compare \odeformer{} with and without additional parameter optimization using existing methods following the protocol described in \cref{sec:results}. We present results for two values of the subsampling parameter $\rho$ and six values of the noise parameter $\sigma$. Whiskers in box plot panels mark minimum and maximum values.}\label{fig:benchmarks}
\end{figure}
\noindent\xhdr{Reconstruction results.}
We present results on both ``Strogatz'' and \odebench{} in \cref{fig:benchmarks}. 
From top to bottom, we ranked methods by their average accuracy across all noise and subsampling levels.

The ranking is similar on the two benchmarks and \odeformer{} achieves the highest average score on both. 
The two leftmost panels show that \odeformer{} is only occasionally outperformed by PySR when the data is very clean -- as noise and subsampling kick in, \odeformer{} gains an increasingly large advantage over all other methods.
In the two rightmost panels of \cref{fig:benchmarks}, we show the distributions of complexity and inference time. \odeformer{} runs on the order of seconds, versus minutes for all other methods except SINDy, while maintaining relatively low and consistent equation complexity even at high noise levels. We show figures for all predictions in \cref{app:odebench}.
\\
\clearpage
\begin{wrapfigure}{r}{0.5\linewidth}
    \vspace{-.6cm}
    \begin{minipage}{1\linewidth}
        \includegraphics[width=\linewidth]{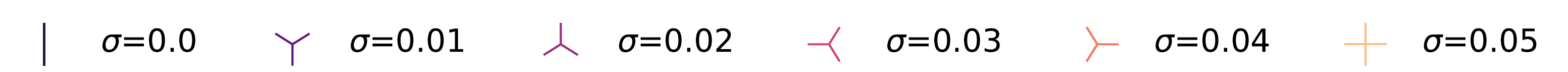}
        \includegraphics[width=\linewidth]{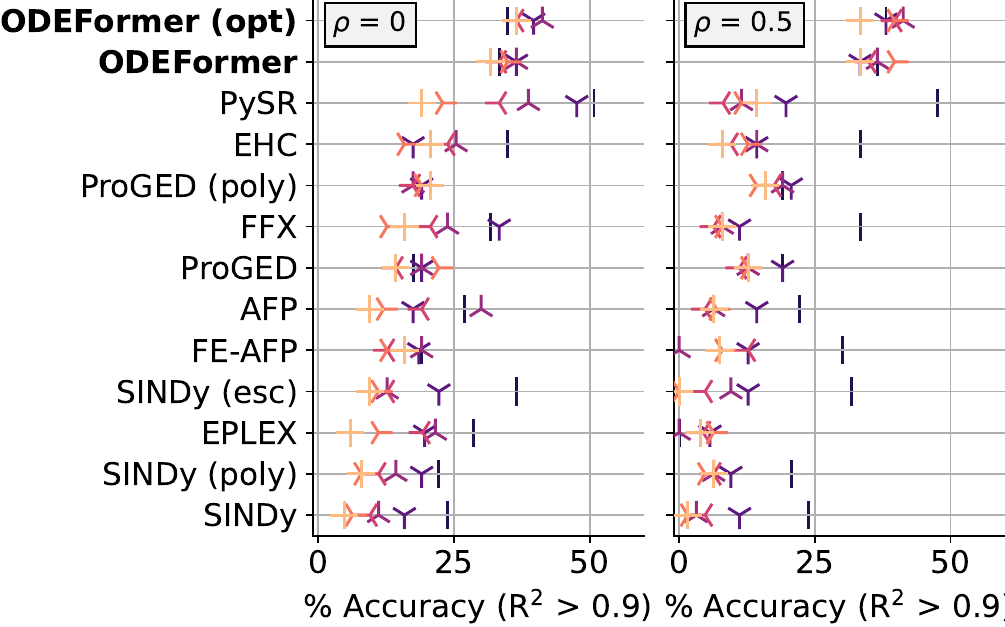}%
    \caption{\textbf{Generalization on \odebench{}.} We consider the same setting as in \cref{fig:benchmarks}.}
    \vspace{-1cm}
        \label{fig:generalization}
    \end{minipage}
\end{wrapfigure}
\noindent\xhdr{Generalization results.} We present generalization results on \odebench{} in \cref{fig:generalization}. Consistently across all models, accuracies drop by about half, meaning that half the correctly reconstructed ODEs do not match the ground truth symbolically. This highlights the importance of evaluating dynamical SR on different initial conditions.
Note, however, that the overall rankings of the different methods is rather consistent with the reconstruction results, and
\odeformer{} achieves the best results on average thanks to its robustness to noise and subsampling.

\section{Discussion and Future Directions}
We presented \odeformer{}, the first transformer capable of inferring multidimensional ODE systems from noisy, irregularly observed solution trajectories, as well as \odebench{}, a novel benchmark dataset for dynamical SR.
In extensive comparisons, we demonstrate that our model outperforms existing methods while allowing faster inference.
We foresee real-world applications of \odeformer{} across the sciences, for hypothesis generation of dynamical laws underlying experimental observations.
However, in the following we also highlight several limitations of the current method, opening up interesting directions for future work.
First, we only considered first order ODEs.
While any higher-order ODE can be written as a system of first order ODEs, this does not immediately allow \odeformer{} to make predictions based only on a solution trajectory, since we would still need to approximate time derivatives up to the order of the ODE. 
While possible in principle via finite differencing schemes, we would suffer similar drawbacks as other methods that rely on finite differences from noisy, irregularly sampled data.
Second, \odeformer{} only works when all variables are observed.
In real-world settings, some relevant variables may be unknown or unobservable.
For example, inferring chemical kinematics may be challenged by the difficulty in measuring the concentration of reaction intermediates~\citep{bures2023organic}.
We may circumvent this issue by randomly masking variables during training (replacing their observations by a dedicated token), to emulate unobserved variables.
While this raises questions around identifiability and robustness, we plan to explore masking in future work.
This technique could also handle higher-order ODEs by considering the derivatives as unobserved.

Third, based on the four representatives in \odebench{}, \odeformer{} (as well as all other benchmarked models) struggles with chaotic systems, which have been argued to provide good assessments for data-driven modeling and forecasting \citep{gilpin2021chaos}.
For chaotic systems, understanding properties of the attractor is often more desirable: nearby trajectories diverge exponentially, rendering identification notoriously challenging.
Dynamical law learning from short (transient) trajectories for chaotic systems remains an interesting direction for future work.

Lastly, all existing methods for dynamical SR, including ours, perform inference based on a single observed trajectory.
In our opinion, one of the most promising directions for future work is to enable inference from multiple solution trajectories of the same ODE.
Key benefits may include averaging out possible noise sources as well as improving identifiability, as we ``explore the domain of $f$''.
However, initial experiments with various forms of logit aggregation in \odeformer{}'s decoder during inference did not yield convincing results.
In future work, we plan to exploit a cross-attention encoder to combine the embeddings of the different trajectories to leverage the combined information, in the spirit of~\cite{liu2023mmvit}.

We conclude by emphasizing that the difficulty and potential ambiguity in dynamical law learning calls for caution during deployment.
Methods like \odeformer{} primarily serve as hypothesis generators, ultimately requiring further experimental verification, something that can be done well based on ODEs (rather than black-box models).
\vspace{-2mm}

\clearpage
\section*{Acknowledgements}

We thank Luca Biggio, François Charton and Tommaso Bendinelli for insightful conversations. We acknowledge funding from the EPFL's AI4science program (SA), Helmholtz Association under the joint research school ``Munich School for Data Science - MUDS" (SB), Swiss National Science Foundation grant (310030$\_$212516) (AM), NCCR Catalysis (grant number 180544), and a National Centre of Competence in Research funded by the Swiss National Science Foundation (PS). This work was supported by the Helmholtz Association’s Initiative and Networking Fund on the HAICORE@FZJ partition.

\printbibliography

\clearpage
\appendix
\section{\odebench{}}\label{app:odebench}

\odebench{} features a selection of ordinary differential equations primarily from Steven Strogatz's book ``Nonlinear Dynamics and Chaos'' with manually chosen parameter values and initial conditions~\citep{strogatz:2000}.
Some other famous known systems have been selected from other sources such as Wikipedia, which are included in the dictionary entries as well.
We selected ODEs primarily based on whether they have actually been suggested as models for real-world phenomena as well as on whether they are `iconic' ODEs in the sense that they are often used as examples in textbooks and/or have recognizable names.
Whenever there were `realistic' parameter values suggested, we chose those.

In this benchmark, we typically include only one set of parameter values per equation.
Many of the ODEs in Strogatz' book are analyzed in terms of the different limit behavior for different parameter settings.
For some systems that exhibit wildly different behavior for different parameter settings, we include multiple sets of parameter values as separate equations (e.g., the Lorenz system in chaotic and non-chaotic regimes).
For each equation, we include two sets of manually chosen initial conditions.

\begin{figure}[!ht]
    \centering
    \includegraphics[width=\linewidth]{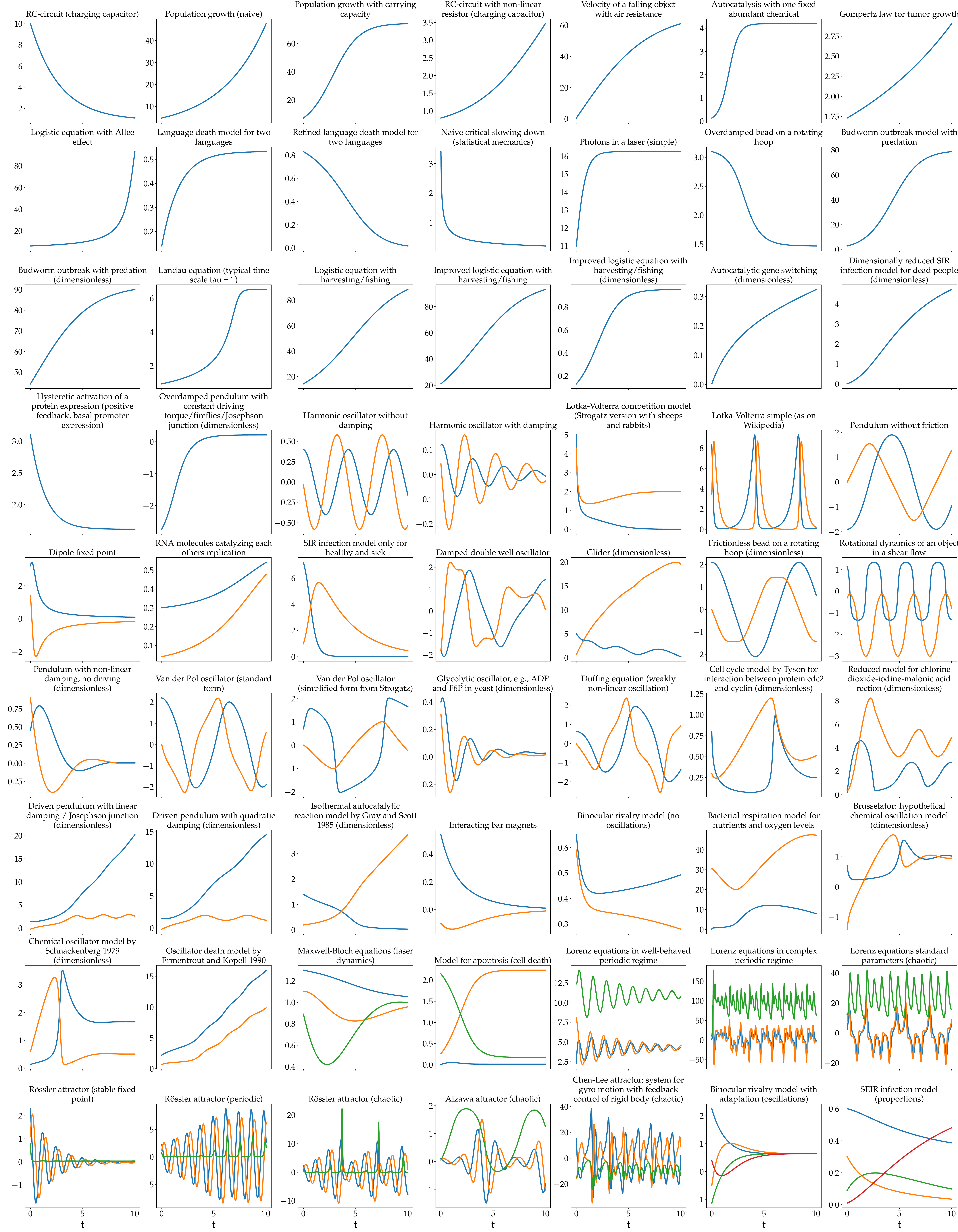}
    \caption{Solution trajectories of all equations in \odebench{} for one of the initial conditions.}
    \label{fig:odebench-plots}
\end{figure}

\begin{figure}[htb]
    \centering
\includegraphics[width=0.13\linewidth]{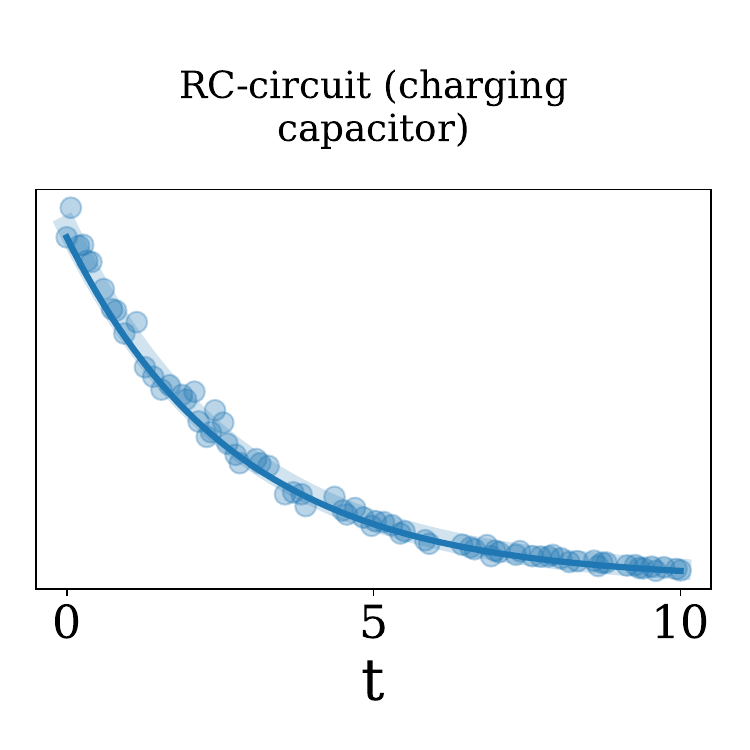}
\includegraphics[width=0.13\linewidth]{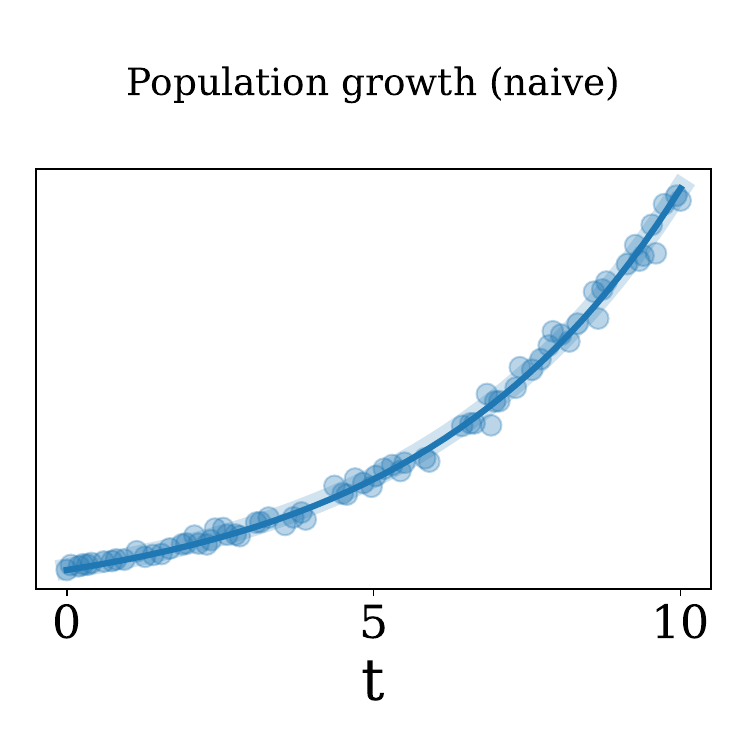}
\includegraphics[width=0.13\linewidth]{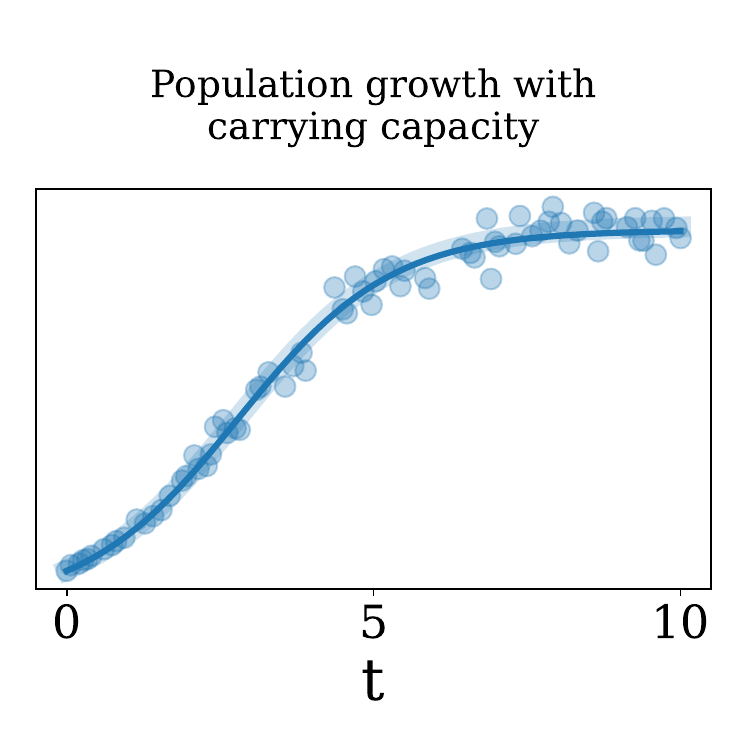}
\includegraphics[width=0.13\linewidth]{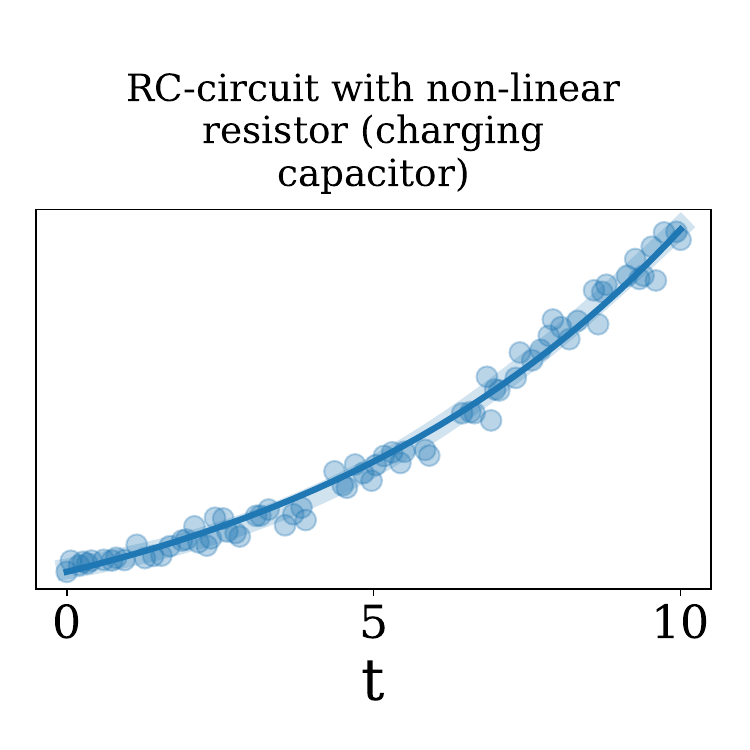}
\includegraphics[width=0.13\linewidth]{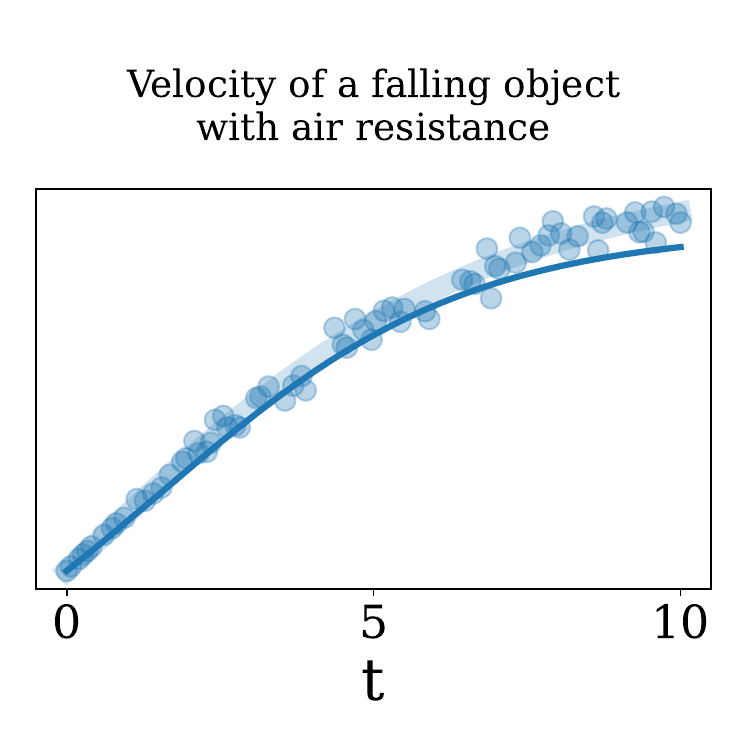}
\includegraphics[width=0.13\linewidth]{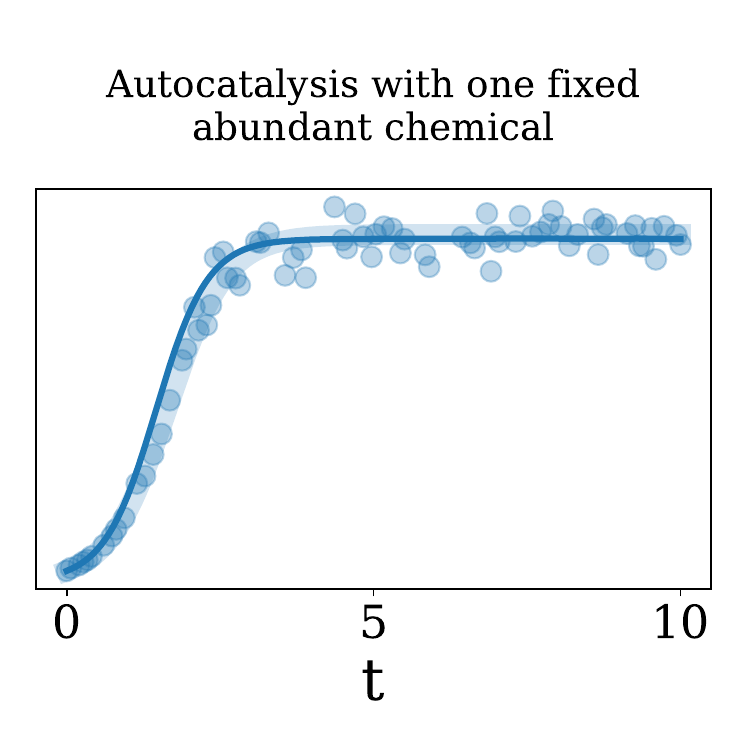}
\includegraphics[width=0.13\linewidth]{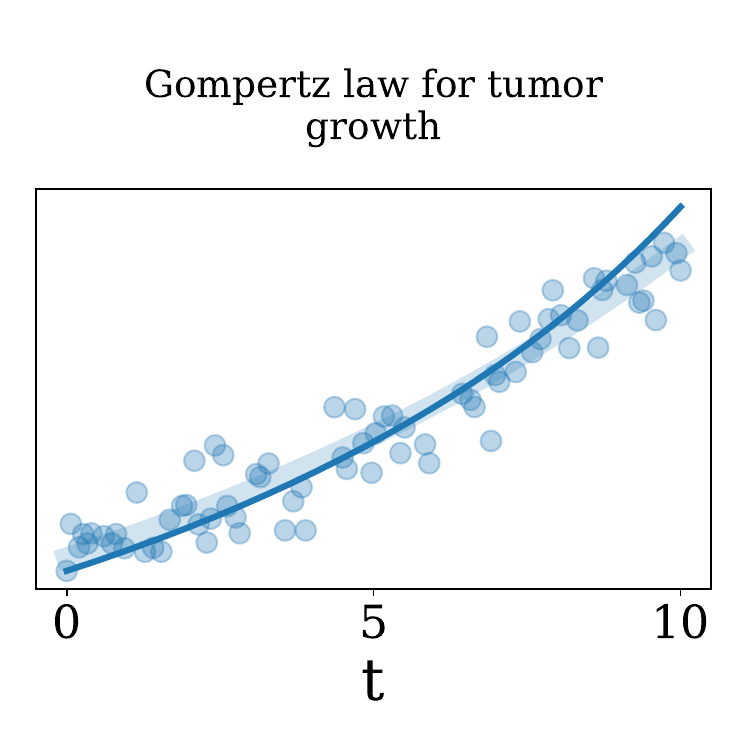}
\includegraphics[width=0.13\linewidth]{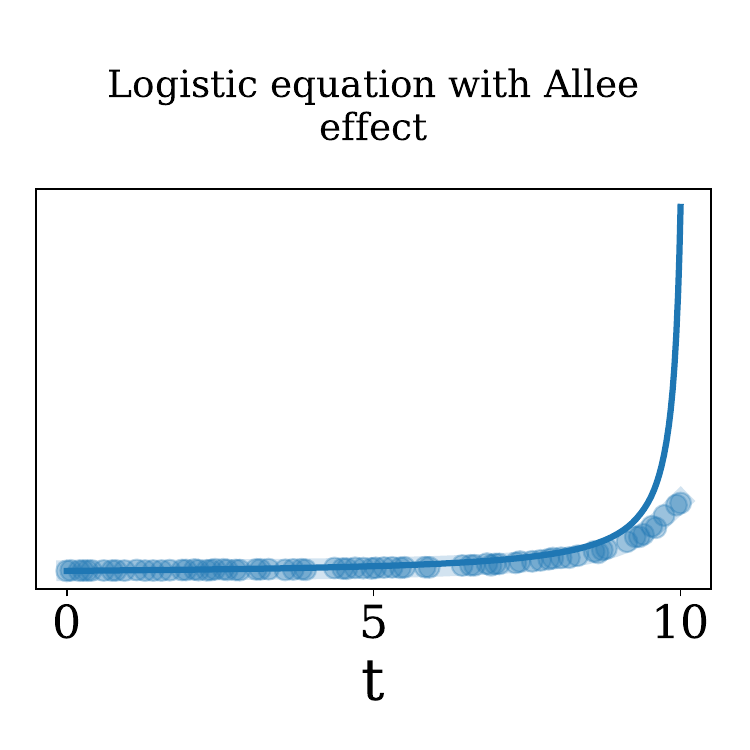}
\includegraphics[width=0.13\linewidth]{figs/odebench/odebench_9.pdf}
\includegraphics[width=0.13\linewidth]{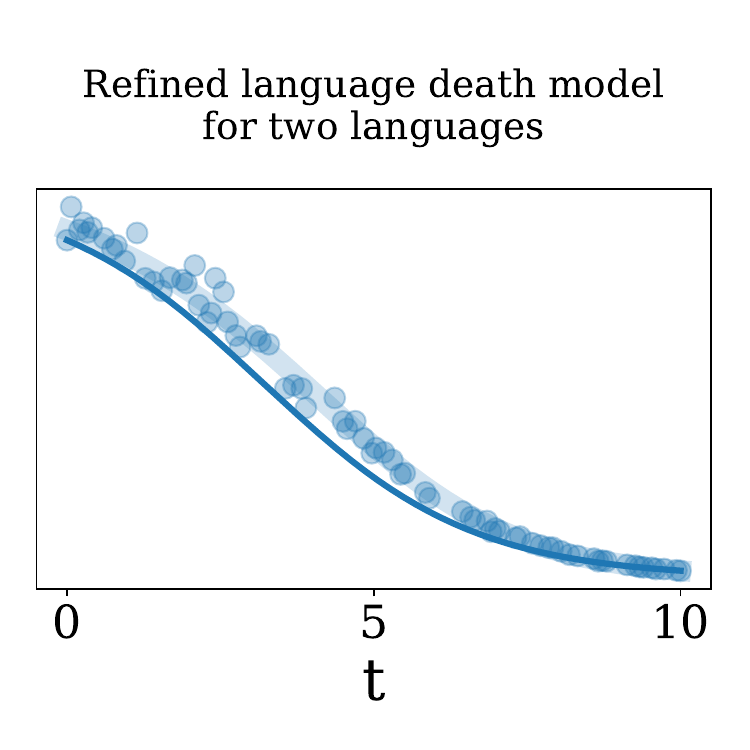}
\includegraphics[width=0.13\linewidth]{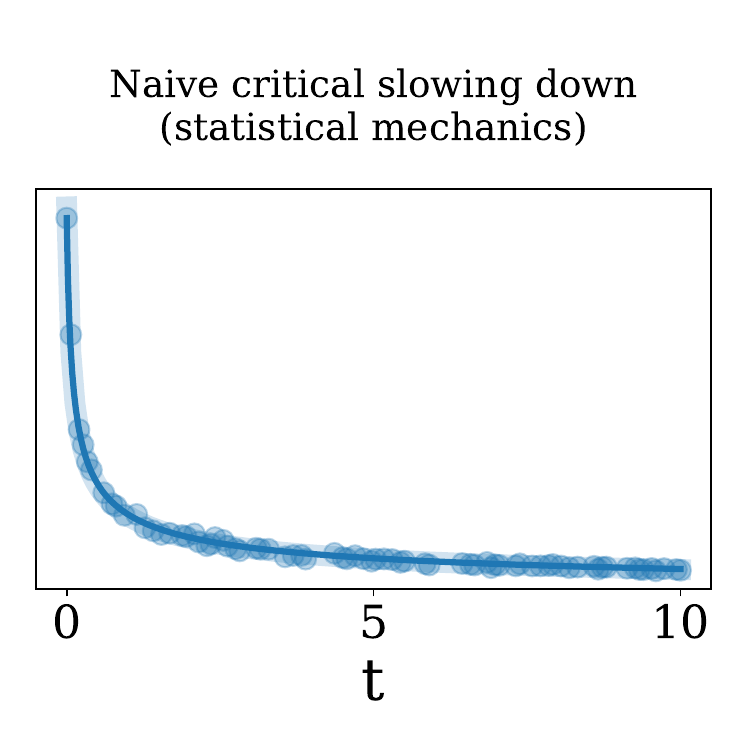}
\includegraphics[width=0.13\linewidth]{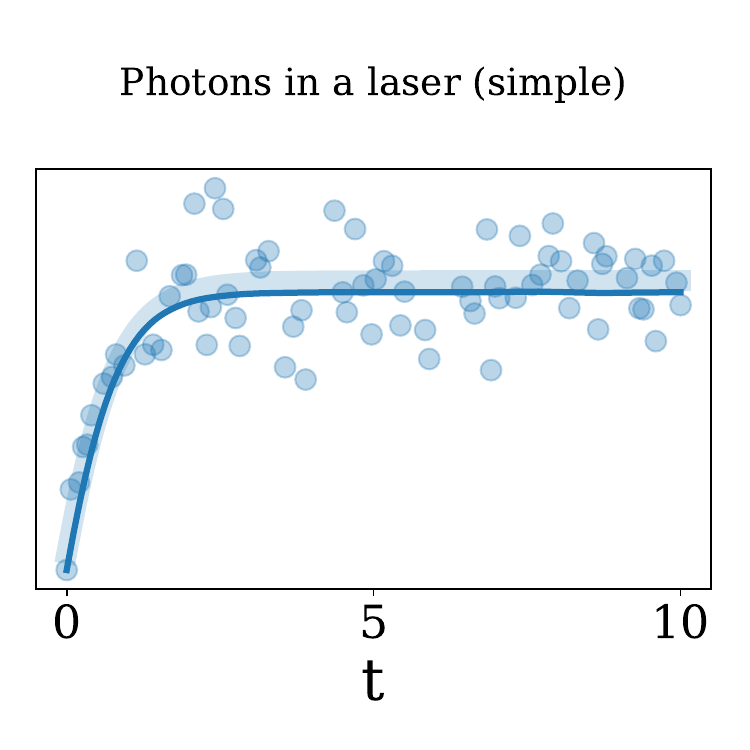}
\includegraphics[width=0.13\linewidth]{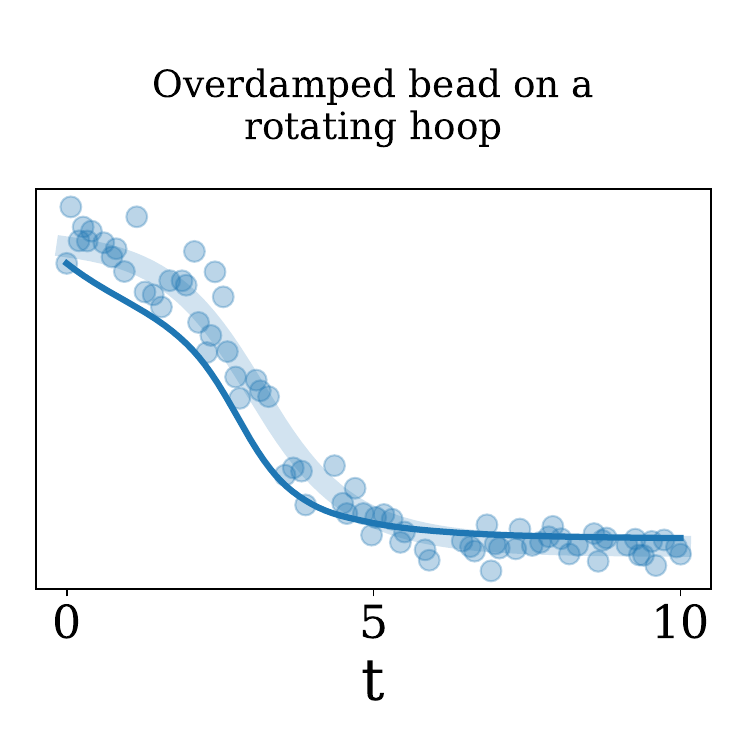}
\includegraphics[width=0.13\linewidth]{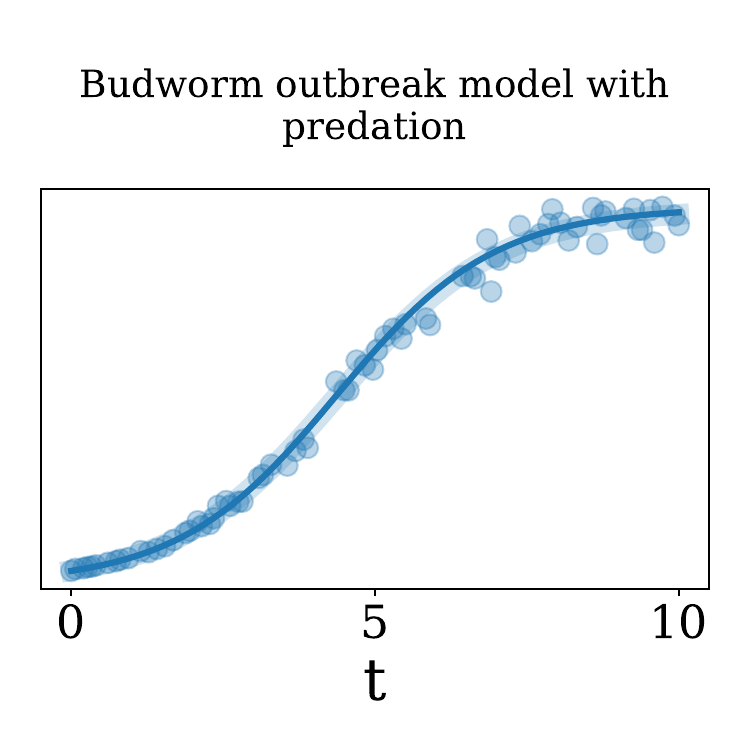}
\includegraphics[width=0.13\linewidth]{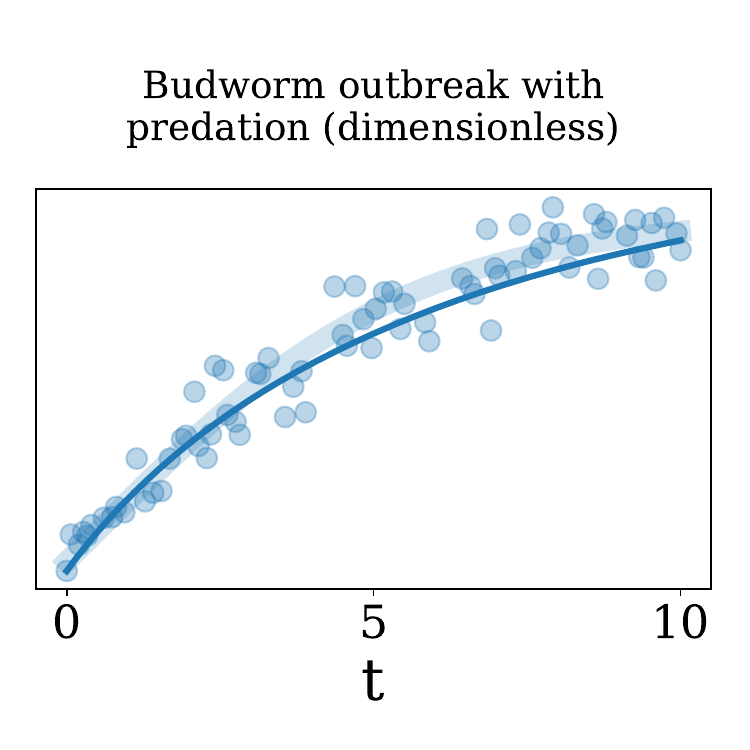}
\includegraphics[width=0.13\linewidth]{figs/odebench/odebench_16.pdf}
\includegraphics[width=0.13\linewidth]{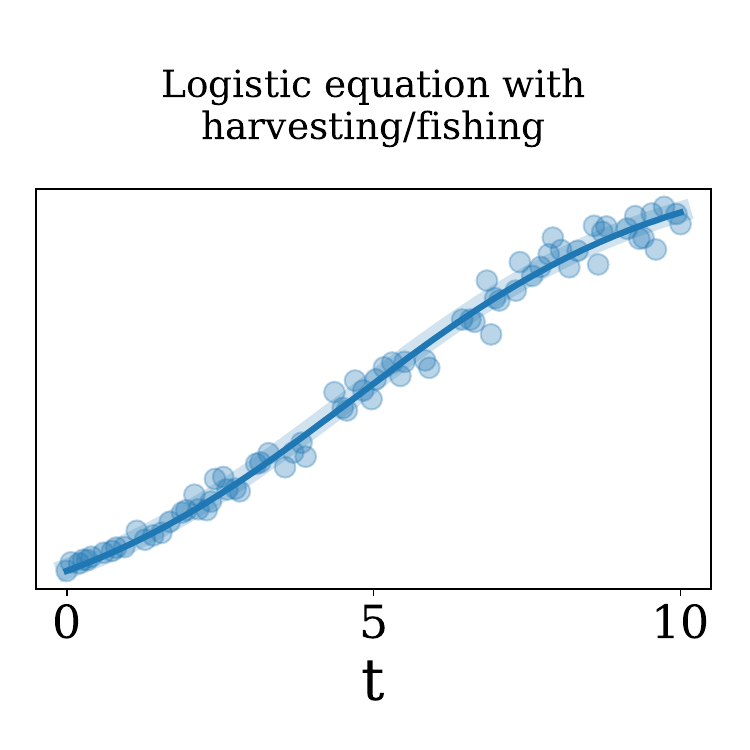}
\includegraphics[width=0.13\linewidth]{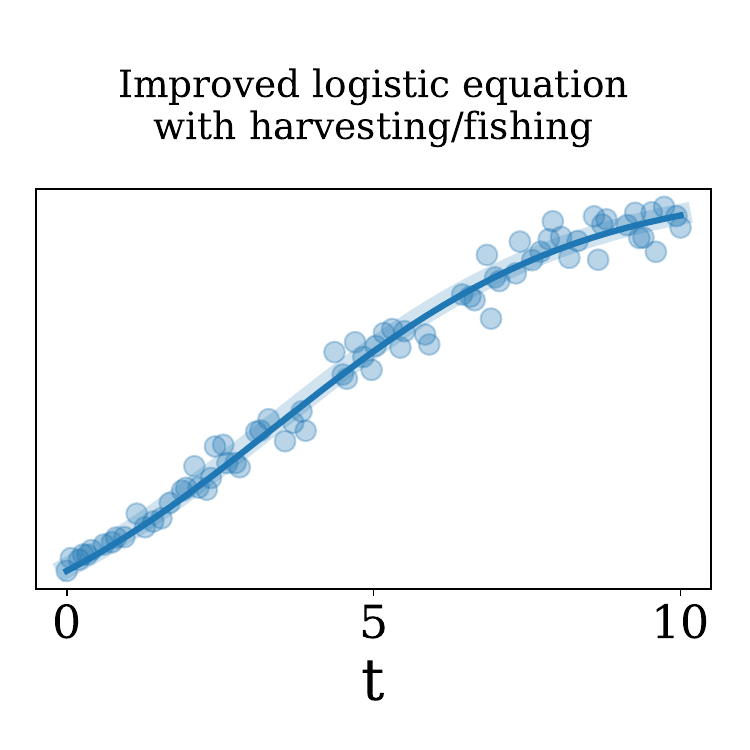}
\includegraphics[width=0.13\linewidth]{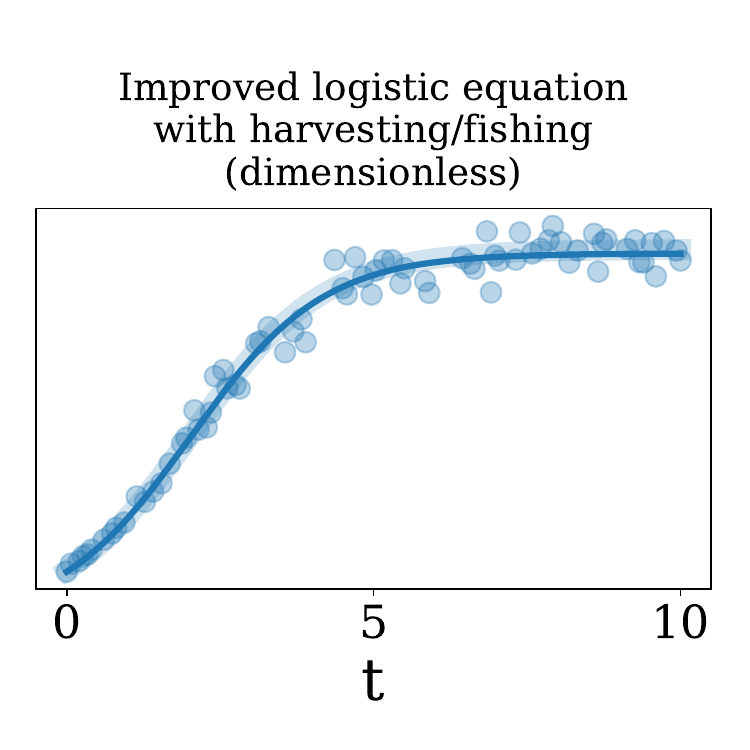}
\includegraphics[width=0.13\linewidth]{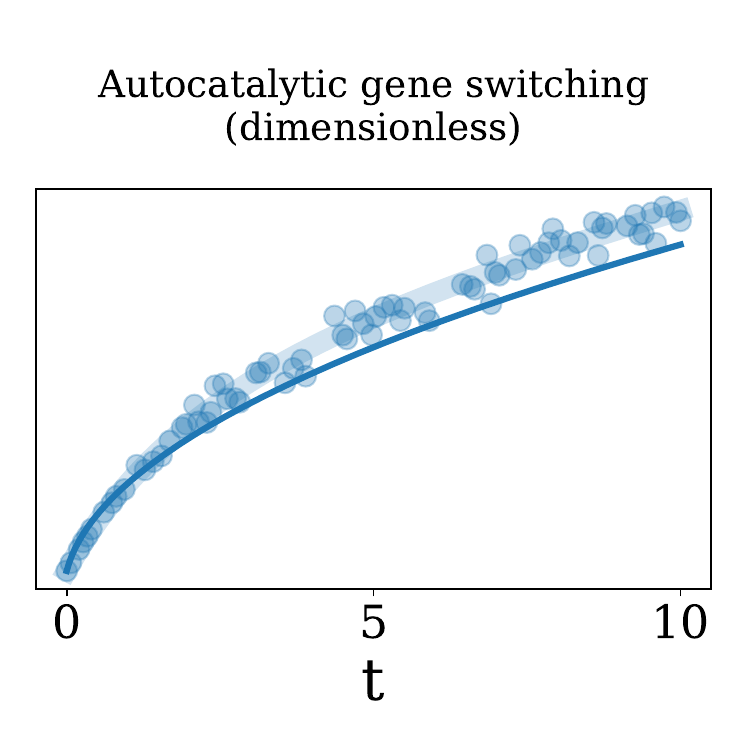}
\includegraphics[width=0.13\linewidth]{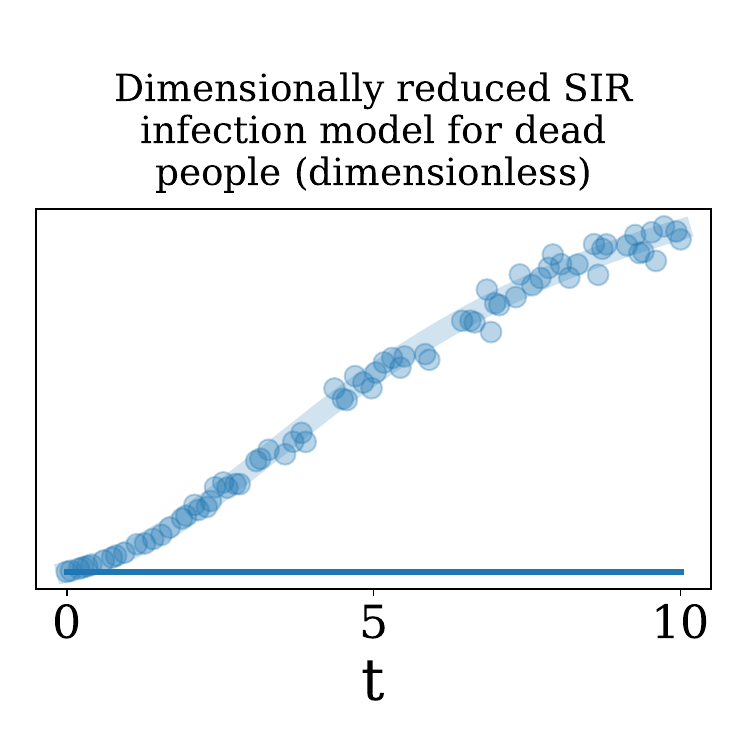}
\includegraphics[width=0.13\linewidth]{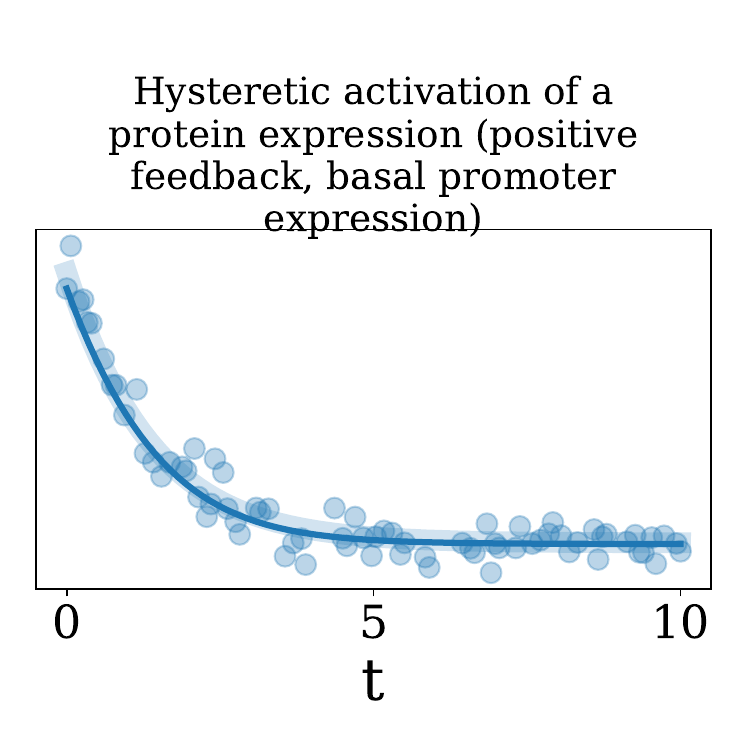}
\includegraphics[width=0.13\linewidth]{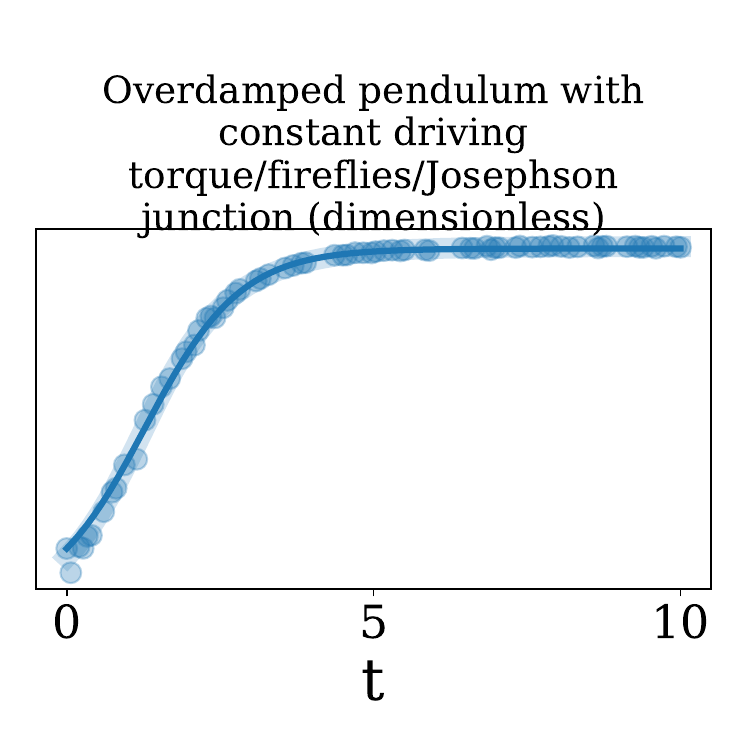}
\includegraphics[width=0.13\linewidth]{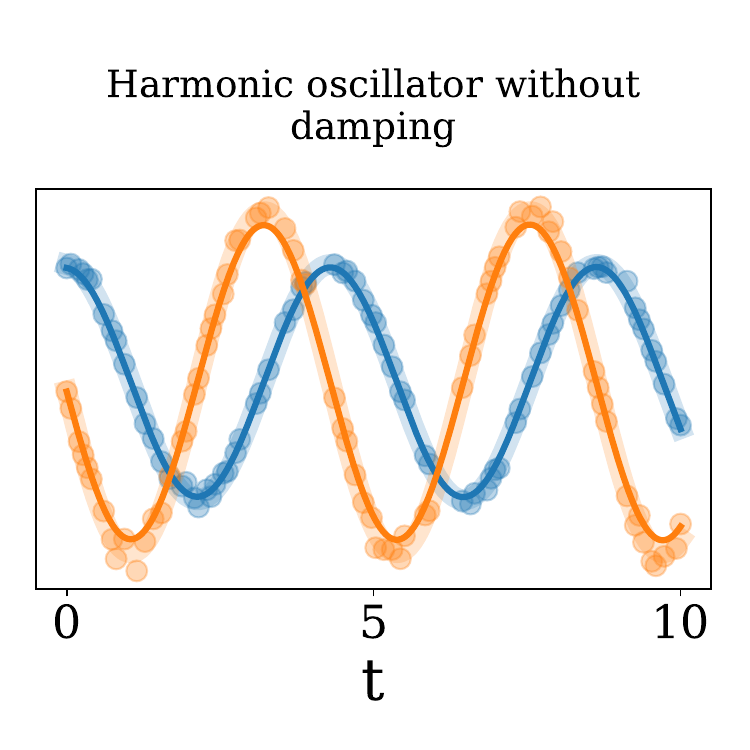}
\includegraphics[width=0.13\linewidth]{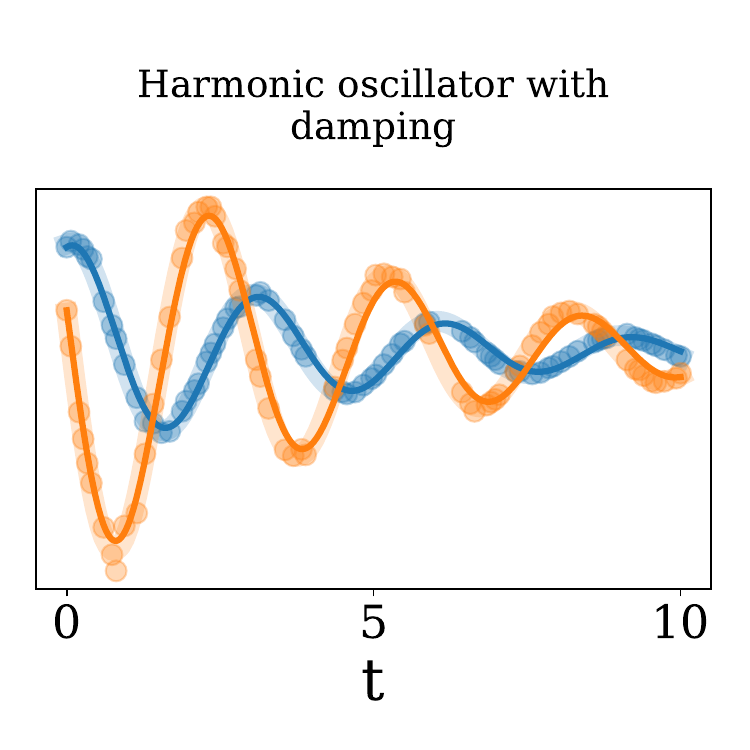}
\includegraphics[width=0.13\linewidth]{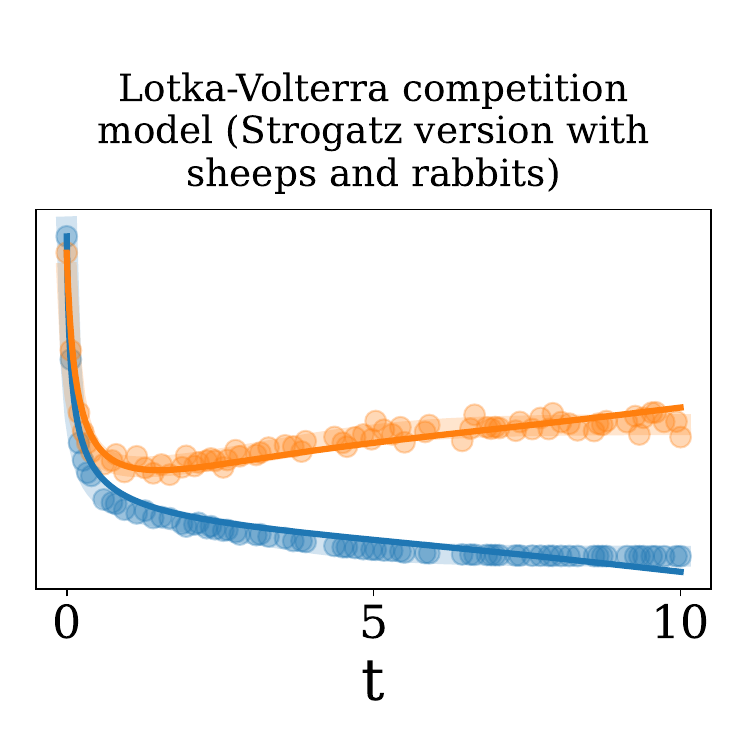}
\includegraphics[width=0.13\linewidth]{figs/odebench/odebench_27.pdf}
\includegraphics[width=0.13\linewidth]{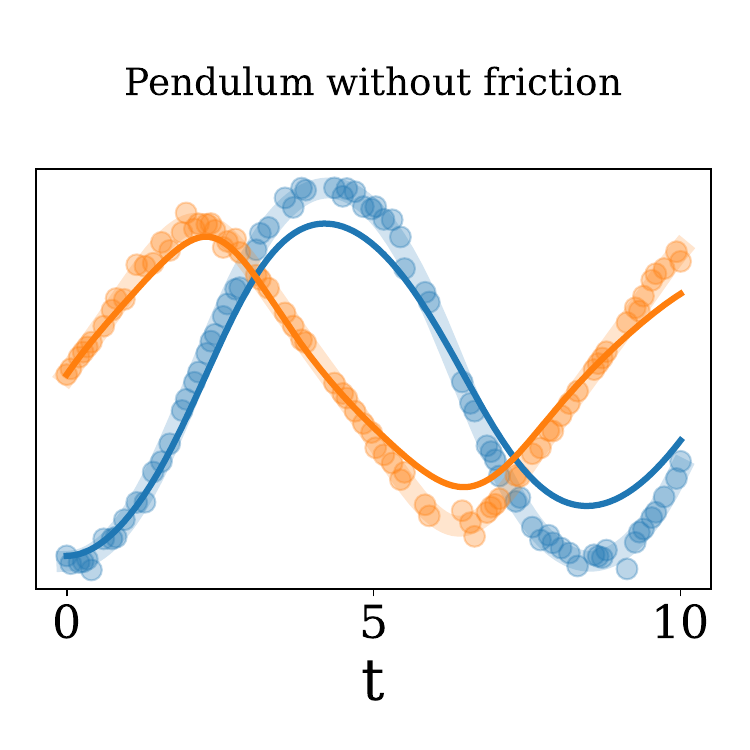}
\includegraphics[width=0.13\linewidth]{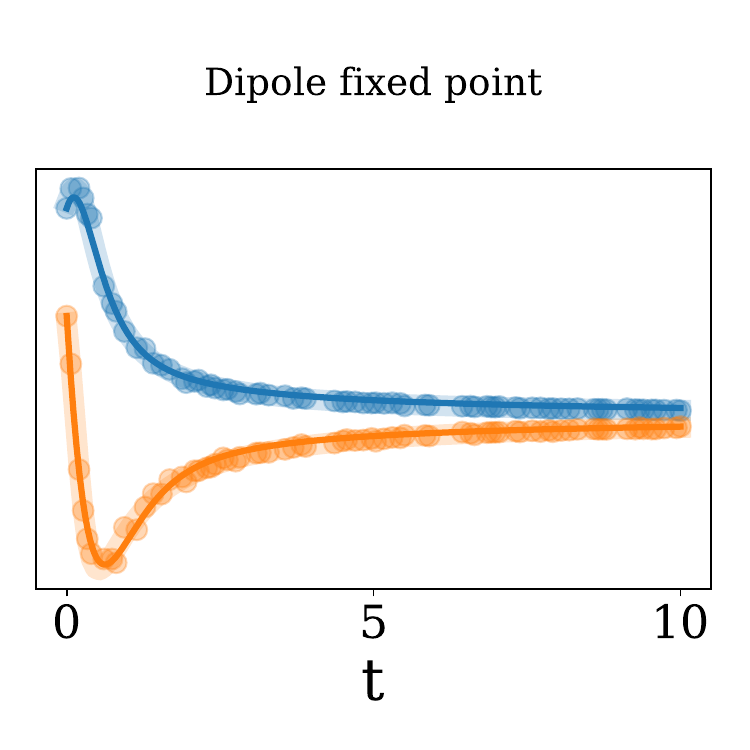}
\includegraphics[width=0.13\linewidth]{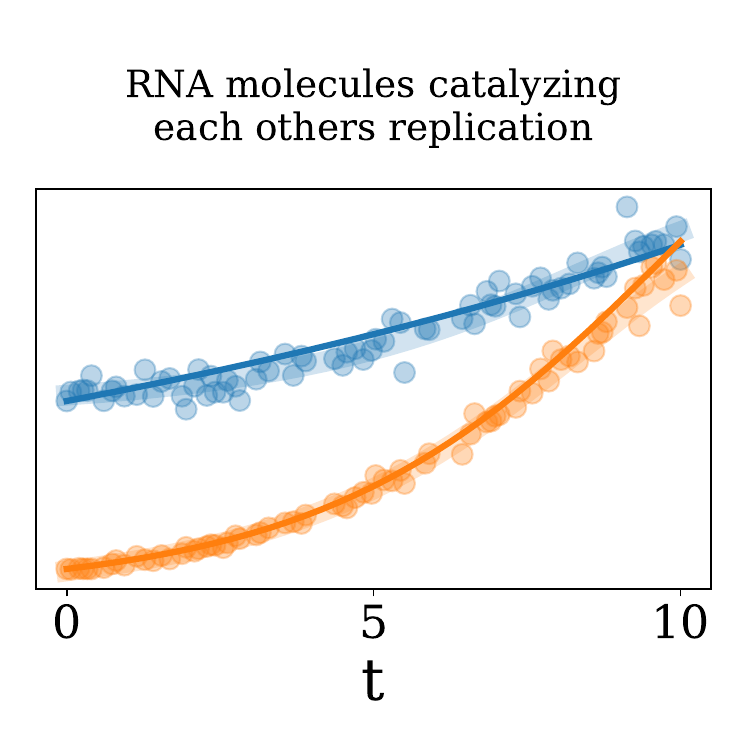}
\includegraphics[width=0.13\linewidth]{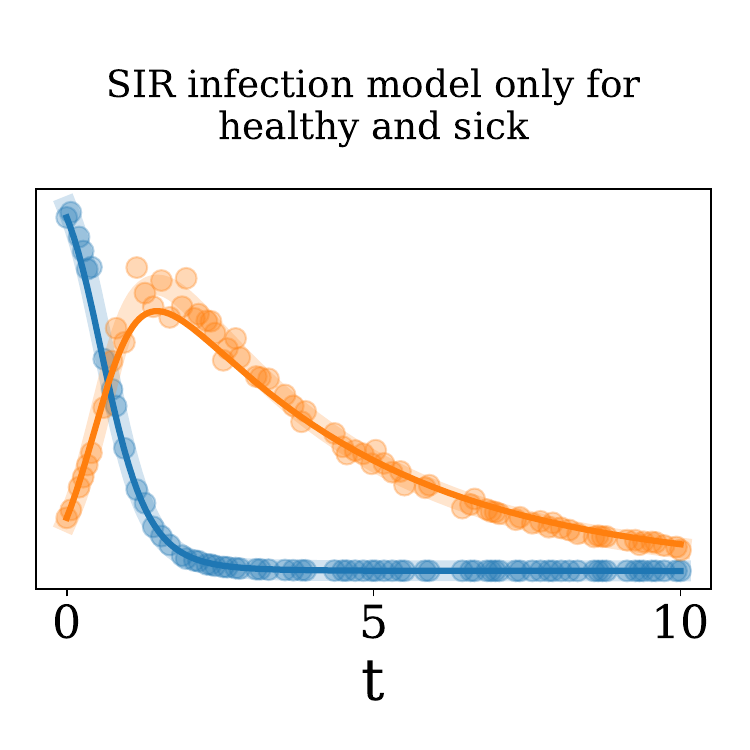}
\includegraphics[width=0.13\linewidth]{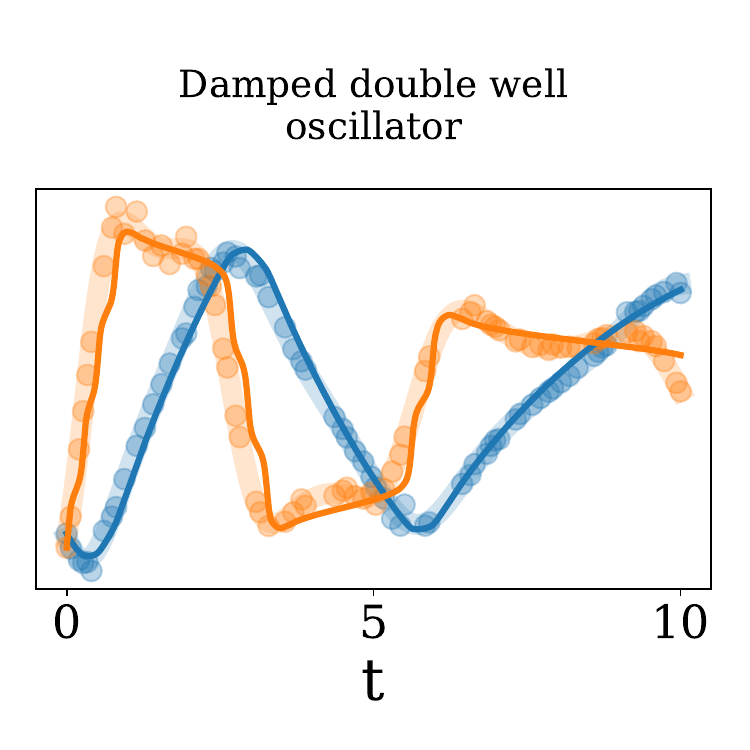}
\includegraphics[width=0.13\linewidth]{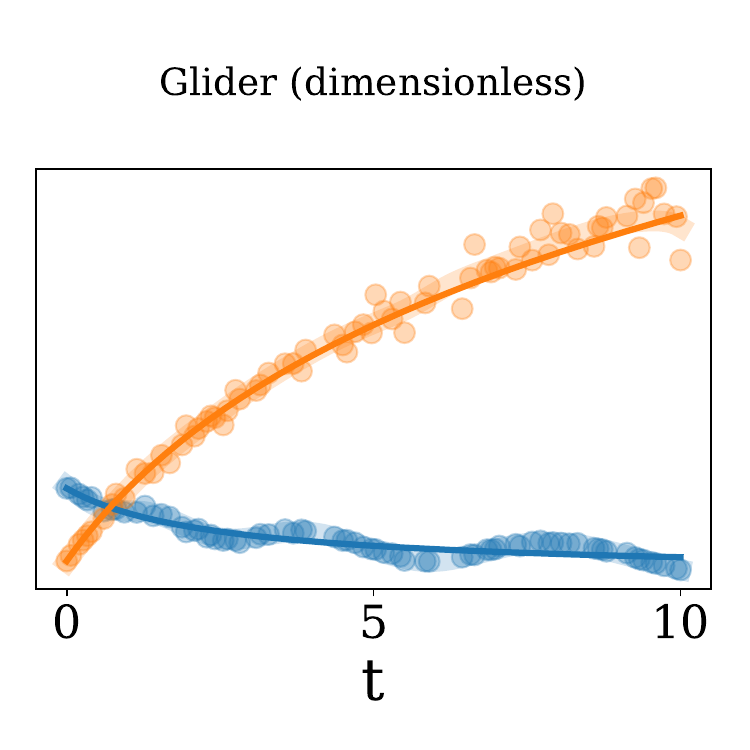}
\includegraphics[width=0.13\linewidth]{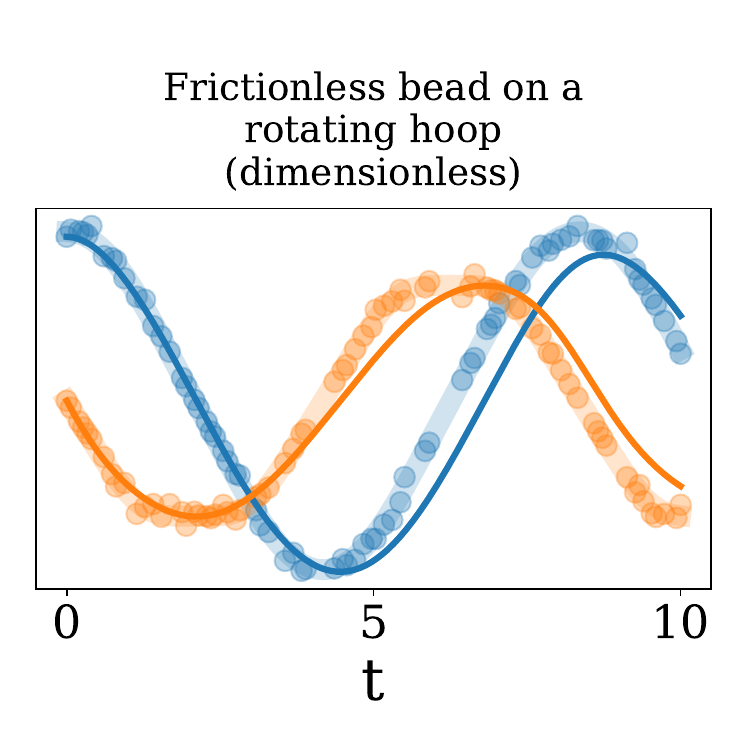}
\includegraphics[width=0.13\linewidth]{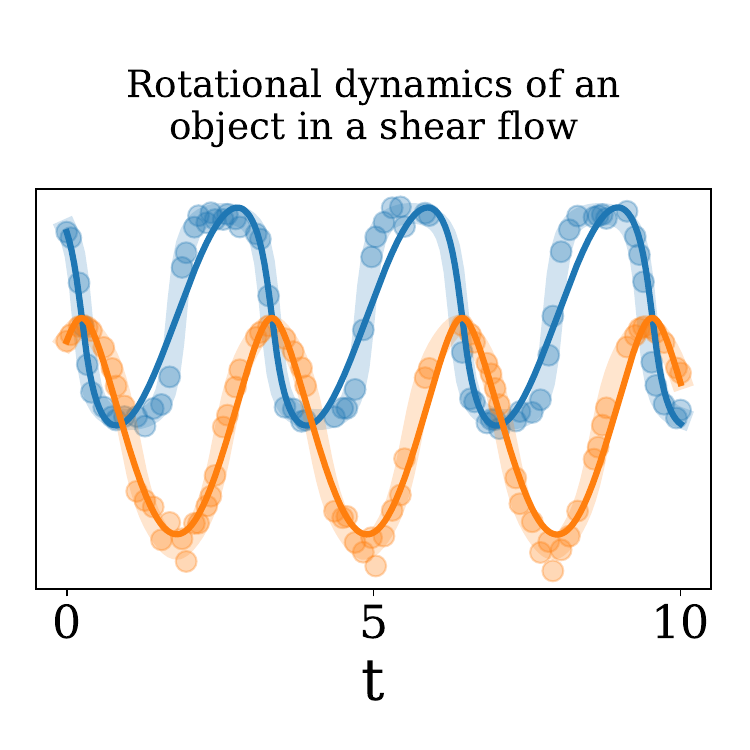}
\includegraphics[width=0.13\linewidth]{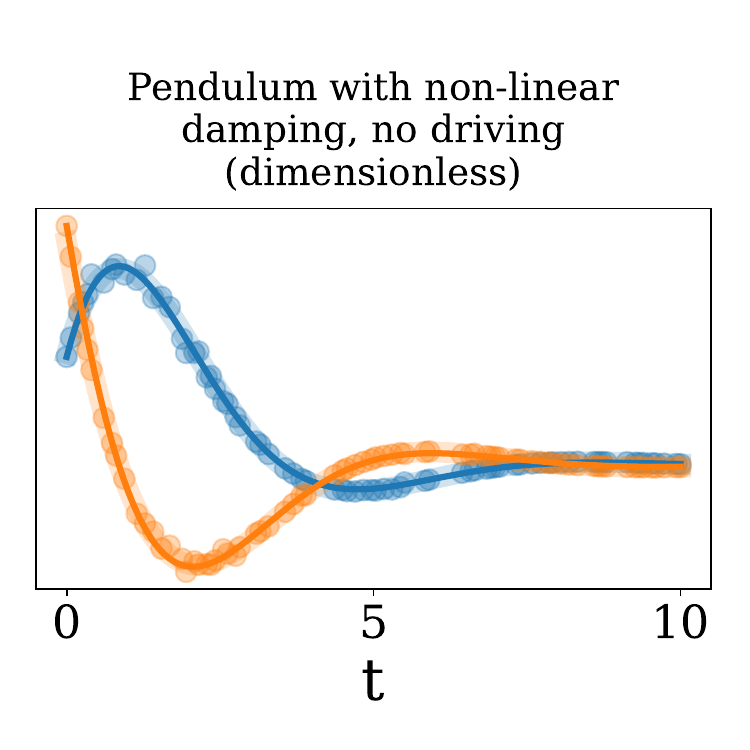}
\includegraphics[width=0.13\linewidth]{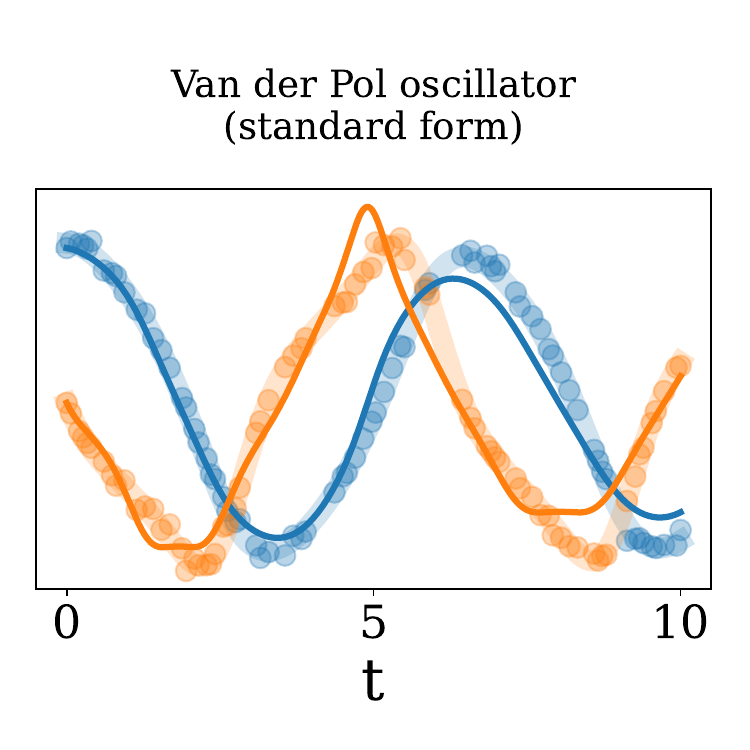}
\includegraphics[width=0.13\linewidth]{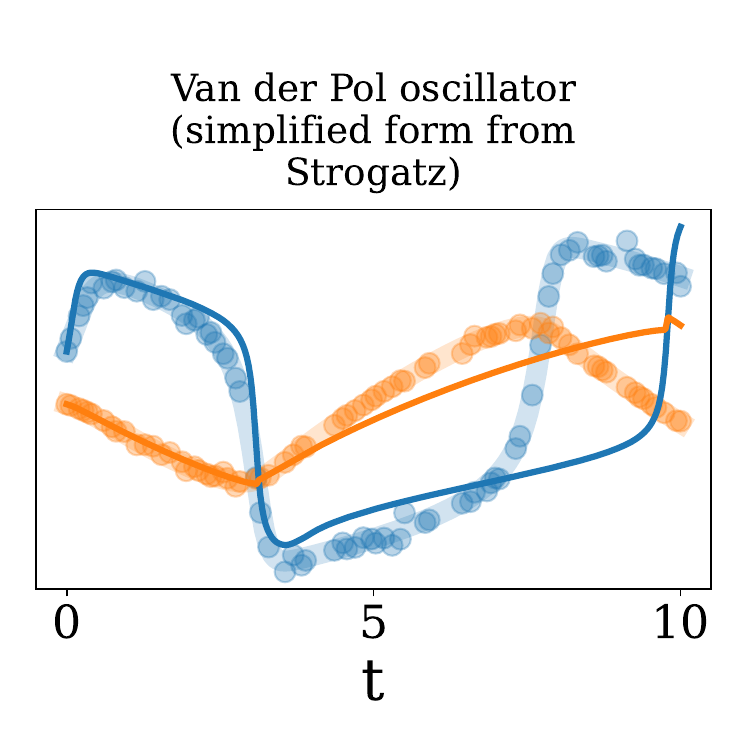}
\includegraphics[width=0.13\linewidth]{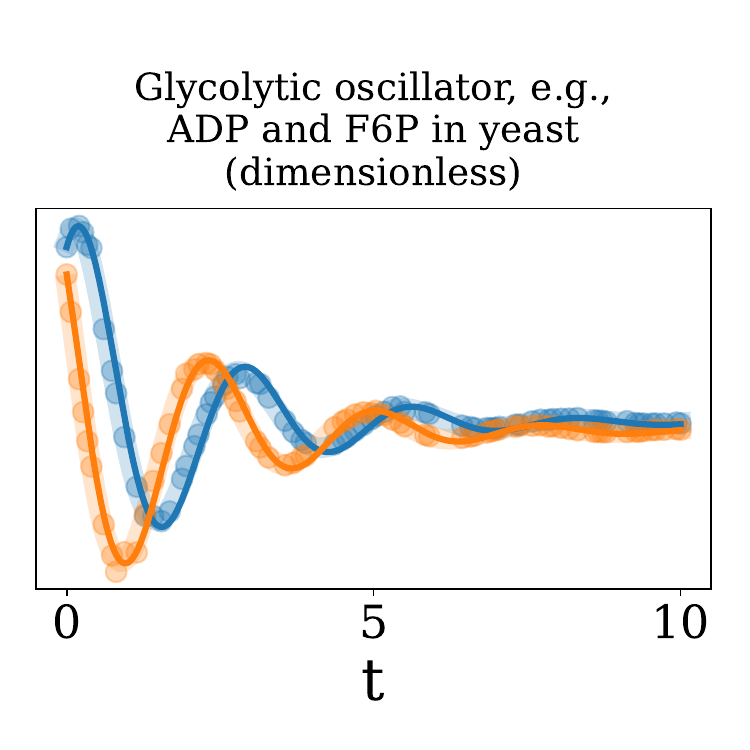}
\includegraphics[width=0.13\linewidth]{figs/odebench/odebench_40.pdf}
\includegraphics[width=0.13\linewidth]{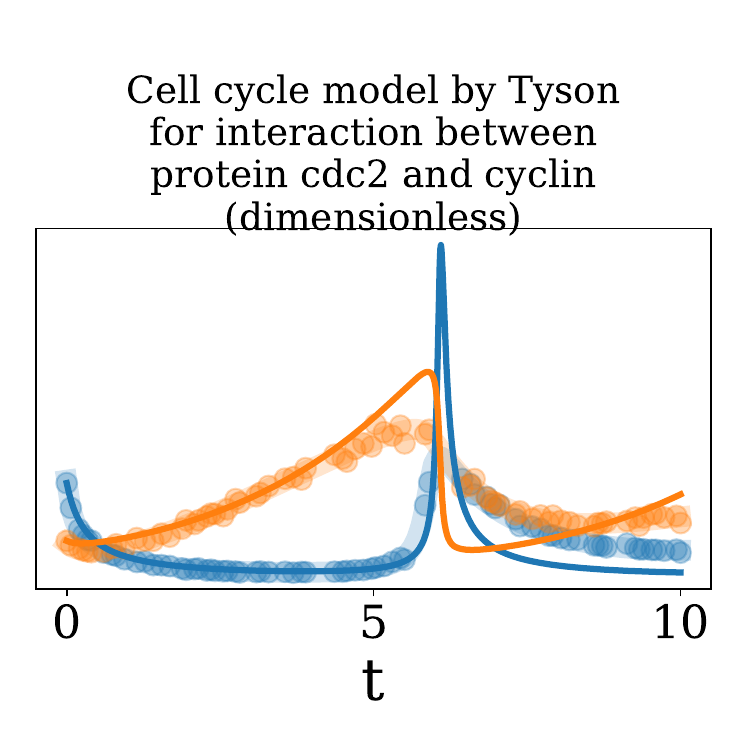}
\includegraphics[width=0.13\linewidth]{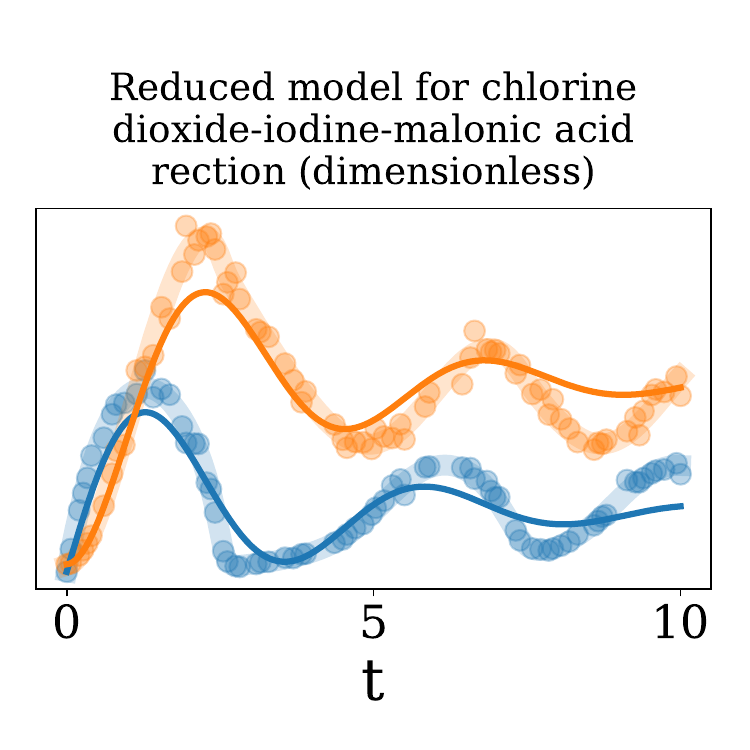}
\includegraphics[width=0.13\linewidth]{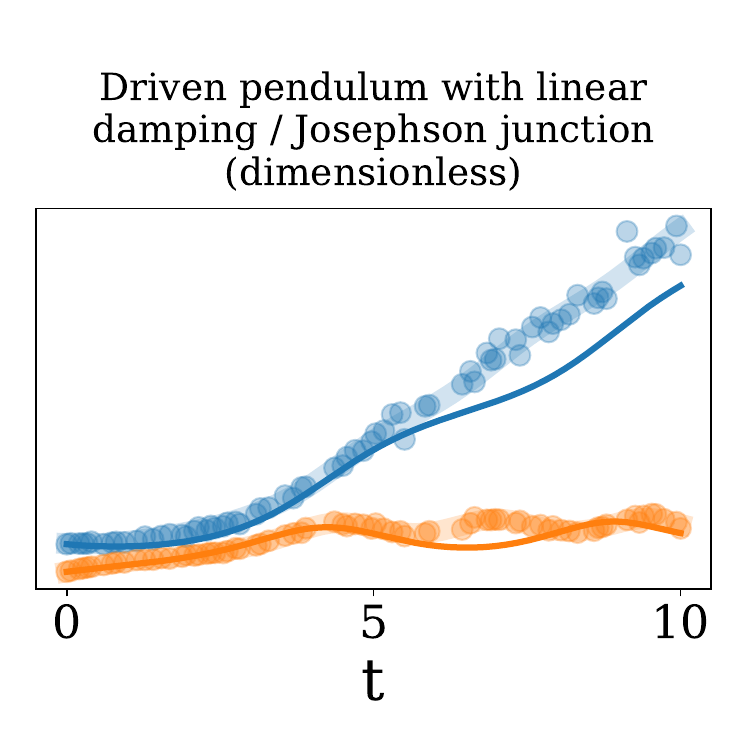}
\includegraphics[width=0.13\linewidth]{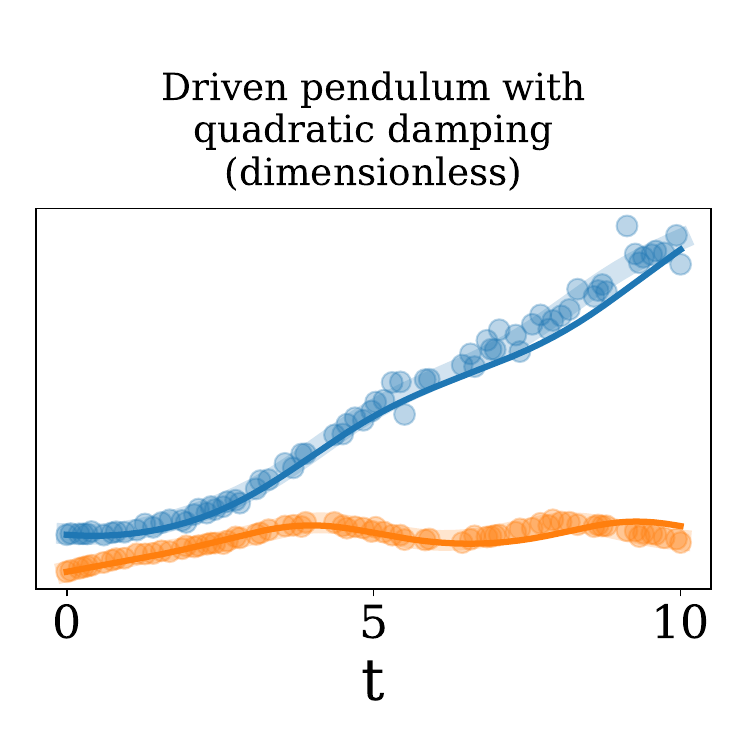}
\includegraphics[width=0.13\linewidth]{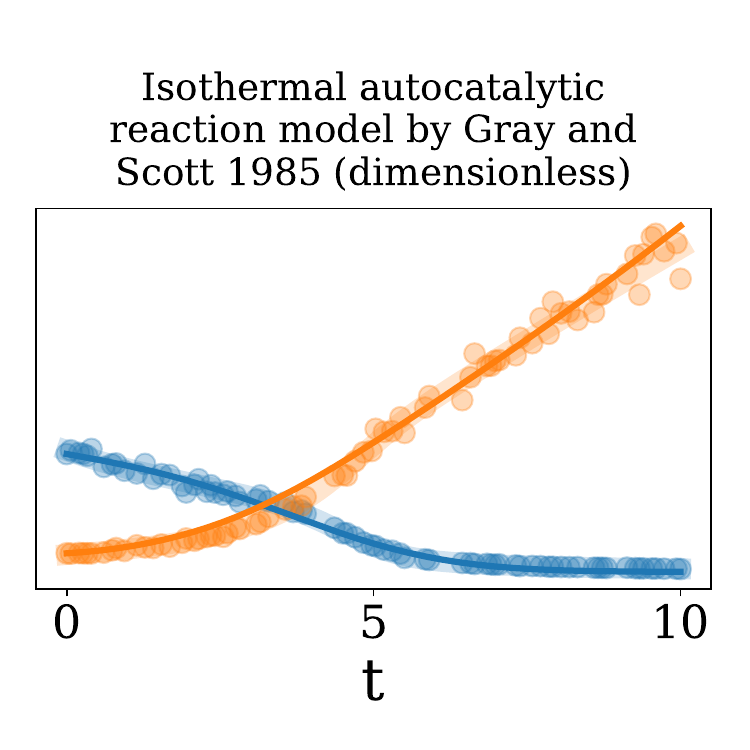}
\includegraphics[width=0.13\linewidth]{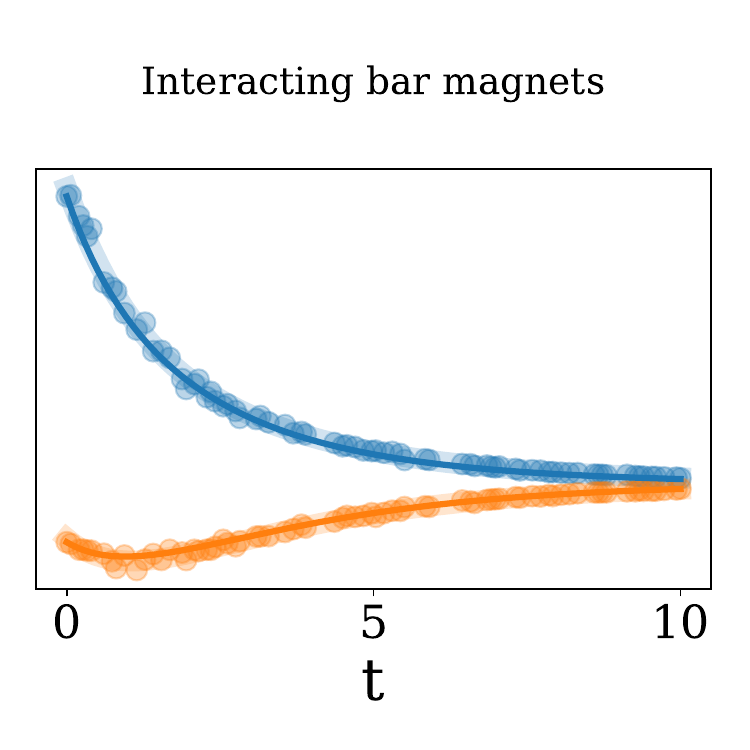}
\includegraphics[width=0.13\linewidth]{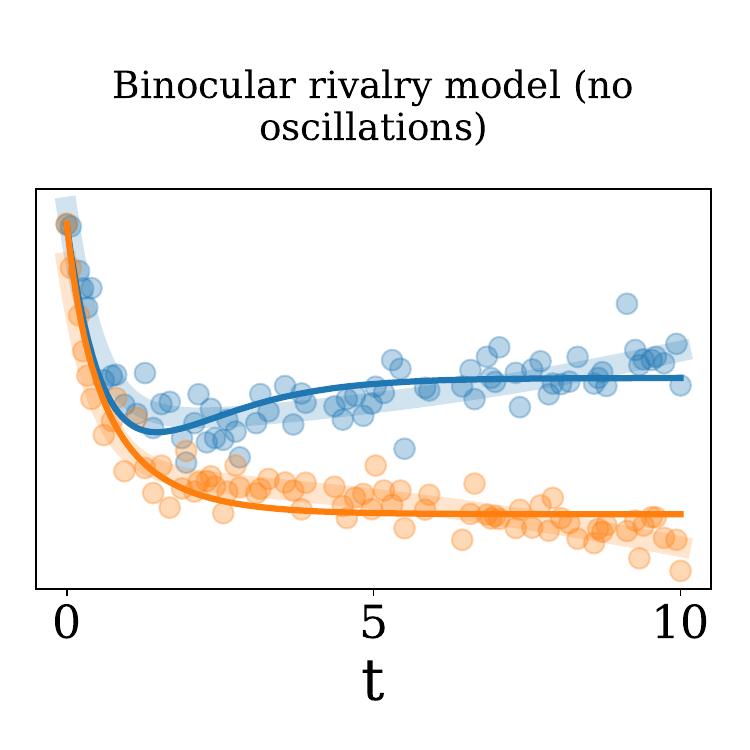}
\includegraphics[width=0.13\linewidth]{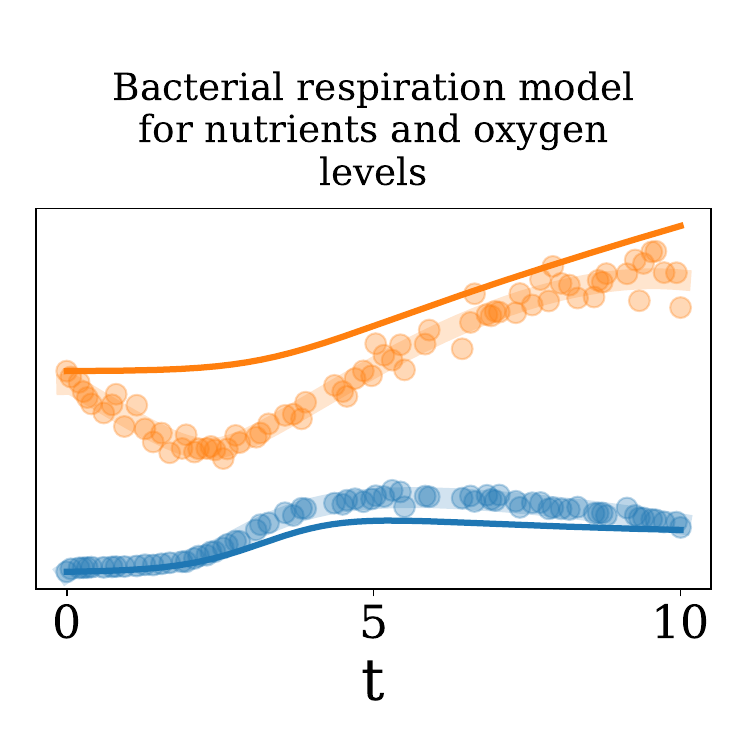}
\includegraphics[width=0.13\linewidth]{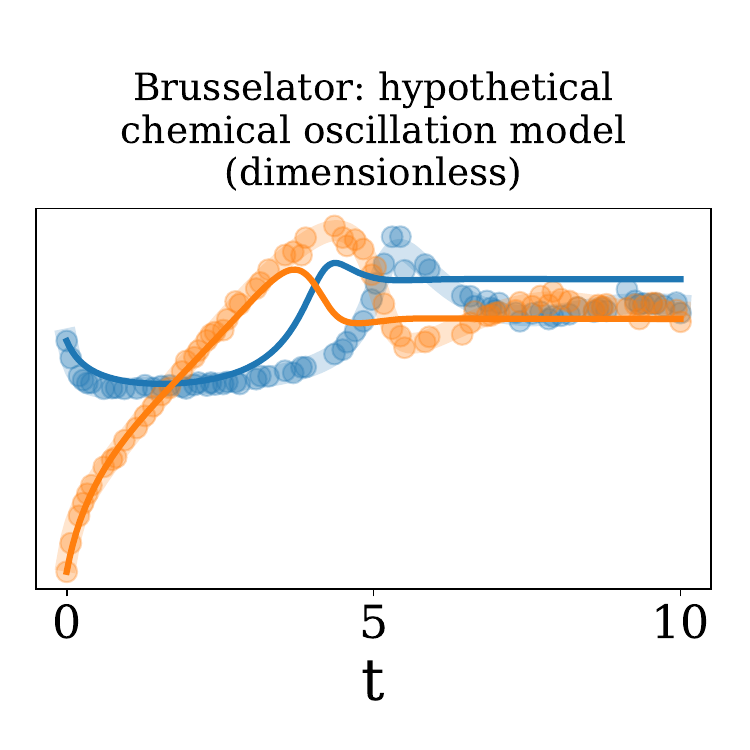}
\includegraphics[width=0.13\linewidth]{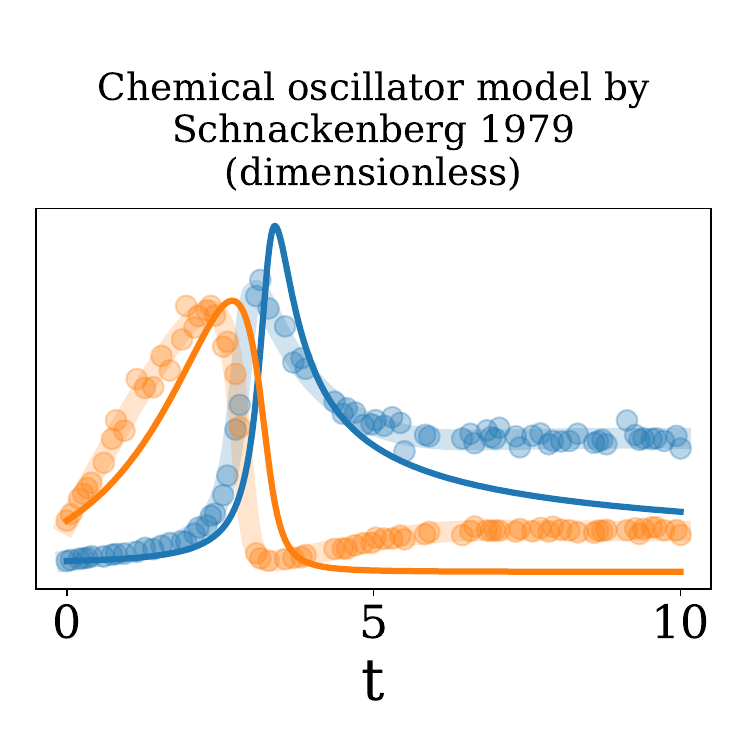}
\includegraphics[width=0.13\linewidth]{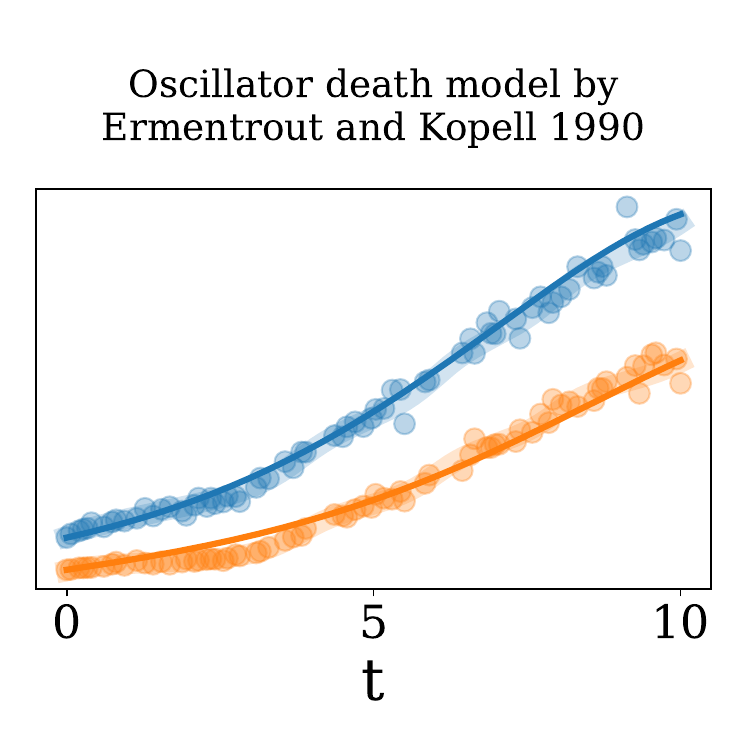}
\includegraphics[width=0.13\linewidth]{figs/odebench/odebench_52.pdf}
\includegraphics[width=0.13\linewidth]{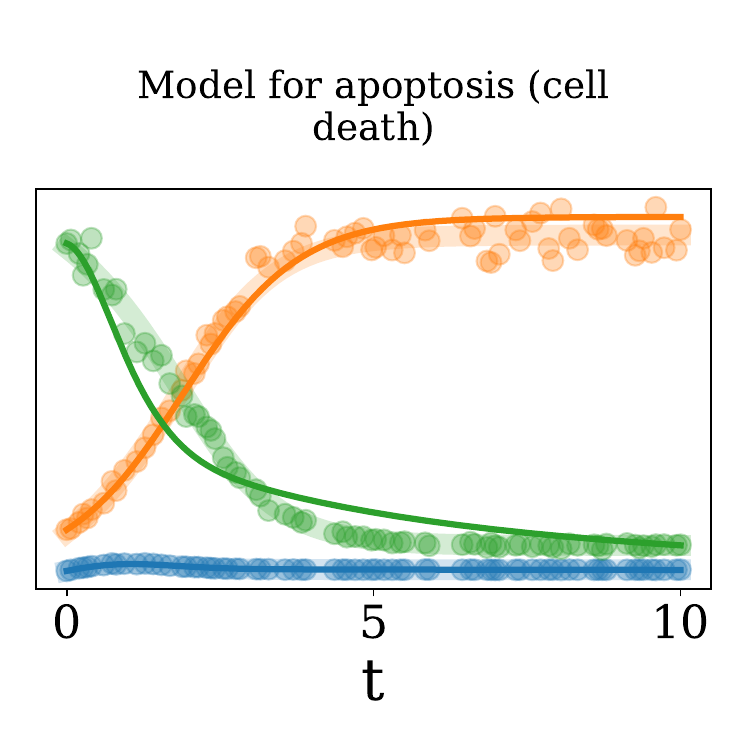}
\includegraphics[width=0.13\linewidth]{figs/odebench/odebench_54.pdf}
\includegraphics[width=0.13\linewidth]{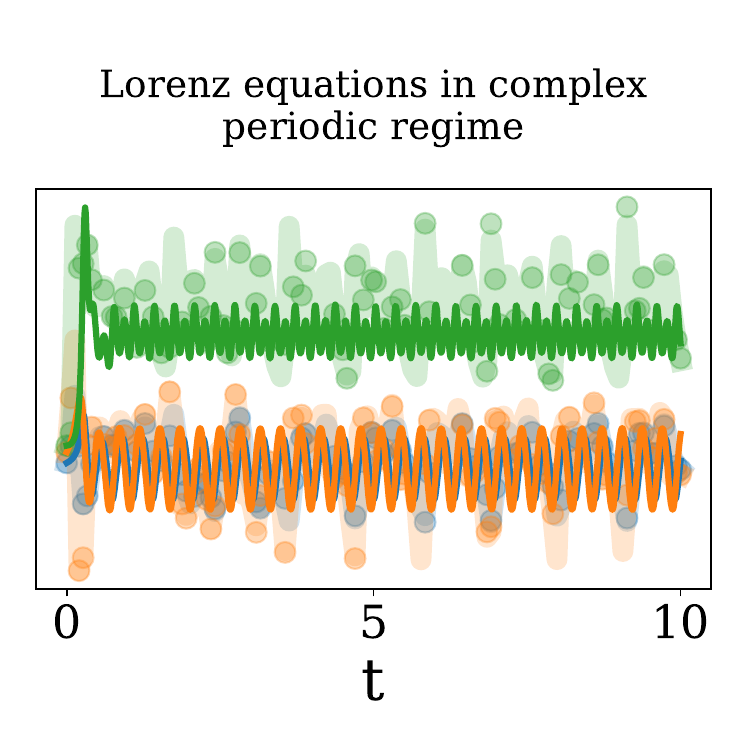}
\includegraphics[width=0.13\linewidth]{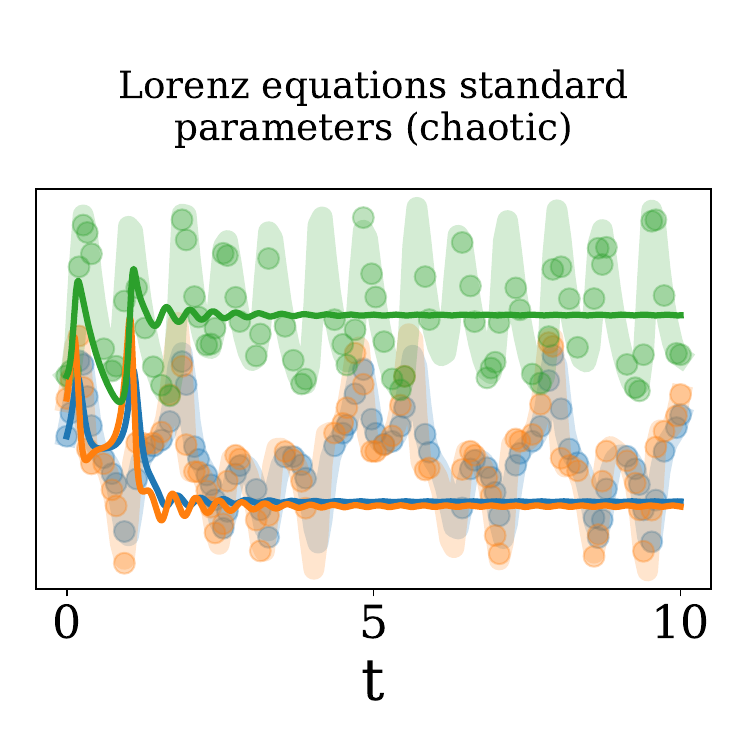}
\includegraphics[width=0.13\linewidth]{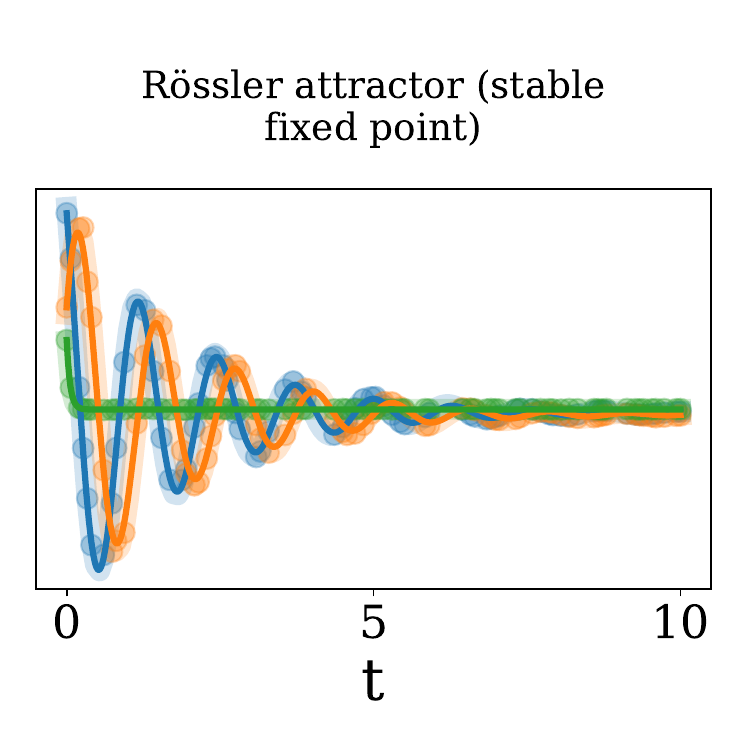}
\includegraphics[width=0.13\linewidth]{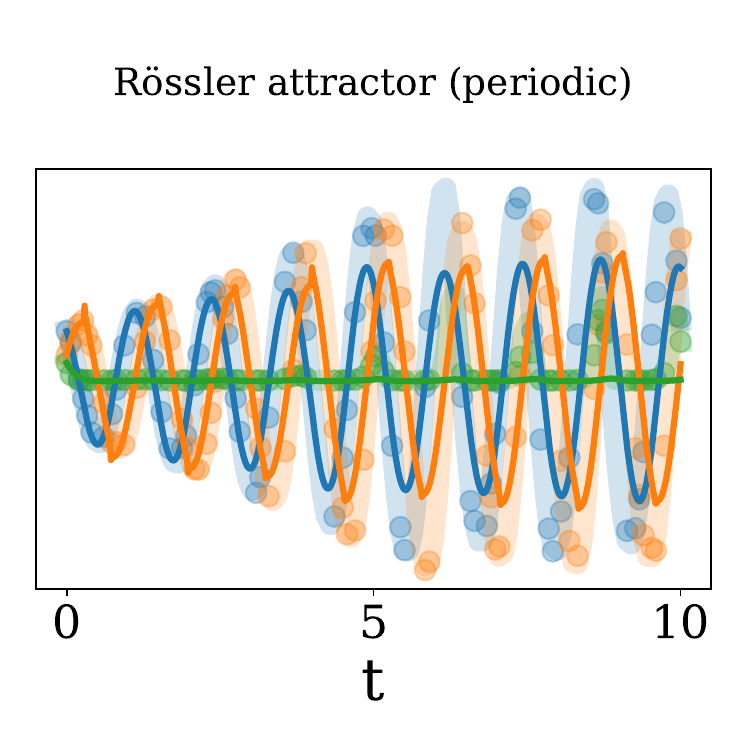}
\includegraphics[width=0.13\linewidth]{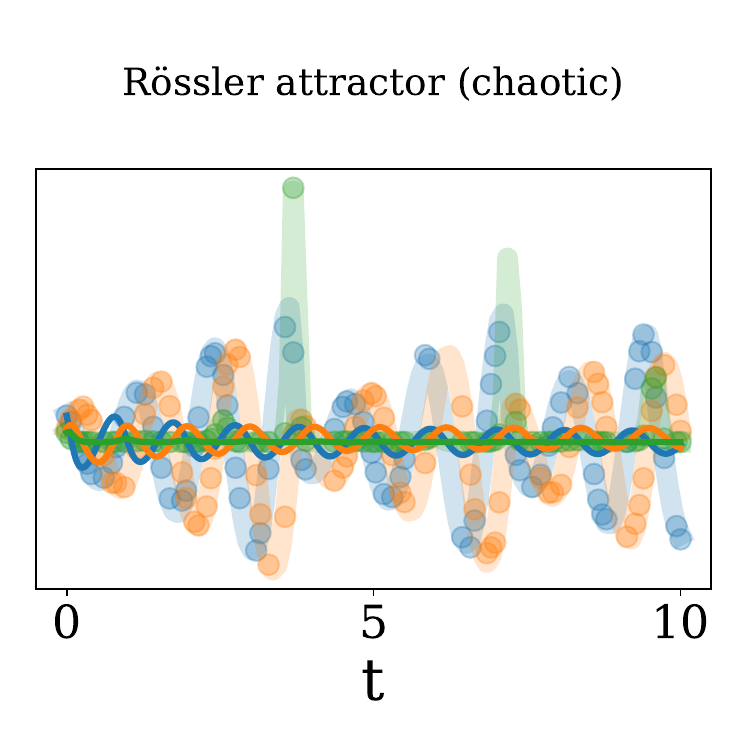}
\includegraphics[width=0.13\linewidth]{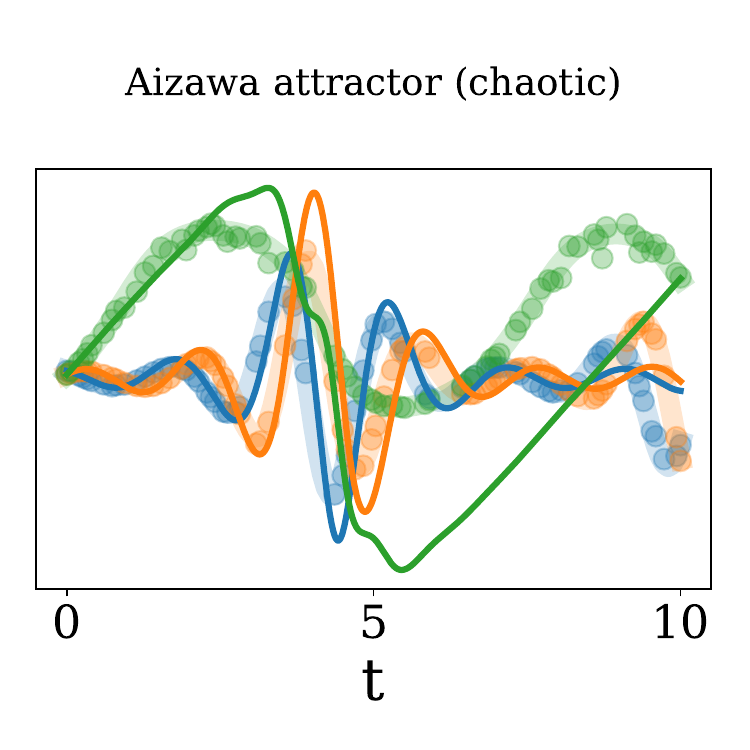}
\includegraphics[width=0.13\linewidth]{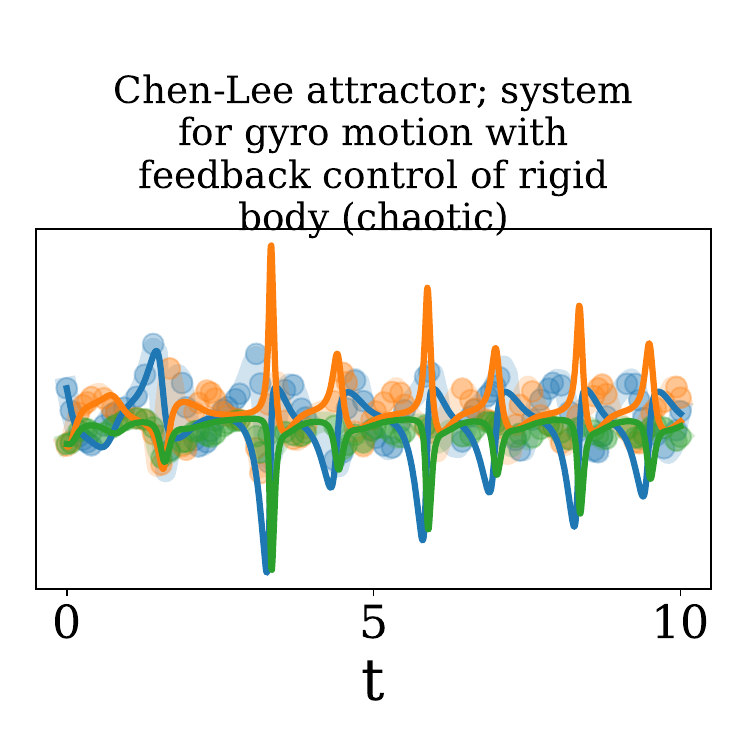}
\includegraphics[width=0.13\linewidth]{figs/odebench/odebench_62.pdf}
\includegraphics[width=0.13\linewidth]{figs/odebench/odebench_63.pdf}
    \caption{Predictions of \odeformer{} for all equations in \odebench{} for the first set of initial conditions.}
    \label{fig:odebench-odeformer}
\end{figure}

There are 23 equations in one dimension, 28 equations in two dimensions, 10 equations in three dimensions, and 2 equations in four dimensions.
This results in a total of 63 equations, 4 of which display chaotic behavior.
We provide the analytical expressions and initial conditions in \cref{tab:odebench-1d,tab:odebench-2d,tab:odebench-3d}, visualizations of a single trajectory for each ODE in \cref{fig:odebench-plots}, and \odeformer{}'s predictions for each ODE in \cref{fig:odebench-odeformer}.

\begin{table}[htb]
    \centering
    \tiny
    \begin{tabular}{c|p{3cm}|l|p{2cm}|p{2cm}}
    \toprule
        ID & System description & Equation & Parameters & Initial values\\
        \midrule
1 & RC-circuit (charging capacitor) & $\frac{c_{0} - \frac{x_{0}}{c_{1}}}{c_{2}}$ & 0.7, 1.2, 2.31 & [10.0], [3.54]\\\hline
2 & Population growth (naive) & $c_{0} x_{0}$ & 0.23 & [4.78], [0.87]\\\hline
3 & Population growth with carrying capacity & $c_{0} x_{0} \cdot \left(1 - \frac{x_{0}}{c_{1}}\right)$ & 0.79, 74.3 & [7.3], [21.0]\\\hline
4 & RC-circuit with non-linear resistor (charging capacitor) & $-0.5 + \frac{1}{e^{c_{0} - \frac{x_{0}}{c_{1}}} + 1}$ & 0.5, 0.96 & [0.8], [0.02]\\\hline
5 & Velocity of a falling object with air resistance & $c_{0} - c_{1} x_{0}^{2}$ & 9.81, 0.0021175 & [0.5], [73.0]\\\hline
6 & Autocatalysis with one fixed abundant chemical & $c_{0} x_{0} - c_{1} x_{0}^{2}$ & 2.1, 0.5 & [0.13], [2.24]\\\hline
7 & Gompertz law for tumor growth & $c_{0} x_{0} \log{\left(c_{1} x_{0} \right)}$ & 0.032, 2.29 & [1.73], [9.5]\\\hline
8 & Logistic equation with Allee effect & $c_{0} x_{0} \left(-1 + \frac{x_{0}}{c_{2}}\right) \left(1 - \frac{x_{0}}{c_{1}}\right)$ & 0.14, 130.0, 4.4 & [6.123], [2.1]\\\hline
9 & Language death model for two languages & $c_{0} \cdot \left(1 - x_{0}\right) - c_{1} x_{0}$ & 0.32, 0.28 & [0.14], [0.55]\\\hline
10 & Refined language death model for two languages & $c_{0} x_{0}^{c_{1}} \cdot \left(1 - x_{0}\right) - x_{0} \cdot \left(1 - c_{0}\right) \left(1 - x_{0}\right)^{c_{1}}$ & 0.2, 1.2 & [0.83], [0.34]\\\hline
11 & Naive critical slowing down (statistical mechanics) & $- x_{0}^{3}$ &  & [3.4], [1.6]\\\hline
12 & Photons in a laser (simple) & $c_{0} x_{0} - c_{1} x_{0}^{2}$ & 1.8, 0.1107 & [11.0], [1.3]\\\hline
13 & Overdamped bead on a rotating hoop & $c_{0} \left(c_{1} \cos{\left(x_{0} \right)} - 1\right) \sin{\left(x_{0} \right)}$ & 0.0981, 9.7 & [3.1], [2.4]\\\hline
14 & Budworm outbreak model with predation & $c_{0} x_{0} \cdot \left(1 - \frac{x_{0}}{c_{1}}\right) - \frac{c_{3} x_{0}^{2}}{c_{2}^{2} + x_{0}^{2}}$ & 0.78, 81.0, 21.2, 0.9 & [2.76], [23.3]\\\hline
15 & Budworm outbreak with predation (dimensionless) & $c_{0} x_{0} \cdot \left(1 - \frac{x_{0}}{c_{1}}\right) - \frac{x_{0}^{2}}{x_{0}^{2} + 1}$ & 0.4, 95.0 & [44.3], [4.5]\\\hline
16 & Landau equation (typical time scale tau = 1) & $c_{0} x_{0} - c_{1} x_{0}^{3} - c_{2} x_{0}^{5}$ & 0.1, -0.04, 0.001 & [0.94], [1.65]\\\hline
17 & Logistic equation with harvesting/fishing & $c_{0} x_{0} \cdot \left(1 - \frac{x_{0}}{c_{1}}\right) - c_{2}$ & 0.4, 100.0, 0.3 & [14.3], [34.2]\\\hline
18 & Improved logistic equation with harvesting/fishing & $c_{0} x_{0} \cdot \left(1 - \frac{x_{0}}{c_{1}}\right) - \frac{c_{2} x_{0}}{c_{3} + x_{0}}$ & 0.4, 100.0, 0.24, 50.0 & [21.1], [44.1]\\\hline
19 & Improved logistic equation with harvesting/fishing (dimensionless) & $- \frac{c_{0} x_{0}}{c_{1} + x_{0}} + x_{0} \cdot \left(1 - x_{0}\right)$ & 0.08, 0.8 & [0.13], [0.03]\\\hline
20 & Autocatalytic gene switching (dimensionless) & $c_{0} - c_{1} x_{0} + \frac{x_{0}^{2}}{x_{0}^{2} + 1}$ & 0.1, 0.55 & [0.002], [0.25]\\\hline
21 & Dimensionally reduced SIR infection model for dead people (dimensionless) & $c_{0} - c_{1} x_{0} - e^{- x_{0}}$ & 1.2, 0.2 & [0.0], [0.8]\\\hline
22 & Hysteretic activation of a protein expression (positive feedback, basal promoter expression) & $c_{0} + \frac{c_{1} x_{0}^{5}}{c_{2} + x_{0}^{5}} - c_{3} x_{0}$ & 1.4, 0.4, 123.0, 0.89 & [3.1], [6.3]\\\hline
23 & Overdamped pendulum with constant driving torque/fireflies/Josephson junction (dimensionless) & $c_{0} - \sin{\left(x_{0} \right)}$ & 0.21 & [-2.74], [1.65]\\

    \bottomrule
    \end{tabular}
    \caption{Scalar ODEs in ODEBench.}
    \label{tab:odebench-1d}
\end{table}
\begin{table}[htb]
    \centering
    \tiny
    \begin{tabular}{c|p{3cm}|l|p{2cm}|p{2cm}}
    \toprule
        ID & System description & Equation & Parameters & Initial values\\
        \midrule
24 & Harmonic oscillator without damping & $\begin{cases}&x_{1}\\&- c_{0} x_{0}\end{cases}$ & 2.1 & [0.4, -0.03], [0.0, 0.2]\\\hline
25 & Harmonic oscillator with damping & $\begin{cases}&x_{1}\\&- c_{0} x_{0} - c_{1} x_{1}\end{cases}$ & 4.5, 0.43 & [0.12, 0.043], [0.0, -0.3]\\\hline
26 & Lotka-Volterra competition model (Strogatz version with sheeps and rabbits) & $\begin{cases}&x_{0} \left(c_{0} - c_{1} x_{1} - x_{0}\right)\\&x_{1} \left(c_{2} - x_{0} - x_{1}\right)\end{cases}$ & 3.0, 2.0, 2.0 & [5.0, 4.3], [2.3, 3.6]\\\hline
27 & Lotka-Volterra simple (as on Wikipedia) & $\begin{cases}&x_{0} \left(c_{0} - c_{1} x_{1}\right)\\&- x_{1} \left(c_{2} - c_{3} x_{0}\right)\end{cases}$ & 1.84, 1.45, 3.0, 1.62 & [8.3, 3.4], [0.4, 0.65]\\\hline
28 & Pendulum without friction & $\begin{cases}&x_{1}\\&- c_{0} \sin{\left(x_{0} \right)}\end{cases}$ & 0.9 & [-1.9, 0.0], [0.3, 0.8]\\\hline
29 & Dipole fixed point & $\begin{cases}&c_{0} x_{0} x_{1}\\&- x_{0}^{2} + x_{1}^{2}\end{cases}$ & 0.65 & [3.2, 1.4], [1.3, 0.2]\\\hline
30 & RNA molecules catalyzing each others replication & $\begin{cases}&x_{0} \left(- c_{0} x_{0} x_{1} + x_{1}\right)\\&x_{1} \left(- c_{0} x_{0} x_{1} + x_{0}\right)\end{cases}$ & 1.61 & [0.3, 0.04], [0.1, 0.21]\\\hline
31 & SIR infection model only for healthy and sick & $\begin{cases}&- c_{0} x_{0} x_{1}\\&c_{0} x_{0} x_{1} - c_{1} x_{1}\end{cases}$ & 0.4, 0.314 & [7.2, 0.98], [20.0, 12.4]\\\hline
32 & Damped double well oscillator & $\begin{cases}&x_{1}\\&- c_{0} x_{1} - x_{0}^{3} + x_{0}\end{cases}$ & 0.18 & [-1.8, -1.8], [5.8, 0.0]\\\hline
33 & Glider (dimensionless) & $\begin{cases}&- c_{0} x_{0}^{2} - \sin{\left(x_{1} \right)}\\&x_{0} - \frac{\cos{\left(x_{1} \right)}}{x_{0}}\end{cases}$ & 0.08 & [5.0, 0.7], [9.81, -0.8]\\\hline
34 & Frictionless bead on a rotating hoop (dimensionless) & $\begin{cases}&x_{1}\\&\left(- c_{0} + \cos{\left(x_{0} \right)}\right) \sin{\left(x_{0} \right)}\end{cases}$ & 0.93 & [2.1, 0.0], [-1.2, -0.2]\\\hline
35 & Rotational dynamics of an object in a shear flow & $\begin{cases}&\cos{\left(x_{0} \right)} \cot{\left(x_{1} \right)}\\&\left(c_{0} \sin^{2}{\left(x_{1} \right)} + \cos^{2}{\left(x_{1} \right)}\right) \sin{\left(x_{0} \right)}\end{cases}$ & 4.2 & [1.13, -0.3], [2.4, 1.7]\\\hline
36 & Pendulum with non-linear damping, no driving (dimensionless) & $\begin{cases}&x_{1}\\&- c_{0} x_{1} \cos{\left(x_{0} \right)} - x_{1} - \sin{\left(x_{0} \right)}\end{cases}$ & 0.07 & [0.45, 0.9], [1.34, -0.8]\\\hline
37 & Van der Pol oscillator (standard form) & $\begin{cases}&x_{1}\\&- c_{0} x_{1} \left(x_{0}^{2} - 1\right) - x_{0}\end{cases}$ & 0.43 & [2.2, 0.0], [0.1, 3.2]\\\hline
38 & Van der Pol oscillator (simplified form from Strogatz) & $\begin{cases}&c_{0} \left(- \frac{x_{0}^{3}}{3} + x_{0} + x_{1}\right)\\&- \frac{x_{0}}{c_{0}}\end{cases}$ & 3.37 & [0.7, 0.0], [-1.1, -0.7]\\\hline
39 & Glycolytic oscillator, e.g., ADP and F6P in yeast (dimensionless) & $\begin{cases}&c_{0} x_{1} + x_{0}^{2} x_{1} - x_{0}\\&- c_{0} x_{0} + c_{1} - x_{0}^{2} x_{1}\end{cases}$ & 2.4, 0.07 & [0.4, 0.31], [0.2, -0.7]\\\hline
40 & Duffing equation (weakly non-linear oscillation) & $\begin{cases}&x_{1}\\&c_{0} x_{1} \cdot \left(1 - x_{0}^{2}\right) - x_{0}\end{cases}$ & 0.886 & [0.63, -0.03], [0.2, 0.2]\\\hline
41 & Cell cycle model by Tyson for interaction between protein cdc2 and cyclin (dimensionless) & $\begin{cases}&c_{0} \left(c_{1} + x_{0}^{2}\right) \left(- x_{0} + x_{1}\right) - x_{0}\\&c_{2} - x_{0}\end{cases}$ & 15.3, 0.001, 0.3 & [0.8, 0.3], [0.02, 1.2]\\\hline
42 & Reduced model for chlorine dioxide-iodine-malonic acid rection (dimensionless) & $\begin{cases}&c_{0} - \frac{c_{1} x_{0} x_{1}}{x_{0}^{2} + 1} - x_{0}\\&c_{2} x_{0} \left(- \frac{x_{1}}{x_{0}^{2} + 1} + 1\right)\end{cases}$ & 8.9, 4.0, 1.4 & [0.2, 0.35], [3.0, 7.8]\\\hline
43 & Driven pendulum with linear damping / Josephson junction (dimensionless) & $\begin{cases}&x_{1}\\&c_{0} - c_{1} x_{1} - \sin{\left(x_{0} \right)}\end{cases}$ & 1.67, 0.64 & [1.47, -0.2], [-1.9, 0.03]\\\hline
44 & Driven pendulum with quadratic damping (dimensionless) & $\begin{cases}&x_{1}\\&c_{0} - c_{1} x_{1} \left|{x_{1}}\right| - \sin{\left(x_{0} \right)}\end{cases}$ & 1.67, 0.64 & [1.47, -0.2], [-1.9, 0.03]\\\hline
45 & Isothermal autocatalytic reaction model by Gray and Scott 1985 (dimensionless) & $\begin{cases}&c_{0} \cdot \left(1 - x_{0}\right) - x_{0} x_{1}^{2}\\&- c_{1} x_{1} + x_{0} x_{1}^{2}\end{cases}$ & 0.5, 0.02 & [1.4, 0.2], [0.32, 0.64]\\\hline
46 & Interacting bar magnets & $\begin{cases}&c_{0} \sin{\left(x_{0} - x_{1} \right)} - \sin{\left(x_{0} \right)}\\&- c_{0} \sin{\left(x_{0} - x_{1} \right)} - \sin{\left(x_{1} \right)}\end{cases}$ & 0.33 & [0.54, -0.1], [0.43, 1.21]\\\hline
47 & Binocular rivalry model (no oscillations) & $\begin{cases}&- x_{0} + \frac{1}{e^{c_{0} x_{1} - c_{1}} + 1}\\&- x_{1} + \frac{1}{e^{c_{0} x_{0} - c_{1}} + 1}\end{cases}$ & 4.89, 1.4 & [0.65, 0.59], [3.2, 10.3]\\\hline
48 & Bacterial respiration model for nutrients and oxygen levels & $\begin{cases}&c_{0} - \frac{x_{0} x_{1}}{c_{1} x_{0}^{2} + 1} - x_{0}\\&c_{2} - \frac{x_{0} x_{1}}{c_{1} x_{0}^{2} + 1}\end{cases}$ & 18.3, 0.48, 11.23 & [0.1, 30.4], [13.2, 5.21]\\\hline
49 & Brusselator: hypothetical chemical oscillation model (dimensionless) & $\begin{cases}&c_{1} x_{0}^{2} x_{1} - x_{0} \left(c_{0} + 1\right) + 1\\&c_{0} x_{0} - c_{1} x_{0}^{2} x_{1}\end{cases}$ & 3.03, 3.1 & [0.7, -1.4], [2.1, 1.3]\\\hline
50 & Chemical oscillator model by Schnackenberg 1979 (dimensionless) & $\begin{cases}&c_{0} + x_{0}^{2} x_{1} - x_{0}\\&c_{1} - x_{0}^{2} x_{1}\end{cases}$ & 0.24, 1.43 & [0.14, 0.6], [1.5, 0.9]\\\hline
51 & Oscillator death model by Ermentrout and Kopell 1990 & $\begin{cases}&c_{0} + \sin{\left(x_{1} \right)} \cos{\left(x_{0} \right)}\\&c_{1} + \sin{\left(x_{1} \right)} \cos{\left(x_{0} \right)}\end{cases}$ & 1.432, 0.972 & [2.2, 0.67], [0.03, -0.12]\\
    \bottomrule
    \end{tabular}
    \caption{2 dimensional ODEs in ODEBench.}
    \label{tab:odebench-2d}
\end{table}
\begin{table}[htb]
    \centering
    \tiny
    \begin{tabular}{c|p{3cm}|l|p{2cm}|p{2cm}}
    \toprule
        ID & System description & Equation & Parameters & Initial values\\
        \midrule
52 & Maxwell-Bloch equations (laser dynamics) & $\begin{cases}&c_{0} \left(- x_{0} + x_{1}\right)\\&c_{1} \left(x_{0} x_{2} - x_{1}\right)\\&c_{2} \left(- c_{3} x_{0} x_{1} + c_{3} - x_{2} + 1\right)\end{cases}$ & 0.1, 0.21, 0.34, 3.1 & [1.3, 1.1, 0.89], [0.89, 1.3, 1.1]\\\hline
53 & Model for apoptosis (cell death) & $\begin{cases}&c_{0} - c_{4} x_{0} - \frac{c_{5} x_{0} x_{1}}{c_{9} + x_{0}}\\&c_{1} x_{2} \left(c_{8} + x_{1}\right) - \frac{c_{2} x_{1}}{c_{6} + x_{1}} - \frac{c_{3} x_{0} x_{1}}{c_{7} + x_{1}}\\&- c_{1} x_{2} \left(c_{8} + x_{1}\right) + \frac{c_{2} x_{1}}{c_{6} + x_{1}} + \frac{c_{3} x_{0} x_{1}}{c_{7} + x_{1}}\end{cases}$ & 0.1, 0.6, 0.2, 7.95, 0.05, 0.4, 0.1, 2.0, 0.1, 0.1 & [0.005, 0.26, 2.15], [0.248, 0.0973, 0.0027]\\\hline
54 & Lorenz equations in well-behaved periodic regime & $\begin{cases}&c_{0} \left(- x_{0} + x_{1}\right)\\&c_{1} x_{0} - x_{0} x_{2} - x_{1}\\&- c_{2} x_{2} + x_{0} x_{1}\end{cases}$ & 5.1, 12.0, 1.67 & [2.3, 8.1, 12.4], [10.0, 20.0, 30.0]\\\hline
55 & Lorenz equations in complex periodic regime & $\begin{cases}&c_{0} \left(- x_{0} + x_{1}\right)\\&c_{1} x_{0} - x_{0} x_{2} - x_{1}\\&- c_{2} x_{2} + x_{0} x_{1}\end{cases}$ & 10.0, 99.96, 8/3 & [2.3, 8.1, 12.4], [10.0, 20.0, 30.0]\\\hline
56 & Lorenz equations standard parameters (chaotic) & $\begin{cases}&c_{0} \left(- x_{0} + x_{1}\right)\\&c_{1} x_{0} - x_{0} x_{2} - x_{1}\\&- c_{2} x_{2} + x_{0} x_{1}\end{cases}$ & 10.0, 28.0, 8/3 & [2.3, 8.1, 12.4], [10.0, 20.0, 30.0]\\\hline
57 & Rössler attractor (stable fixed point) & $\begin{cases}&c_{3} \left(- x_{1} - x_{2}\right)\\&c_{3} \left(c_{0} x_{1} + x_{0}\right)\\&c_{3} \left(c_{1} + x_{2} \left(- c_{2} + x_{0}\right)\right)\end{cases}$ & -0.2, 0.2, 5.7, 5.0 & [2.3, 1.1, 0.8], [-0.1, 4.1, -2.1]\\\hline
58 & Rössler attractor (periodic) & $\begin{cases}&c_{3} \left(- x_{1} - x_{2}\right)\\&c_{3} \left(c_{0} x_{1} + x_{0}\right)\\&c_{3} \left(c_{1} + x_{2} \left(- c_{2} + x_{0}\right)\right)\end{cases}$ & 0.1, 0.2, 5.7, 5.0 & [2.3, 1.1, 0.8], [-0.1, 4.1, -2.1]\\\hline
59 & Rössler attractor (chaotic) & $\begin{cases}&c_{3} \left(- x_{1} - x_{2}\right)\\&c_{3} \left(c_{0} x_{1} + x_{0}\right)\\&c_{3} \left(c_{1} + x_{2} \left(- c_{2} + x_{0}\right)\right)\end{cases}$ & 0.2, 0.2, 5.7, 5.0 & [2.3, 1.1, 0.8], [-0.1, 4.1, -2.1]\\\hline
60 & Aizawa attractor (chaotic) & $\begin{cases}&- c_{3} x_{1} + x_{0} \left(- c_{1} + x_{2}\right)\\&c_{3} x_{0} + x_{1} \left(- c_{1} + x_{2}\right)\\&c_{0} x_{2} + c_{2} + c_{5} x_{0}^{3} x_{2} - 1/3 x_{2}^{3} - \left(x_{0}^{2} + x_{1}^{2}\right) \left(c_{4} x_{2} + 1\right)\end{cases}$ & 0.95, 0.7, 0.65, 3.5, 0.25, 0.1 & [0.1, 0.05, 0.05], [-0.3, 0.2, 0.1]\\\hline
61 & Chen-Lee attractor; system for gyro motion with feedback control of rigid body (chaotic) & $\begin{cases}&c_{0} x_{0} - x_{1} x_{2}\\&c_{1} x_{1} + x_{0} x_{2}\\&c_{2} x_{2} + \frac{x_{0} x_{1}}{c_{3}}\end{cases}$ & 5.0, -10.0, -3.8, 3.0 & [15, -15, -15], [8, 14, -10]\\\hline
62 & Binocular rivalry model with adaptation (oscillations) & $\begin{cases}&- x_{0} + \frac{1}{e^{c_{0} x_{2} + c_{1} x_{1} - c_{2}} + 1}\\&c_{3} \left(x_{0} - x_{1}\right)\\&- x_{2} + \frac{1}{e^{c_{0} x_{0} + c_{1} x_{3} - c_{2}} + 1}\\&c_{3} \left(x_{2} - x_{3}\right)\end{cases}$ & 0.89, 0.4, 1.4, 1.0 & [2.25, -0.5, -1.13, 0.4], [0.342, -0.431, -0.86, 0.041]\\\hline
63 & SEIR infection model (proportions) & $\begin{cases}&- c_{1} x_{0} x_{2}\\&- c_{0} x_{1} + c_{1} x_{0} x_{2}\\&c_{0} x_{1} - c_{2} x_{2}\\&c_{2} x_{2}\end{cases}$ & 0.47, 0.28, 0.3 & [0.6, 0.3, 0.09, 0.01], [0.4, 0.3, 0.25, 0.05]\\
    \bottomrule
    \end{tabular}
    \caption{3 and 4 dimensional ODEs in ODEBench.}
    \label{tab:odebench-3d}
\end{table}

\clearpage


\clearpage

\clearpage

\section{Identifiability}\label{app:identifiability}

Traditionally, when inferring dynamical laws as ODEs from observations, one assumed a parametric form of the ODE to be known.
In addition, typical assumptions included that the state of the system may not be fully observed, and that one may not be interested in identifying the system fully (all parameters), but that there are only certain (combinations of) parameters that need to be identified.
For example, in a typical epidemiological model of disease spread, e.g., modelling the fractions of susceptible, infected, and recovered people (SIR model) in a population, one may only be able to measure the number of infected people, but may also only be interested in estimating the reproduction rate (a ratio of two parameters in the full equations).
Hence a typical identifiability question would start from a known system, a known observation function, and a specific target quantity.
Researchers have then developed various general methods and procedures to decide whether such a query is solvable, i.e., whether the target is identified from the observations within the assumptions   \citep{aastrom1971system,miao2011identifiability,villaverde2016structural,hamelin2020observability}.

The study of identifiability from observations within larger non-parametric function classes from full state observations has only been taken up more recently.
For example, linear autonomous systems as well as autonomous systems that are linear in parameters are fairly well understood \citep{stanhope2014identifiability,duan2020identification} with \citet{qiu2022identifiability} recently essentially providing closure to this question.
A broad summary of these findings is that linear (in parameters) autonomous systems are almost always (almost surely) identifiable from a single trajectory for many reasonable measures (probability distributions) over parameters, i.e., over ODE systems.

When it comes to non-parametric classes such as analytic, algebraic, continuous, or smooth functions, \citet{scholl2023uniqueness} have recently presented the first detailed analysis and results for a broad class of scalar PDEs.
These include ODEs as special cases, however their results only apply to scalar ODEs.
Even though they consider ODEs beyond first order, this still does not include multivariate ODE systems.
In the scalar case, identification from a single solution trajectory is possible for analytic functions $f$, but essentially impossible for continuous (or smooth) $f$ \citep{scholl2023uniqueness}.\footnote{The intuition here is that linear, polynomial, or more broadly even real analytic scalar univariate functions can be uniquely extrapolated to all of $\R$ when they are known on any open interval. On the other hand, there are infinitely many ways to extrapolate continuous or smooth functions beyond any interval.}

The function class considered by \odeformer{}, i.e., the distribution implied by our generator, predominantly contains real analytic functions, but not exclusively since $x \to 1/x$ is analytic only on $\R \setminus \{0\}$.
Hence, it is not clear in which category we fall regarding identifiability.

Crucially, all these theoretical results assume the entire continuous, non-noisy solution trajectory to be known.
Little is known for discrete and noisy observations, where identifiability likely turns from a yes/no question into one of probabilistic claims given a prior over functions and the assumed noise model.
Hence, current theoretical results do not conclude whether non-linear ODEs can practically be inferred from data.

\section{Training details for \odeformer{}}\label{app:training_details}

We optimize a cross-entropy loss with the Adam optimizer (with default parameters suggested by~\cite{kingma2014adam}), with a learning rate warming up from $10^{-7}$ to $2\times 10^{-4}$ across the initial 10,000 steps, by decaying via the cosine schedule for the next 300,000 steps. The annealing cycle then restarts with a damping factor of 3/2 as per~\cite{kingma2014adam}, resulting in approximately 800,000 optimization steps. We do not use any regularization such as weight decay or dropout. To efficiently manage the greatly varying input sequence lengths, we group examples of similar lengths in batches, with the constraint that each batch contains 10,000 tokens. Our model is trained on a set of about 50M examples pre-generated with 80 CPU cores. When run on a single NVIDIA A100 GPU with 80GB memory and 8 CPU cores, \odeformer{}’s training process takes roughly three days.

\section{Parameter optimization in \odeformer{} (opt)}\label{app:optional_parameter_optimization}

In contrast to all baseline models, \odeformer{} is a pretrained model and predicted ODEs are not explicitly fit to the data observed at inference time. However, similar to \cite{kamienny2022end} we can post-hoc optimize the parameters of a predicted ODE to improve the data fit. Although parameter estimation for dynamical system is known to be a challenging inference problem, we use the Broyden-Fletcher-Goldfarb-Shanno algorithm (BFGS) \citep{NoceWrig06bfgs} as implemented in \texttt{scipy.optimize.minimize} \citep{virtanen2020scipy} and thus opt for a comparatively simple local, gradient-based method in the hope that the parameter values predicted by \odeformer{} only need slight refinement. The optimizer solves the following problem
\begin{align*}
    \argmin_{\{p_1, \ldots, p_k\}} \text{loss}\left[(x(t_0), \ldots, x(t_n)), \mathtt{ solve\_ivp}(\hat{f}(x; \{p_1, \ldots, p_k\}), x_0=x(t_0), t=(t_0, \ldots, t_n))\right]
\end{align*}
where $\{p_1, \ldots, p_k\}$ denotes the set of parameters of the ODE $\hat{f}$ that was predicted by \odeformer{}, and where $(x(t_0) \ldots, x(t_n))$ represents the (potentially noisy) observations. We use the negative variance-weighted $R^2$ score as optimization loss.

\section{Evaluation of baseline models}
\label{app:baselines}

\begin{table}[htb]
    \caption{Hyperparameter names and values for optimization of baseline models. For FFX and PySR, we optimize over finite difference order and smoother window length but no additional hyper parameters.}
    \begin{center}
    \begin{tabular}{c|l|c}
    \toprule
        Model & Hyperparameter & Values \\ 
        \hline
        \multirow{2}{*}{\bf All models} & finite difference order & 2, 3, 4 \\
                   & smoother window length & None, 15\\
        \hline
        \multirow{3}{*}{\bf AFP} & population size & 100, 500, 1000\\
        & generations & 2500, 500, 250\\
        & operators & \makecell{[n, v, +, -, *, /, exp, log, 2, 3, sqrt], \\\ [n, v, +, -, *, /,  exp, log, 2, 3, sqrt, sin, cos]}\\
        \hline
        \multirow{3}{*}{\bf EHC} & population size & 100, 500, 1000\\
        & generations & 1000, 200, 100\\
        & operators & \makecell{[n, v, +, -, *, /, exp, log, 2, 3, sqrt], \\\ [n, v, +, -, *, /,  exp, log, 2, 3, sqrt, sin, cos]}\\
        \hline
        \multirow{3}{*}{\bf EPLEX} & population size & 100, 500, 1000\\
        & generations & 2500, 500, 250\\
        & operators & \makecell{[n, v, +, -, *, /, exp, log, 2, 3, sqrt], \\\ [n, v, +, -, *, /,  exp, log, 2, 3, sqrt, sin, cos]}\\
        \hline
        \multirow{3}{*}{\bf FE-AFP} & population size & 100, 500, 1000\\
        & generations & 2500, 500, 250\\
        & operators & \makecell{[n, v, +, -, *, /, exp, log, 2, 3, sqrt], \\\ [n, v, +, -, *, /,  exp, log, 2, 3, sqrt, sin, cos]}\\
        \hline
        \bf ProGED & grammar & \makecell{universal, rational \\ simplerational, trigonometric \\ polynomial}\\
        \hline
        \multirow{5}{*}{\bf SINDy} & polynomial degree & 1, 2, 3, 4, 5, 6, 7, 8, 9, 10\\
                    & basis functions   & \makecell{ [polynomials], \\\ [polynomials, sin, cos, exp], \\\ [polynomials, sin, cos, exp, log, sqrt, 1/x] }  \\
                    & optimizer threshold & 0.05, 0.1, 0.15 \\
                    & optimizer alpha & 0.025, 0.05, 0.075 \\
                    & optimizer max iterations & 20, 100 \\
                    \hline
        \multirow{5}{*}{\bf SINDy (esc)}  & polynomial degree & 1, 2, 3, 4, 5, 6, 7, 8, 9, 10 \\
                    & basis functions & \makecell{ [polynomials], \\\ [polynomials, sin, cos, exp]}  \\
                    & optimizer threshold & 0.05, 0.1, 0.15 \\
                    & optimizer alpha & 0.025, 0.05, 0.075 \\
                    & optimizer max iterations & 20, 100 \\
                    \hline
        \multirow{5}{*}{\bf SINDy (poly)}  & polynomial degree & 1, 2, 3, 4, 5, 6, 7, 8, 9, 10 \\
                    & optimizer threshold & 0.05, 0.1, 0.15 \\
                    & optimizer alpha & 0.025, 0.05, 0.075 \\
                    & optimizer max iterations & 20, 100 \\
    \bottomrule
    \end{tabular}
    \end{center}
    \label{tab:baseline_hyperparams}
\end{table}

\paragraph{Hyperparameter optimization.}
All baseline models are fitted to each trajectory separately and each fit involves a separate hyperparameter optimization. Hyperparameters that are searched over are listed in \cref{tab:baseline_hyperparams}, all other hyperparameters are set to their respective default values. For each combination of hyperparameters, the model is fitted on the first 70\% and scored on the remaining 30\% of a trajectory. To reduce runtime, we parallelize optimization according to \texttt{GridSearchCV} from \texttt{scikit-learn} \citep{pedregosa2011scikit} and set the number of parallel jobs to $\min($\# combinations, \# cpu cores (= 48)$)$. After selecting the combination with highest $R^2$ score, the final model is fitted on the full trajectory.

\paragraph{Finite difference approximations.}
Except for ProGED, all baseline models require approximations of temporal derivatives of all state variables of an ODE system as regression targets. To estimate temporal derivatives we use the central finite difference algorithm as implemented by \texttt{FiniteDifference} in the \texttt{pysindy} software package \citep{pysindy} and include the approximation order in the hyperparameter search. For a fair comparison on noisy trajectories we extend the hyperparameter search to also include optional smoothing of trajectories with a Savitzky-Savgol filter with a window length of 15 as implemented by \texttt{SmoothedFiniteDifference} \citep{pysindy}.

\paragraph{Vector-valued functions.}
Some of the baseline implementations (AFP, FE-AFP, EPLEX, EHC, FFX) do not readily support vector-valued functions ($f:\mathbb R^D\to \mathbb R^D$) but only scalar-valued functions ($f:\mathbb R^D\to \mathbb R$). To evaluate these baselines on systems of ODEs, we run them separately for each component $f_i:\mathbb R^D\to \mathbb R$ of the system and combine the predictions for all components $i \in \{1, \ldots, D\}$ via the Cartesian product $\{f_1^1, \ldots, f_1^{K_1} \} \times \ldots \times \{f_D^1, \ldots, f_D^{K_D} \}$ where $K_i$ represents the number of predictions, e.g., the length of the Pareto front, obtained for component $i$.

\paragraph{Candidate selection.}
In symbolic regression, one typically faces a trade-off between accuracy (how well the function/trajectory is recovered) and complexity of the proposed expression.
There are different strategies in the literature to select a single, final equation from the accuracy-complexity Pareto front, which may bias comparisons across methods along one or the other dimension.
For a fair comparison, we evaluate all equations of a model's Pareto front and pick the final equation based on accuracy.

\clearpage
\section{Effect of beam size}\label{app:beamsize}

\begin{figure}[htb]
    \begin{minipage}{.53\linewidth}
        \includegraphics[width=\linewidth]{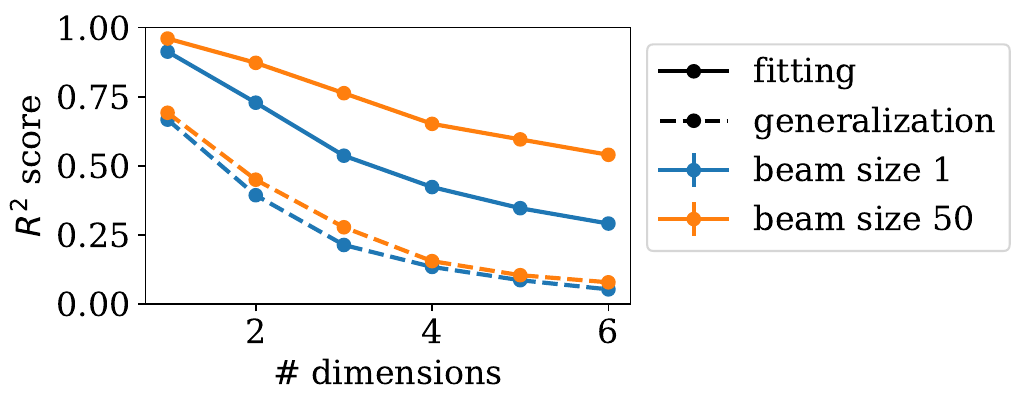}
    \end{minipage}
    \hfill
    \begin{minipage}{.45\linewidth}
        \caption{\textbf{Increasing the beam size improves reconstruction, but not generalization.} We plot the average reconstruction and generalization $R^2$-score on 10,000 noise-free, densely samples synthetic examples for various beam sizes and a temperature of 0.1.}
        \label{fig:beam}
    \end{minipage}
\end{figure}

In \cref{fig:beam}, we study the impact of the beam size on reconstruction and generalization performance.
While reconstruction improves with the beam size, generalization hardly changes. This highlights the importance of using both metrics: the two are not necessarily correlated, and the latter is a much better proxy of symbolic recovery than the former.

\section{Additional results on benchmarks}\label{app:results}

\begin{figure}[htb]
    \begin{subfigure}[b]{.33\linewidth}
        \includegraphics[width=\linewidth]{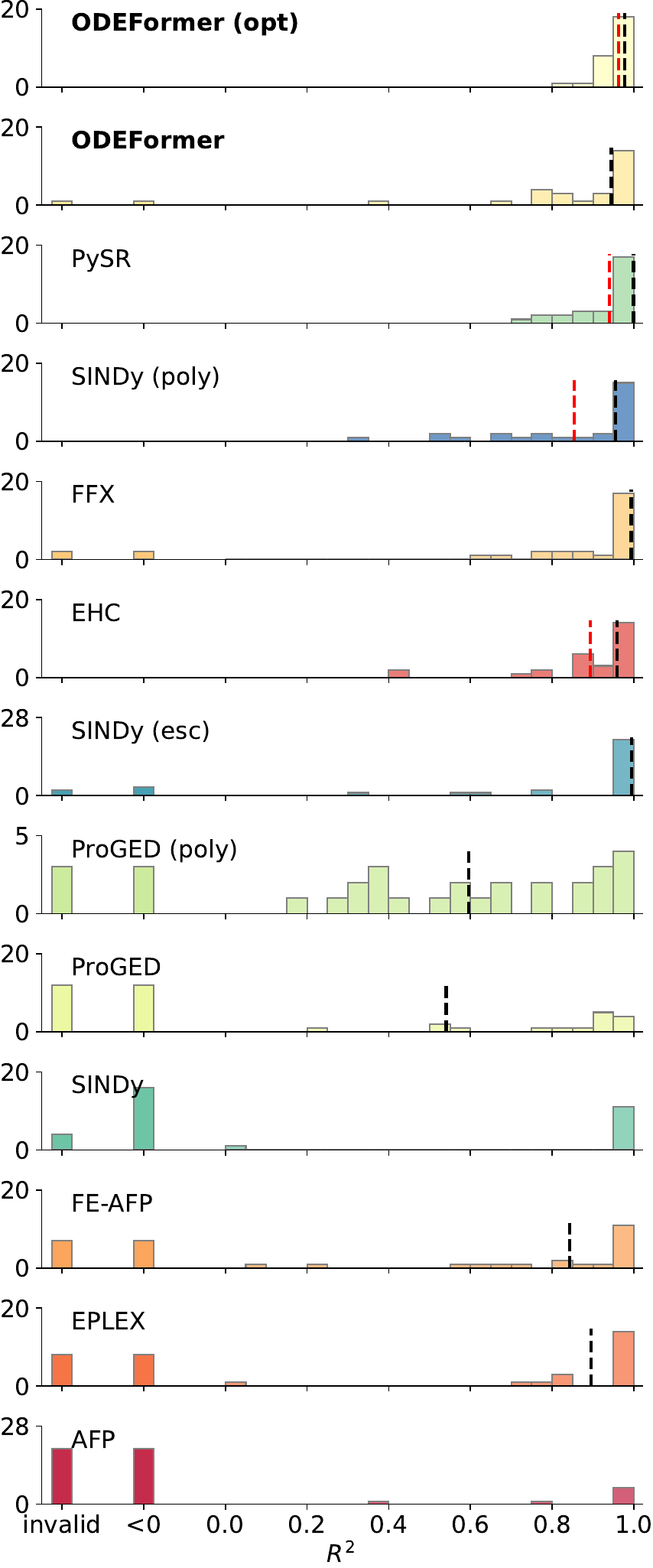}
    \caption[]{$\sigma$=0}
    \end{subfigure}
    \begin{subfigure}[b]{.33\linewidth}
        \includegraphics[width=\linewidth]{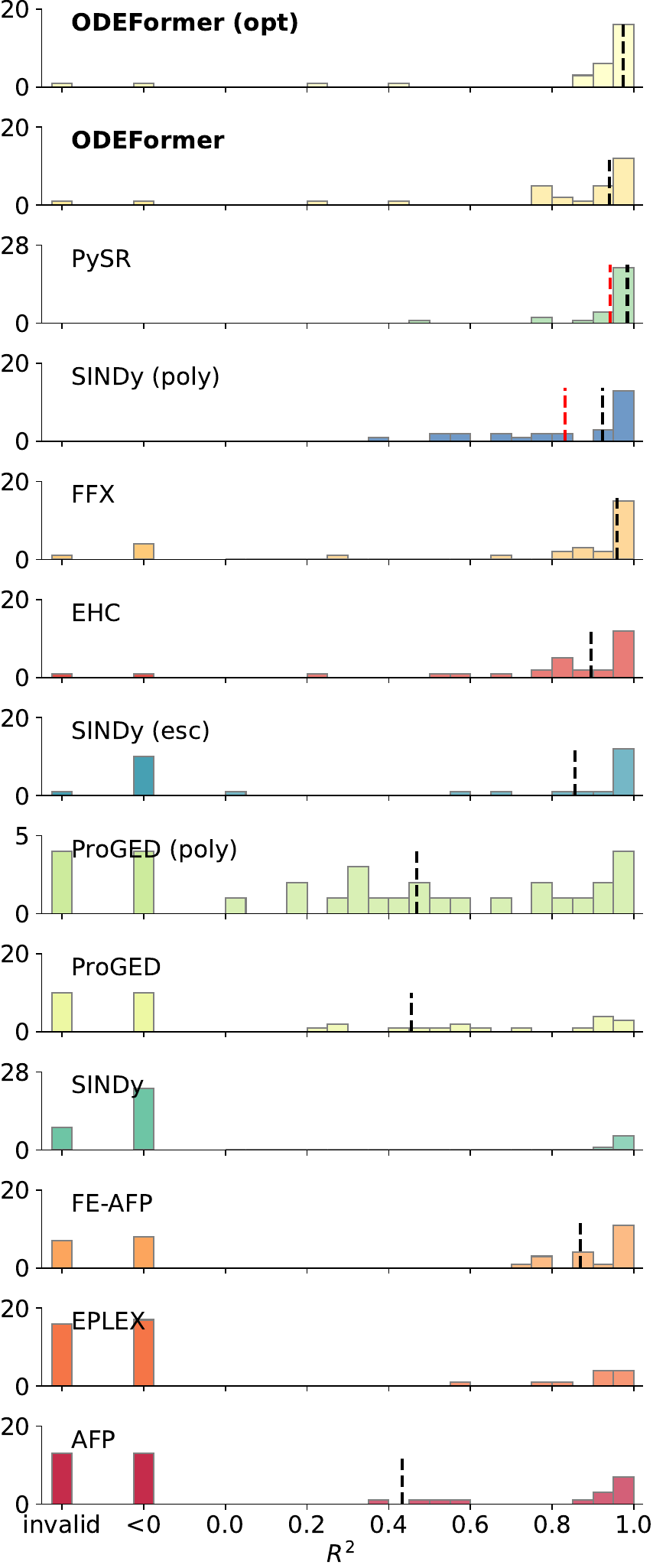}
    \caption[]{$\sigma$=0.01}
    \end{subfigure}
    \begin{subfigure}[b]{.33\linewidth}
        \includegraphics[width=\linewidth]{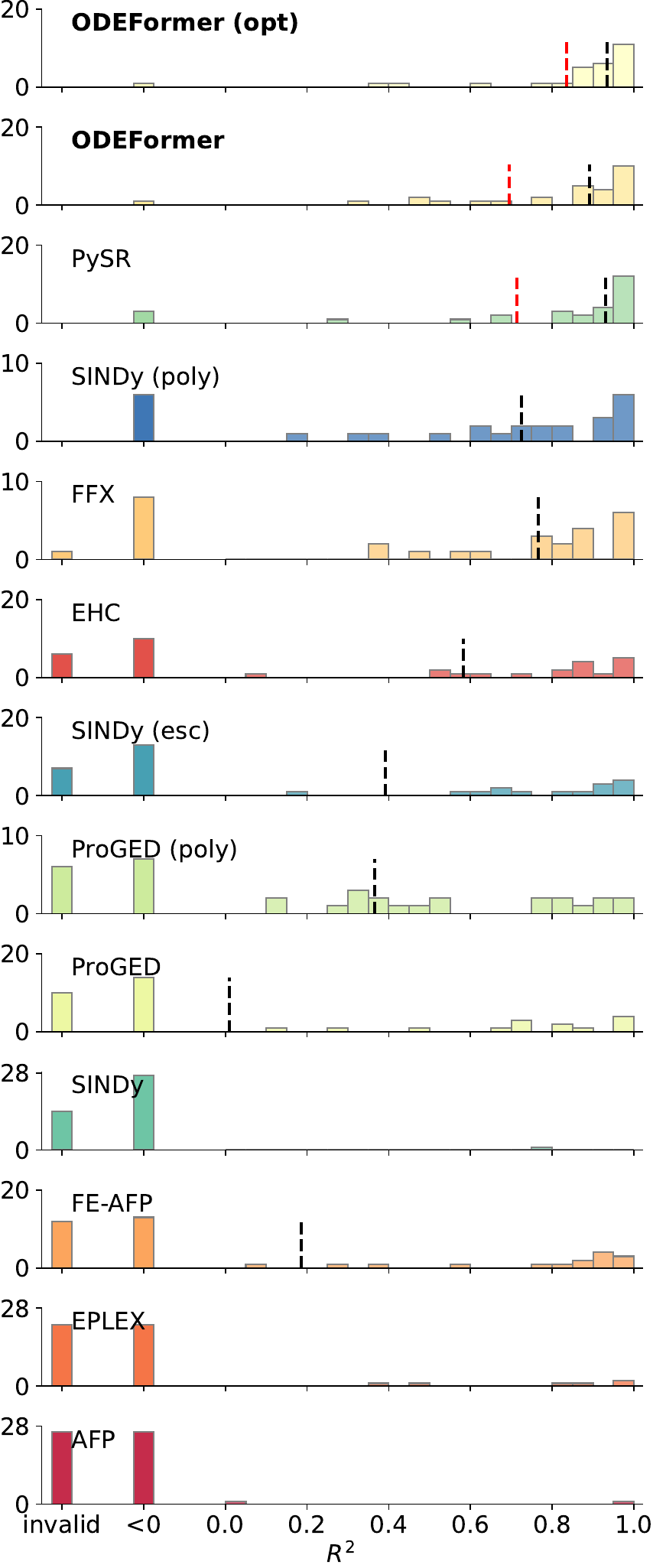}
    \caption[]{$\sigma$=0.05}
    \end{subfigure}
    \caption[]{Histogram of per equation $R^2$ scores for the \textbf{reconstruction} task on \textbf{Strogatz}. Subfigures correspond to different noise levels. The y-axis represents counts and is scaled per model for better visibility of the distribution of scores. The x-axis annotations ``invalid'' and ``<0'' respectively denote the number of invalid predictions as well as the number of predictions that yielded an $R^2$ score below 0. The red dashed line corresponds to the mean $R^2$ score across equations, the black dashed line corresponds to the median $R^2$ score.}
    \label{fig:histogram-strogatz-dense}
\end{figure}

\begin{figure}[htb]
    \begin{subfigure}[b]{.33\linewidth}
        \includegraphics[width=\linewidth]{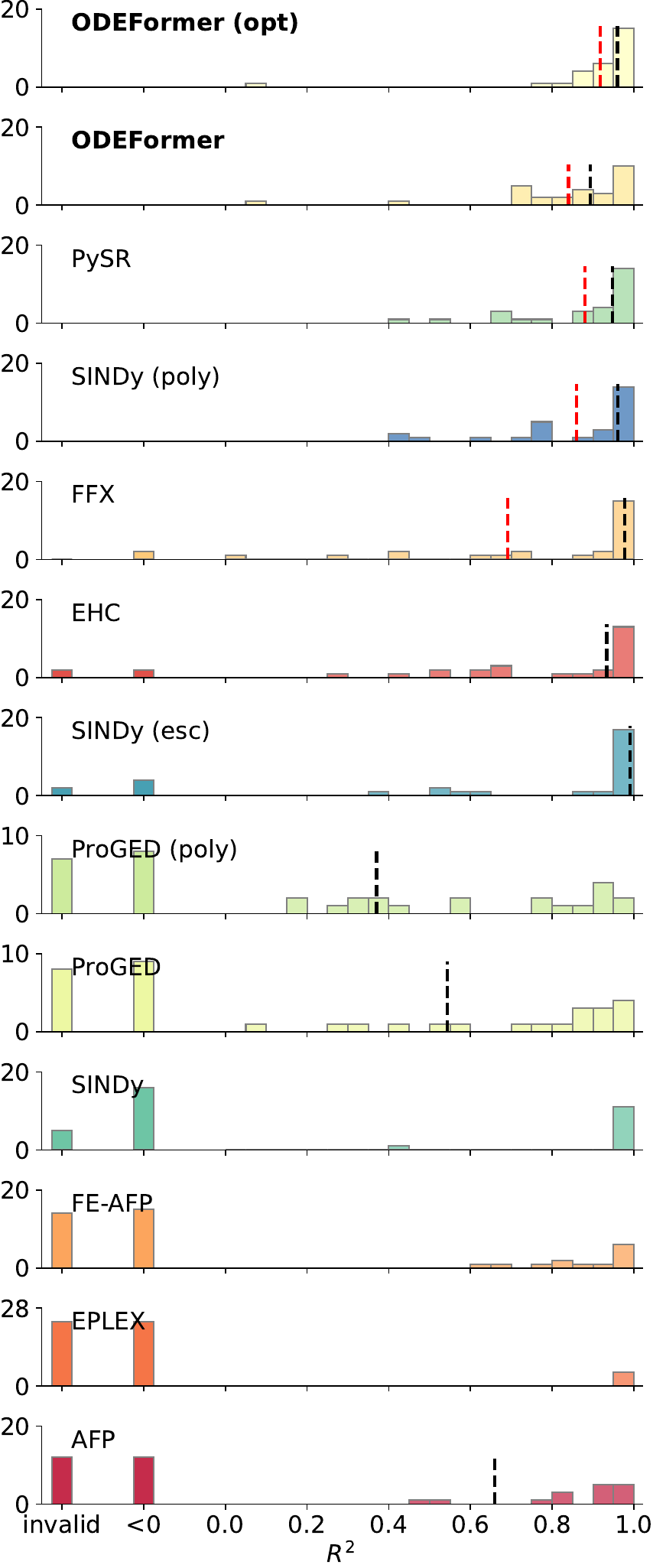}
    \caption[]{$\sigma$=0}
    \end{subfigure}
    \begin{subfigure}[b]{.33\linewidth}
        \includegraphics[width=\linewidth]{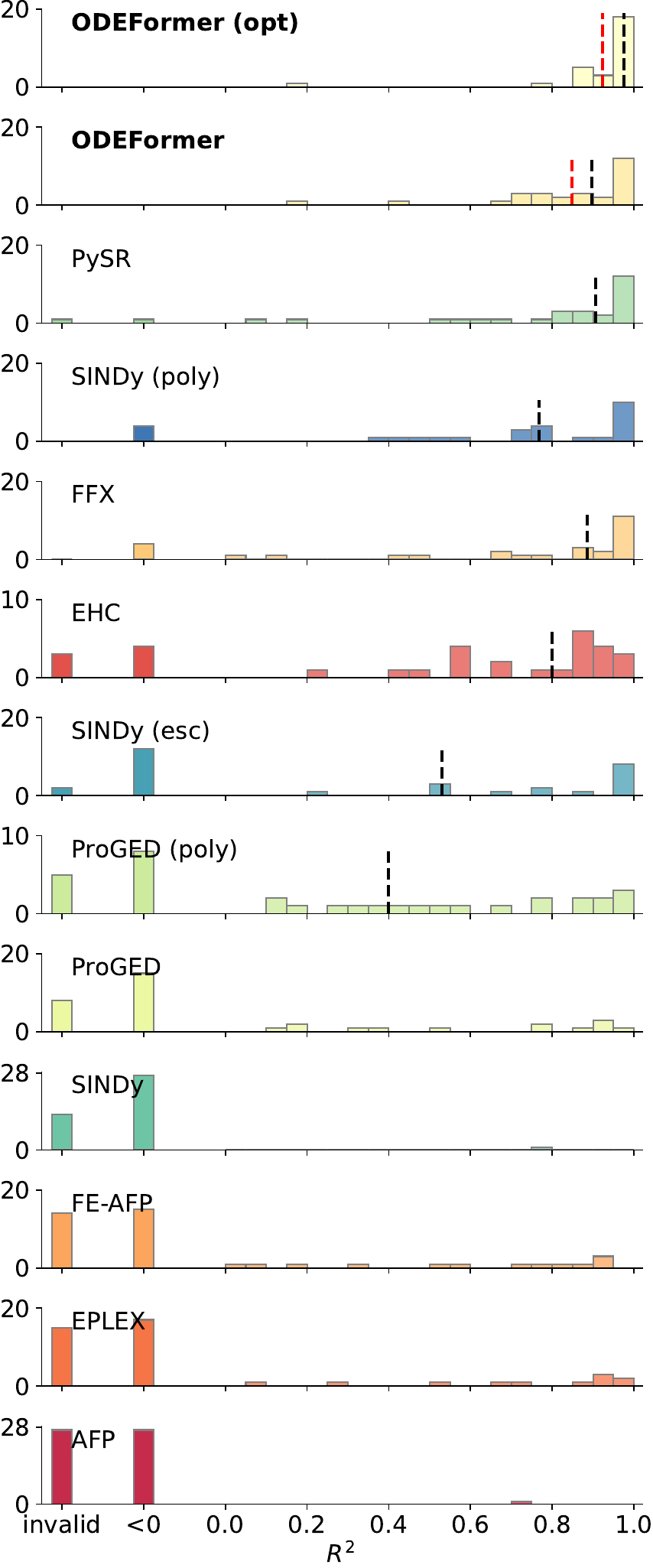}
    \caption[]{$\sigma$=0.01}
    \end{subfigure}
    \begin{subfigure}[b]{.33\linewidth}
        \includegraphics[width=\linewidth]{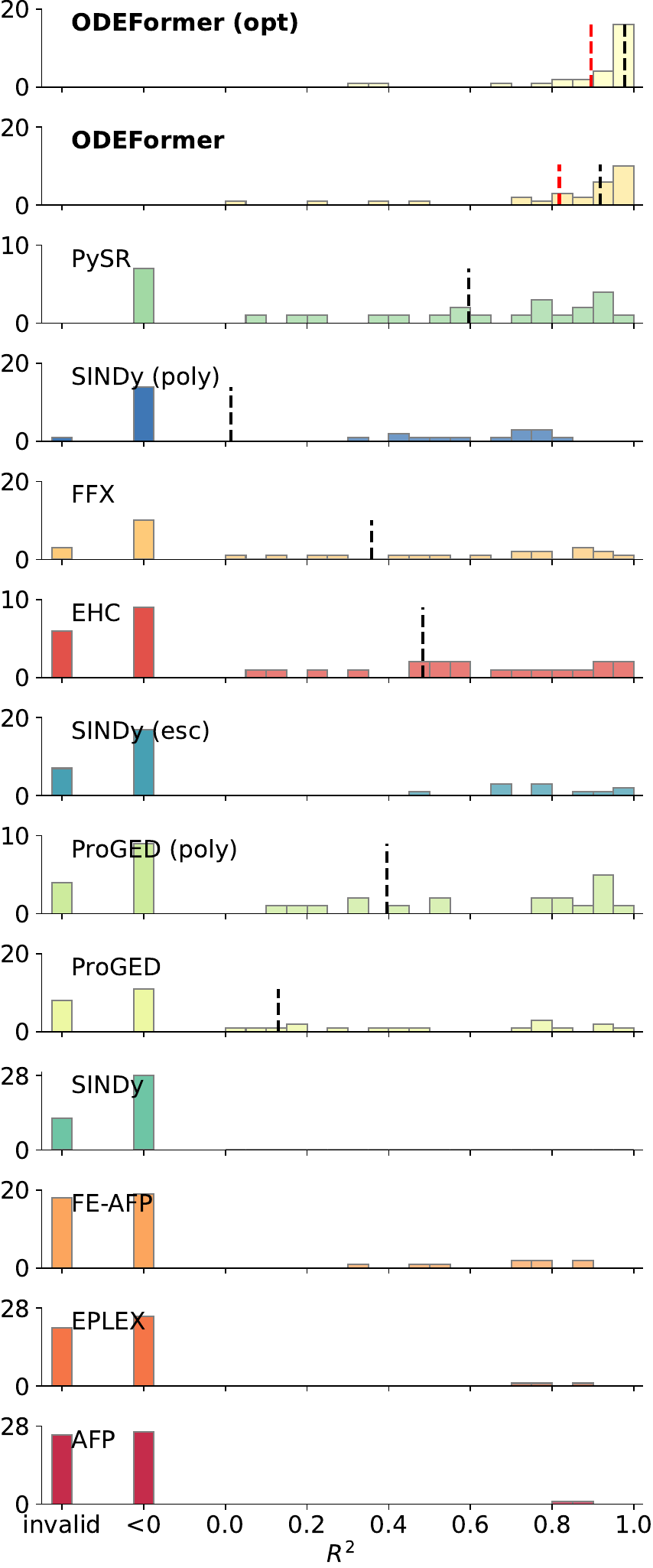}
    \caption[]{$\sigma$=0.05}
    \end{subfigure}
    \caption[]{Histogram of per equation $R^2$ scores for the \textbf{reconstruction} task on \textbf{Strogatz} where \textbf{50\%} of the trajectory are dropped uniformly at random ($\rho=0.5$). Subfigures correspond to different noise levels. The y-axis represents counts and is scaled per model for better visibility of the distribution of scores. The x-axis annotations ``invalid'' and ``<0'' respectively denote the number of invalid predictions as well as the number of predictions that yielded an $R^2$ score below 0. The red dashed line corresponds to the mean $R^2$ score across equations, the black dashed line corresponds to the median $R^2$ score.}
    \label{fig:histogram-strogatz-sparse}
\end{figure}

\begin{figure}[htb]
    \begin{subfigure}[b]{.33\linewidth}
        \includegraphics[width=\linewidth]{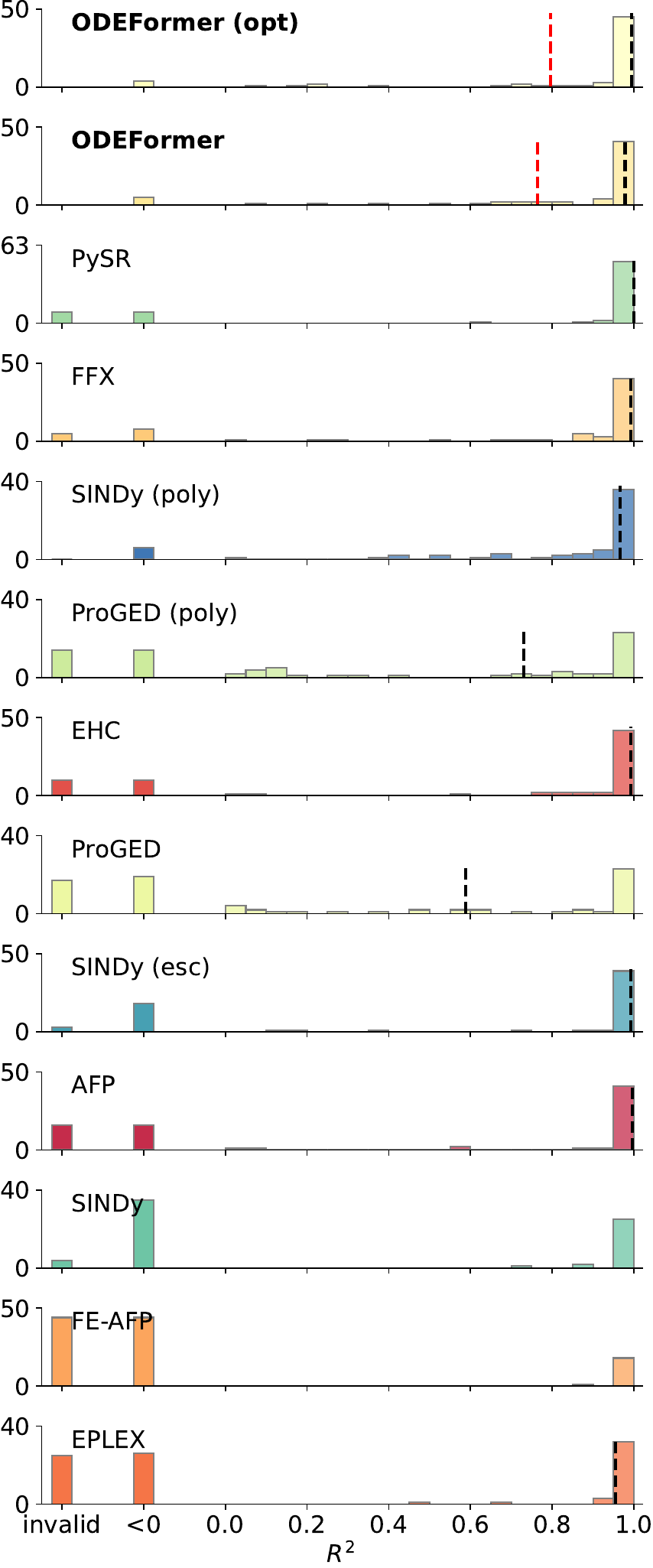}
    \caption[]{$\sigma$=0}
    \end{subfigure}
    \begin{subfigure}[b]{.33\linewidth}
        \includegraphics[width=\linewidth]{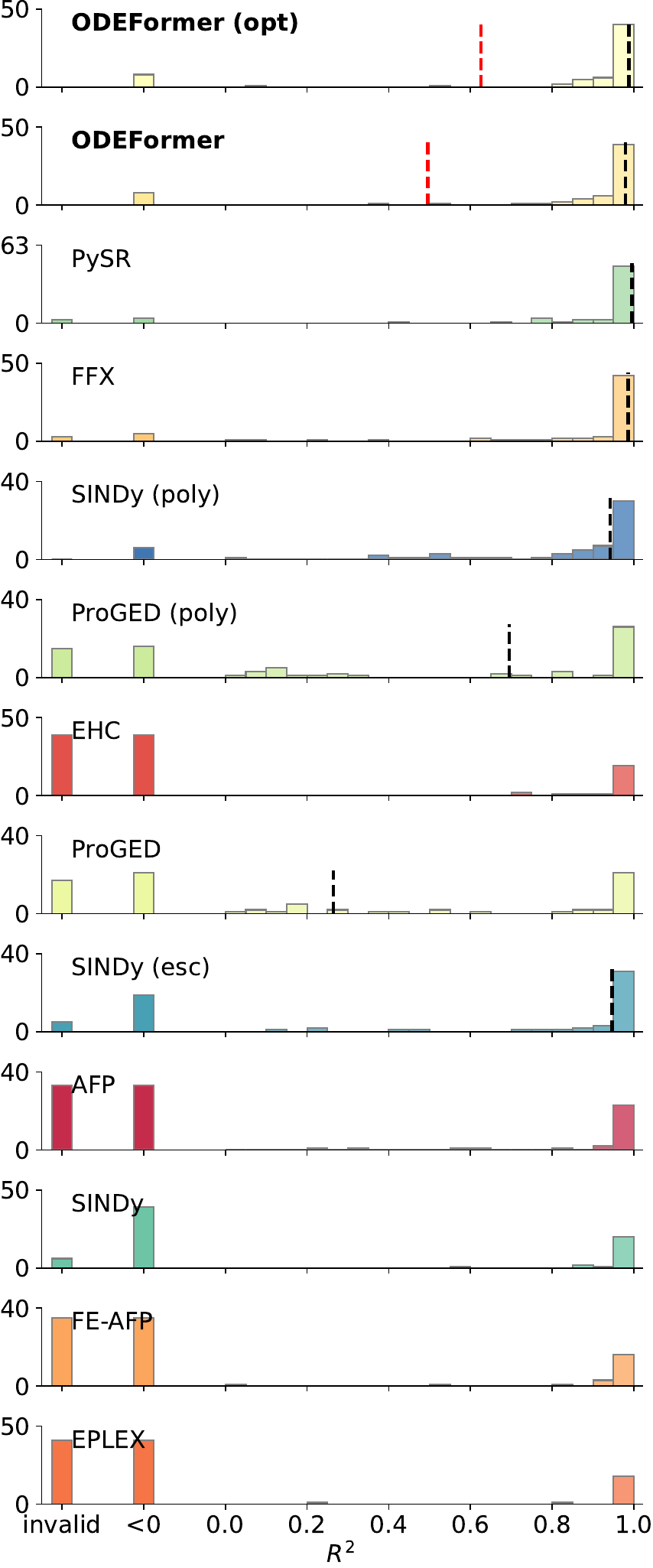}
    \caption[]{$\sigma$=0.01}
    \end{subfigure}
    \begin{subfigure}[b]{.33\linewidth}
        \includegraphics[width=\linewidth]{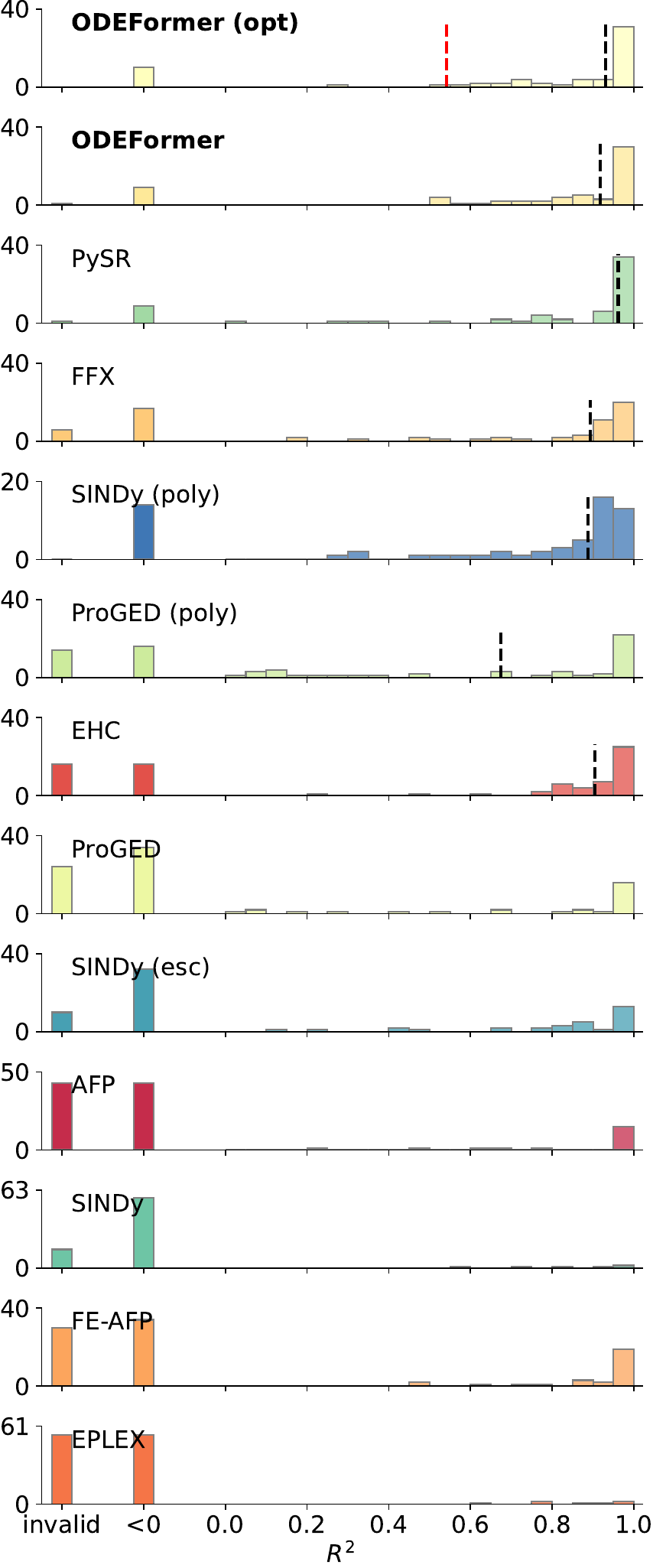}
    \caption[]{$\sigma$=0.05}
    \end{subfigure}
    \caption[]{Histogram of per equation $R^2$ scores for the \textbf{reconstruction} task on \textbf{ODEBench}. Subfigures correspond to different noise levels. The y-axis represents counts and is scaled per model for better visibility of the distribution of scores. The x-axis annotations ``invalid'' and ``<0'' respectively denote the number of invalid predictions as well as the number of predictions that yielded an $R^2$ score below 0. The red dashed line corresponds to the mean $R^2$ score across equations, the black dashed line corresponds to the median $R^2$ score.}
    \label{fig:histogram-odebench-dense}
\end{figure}

\begin{figure}[htb]
    \begin{subfigure}[b]{.33\linewidth}
        \includegraphics[width=\linewidth]{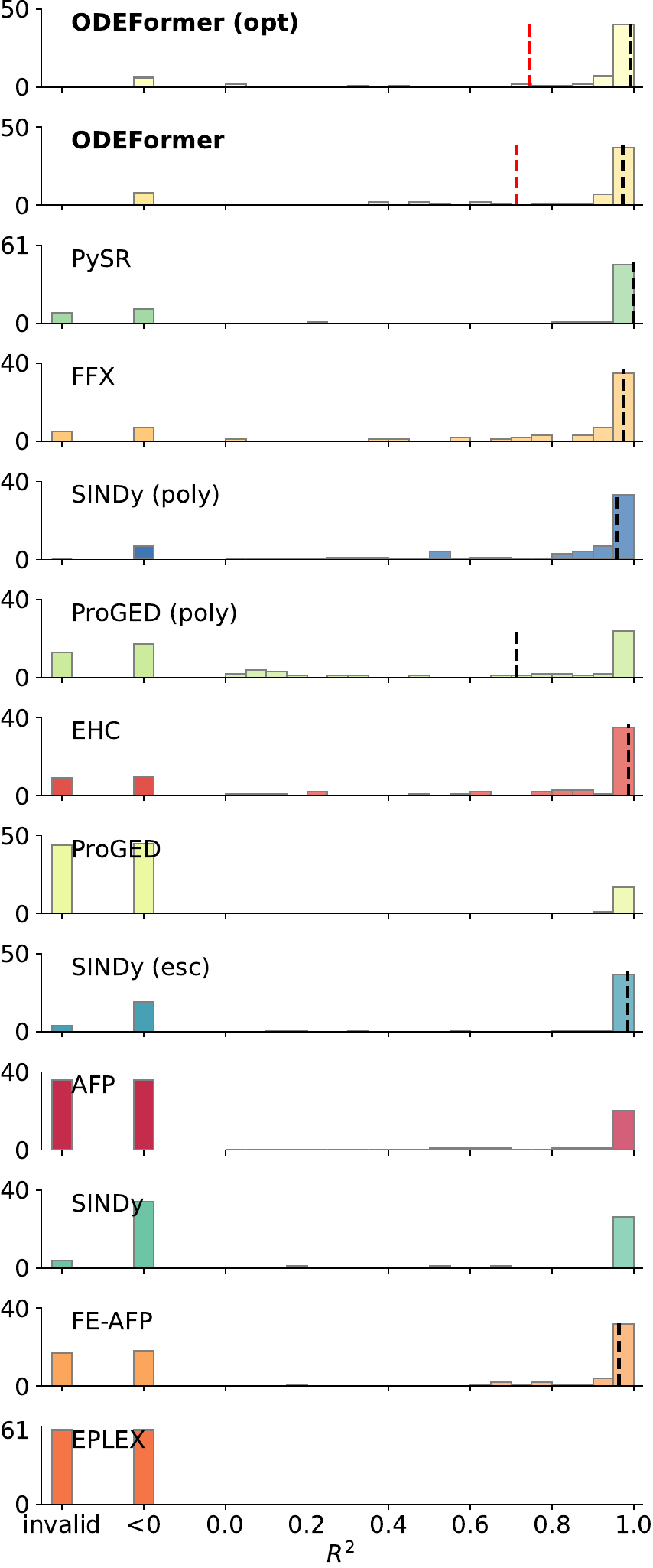}
    \caption[]{$\sigma$=0}
    \end{subfigure}
    \begin{subfigure}[b]{.33\linewidth}
        \includegraphics[width=\linewidth]{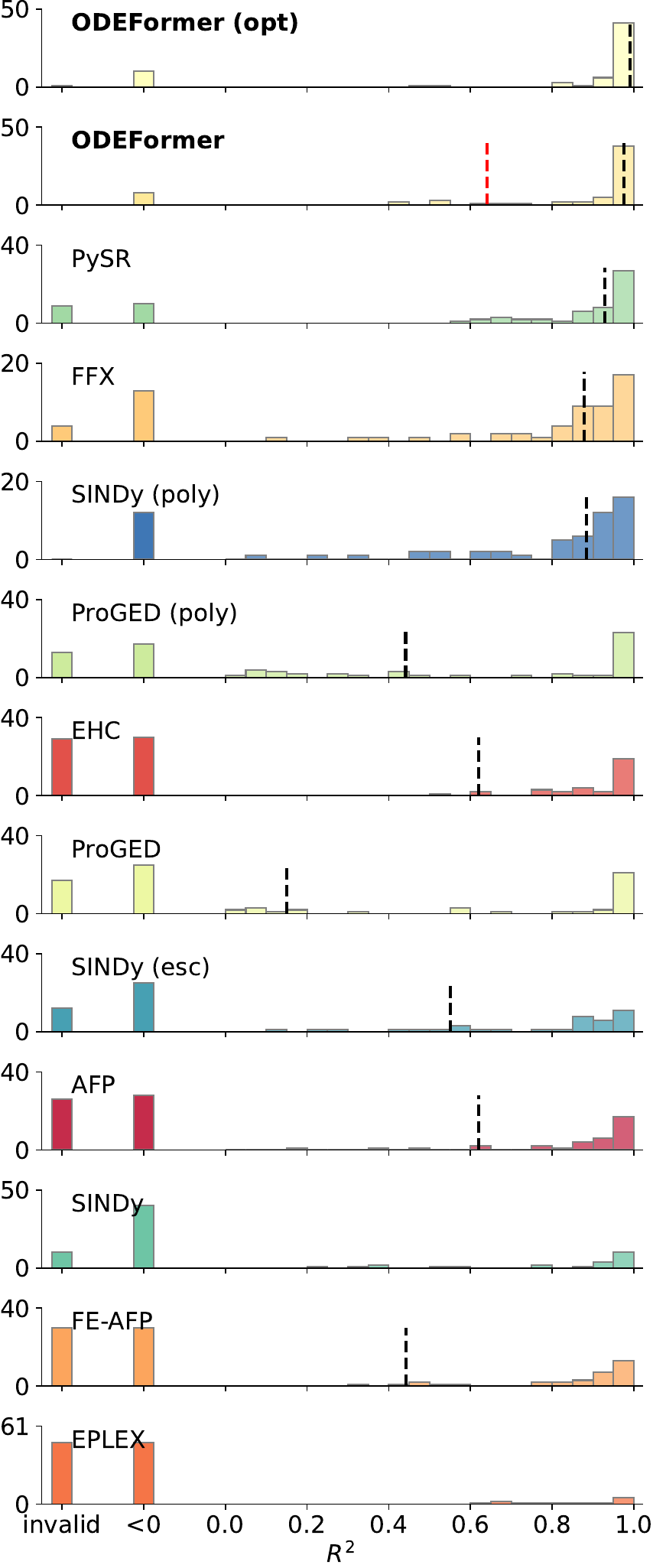}
    \caption[]{$\sigma$=0.01}
    \end{subfigure}
    \begin{subfigure}[b]{.33\linewidth}
        \includegraphics[width=\linewidth]{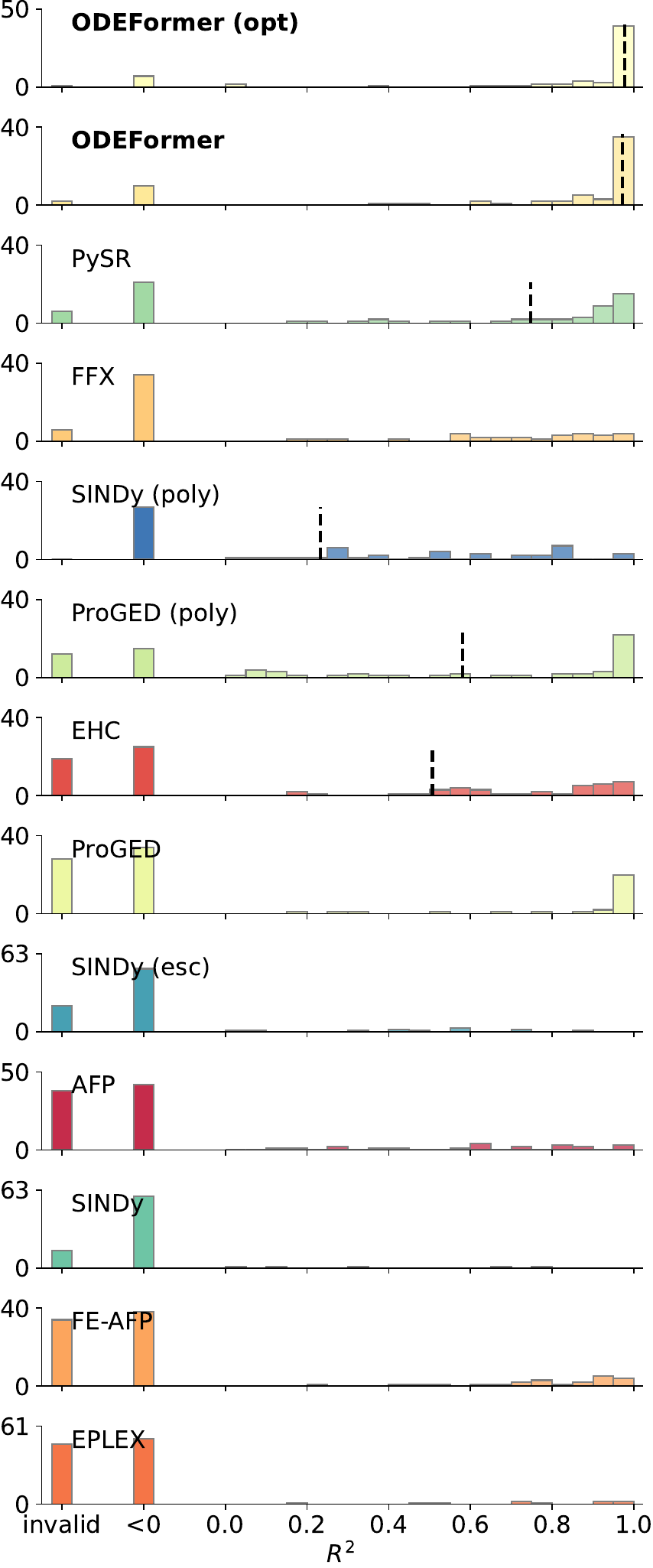}
    \caption[]{$\sigma$=0.05}
    \end{subfigure}
    \caption[]{Histogram of per equation $R^2$ scores for the \textbf{reconstruction} task on \textbf{ODEBench} where \textbf{50\%} of the trajectory are dropped uniformly at random ($\rho=0.5$). Subfigures correspond to different noise levels. The y-axis represents counts and is scaled per model for better visibility of the distribution of scores. The x-axis annotations ``invalid'' and ``<0'' respectively denote the number of invalid predictions as well as the number of predictions that yielded an $R^2$ score below 0. The red dashed line corresponds to the mean $R^2$ score across equations, the black dashed line corresponds to the median $R^2$ score.}
    \label{fig:histogram-odebench-sparse}
\end{figure}

\begin{figure}[htb]
    \begin{subfigure}[b]{.33\linewidth}
        \includegraphics[width=\linewidth]{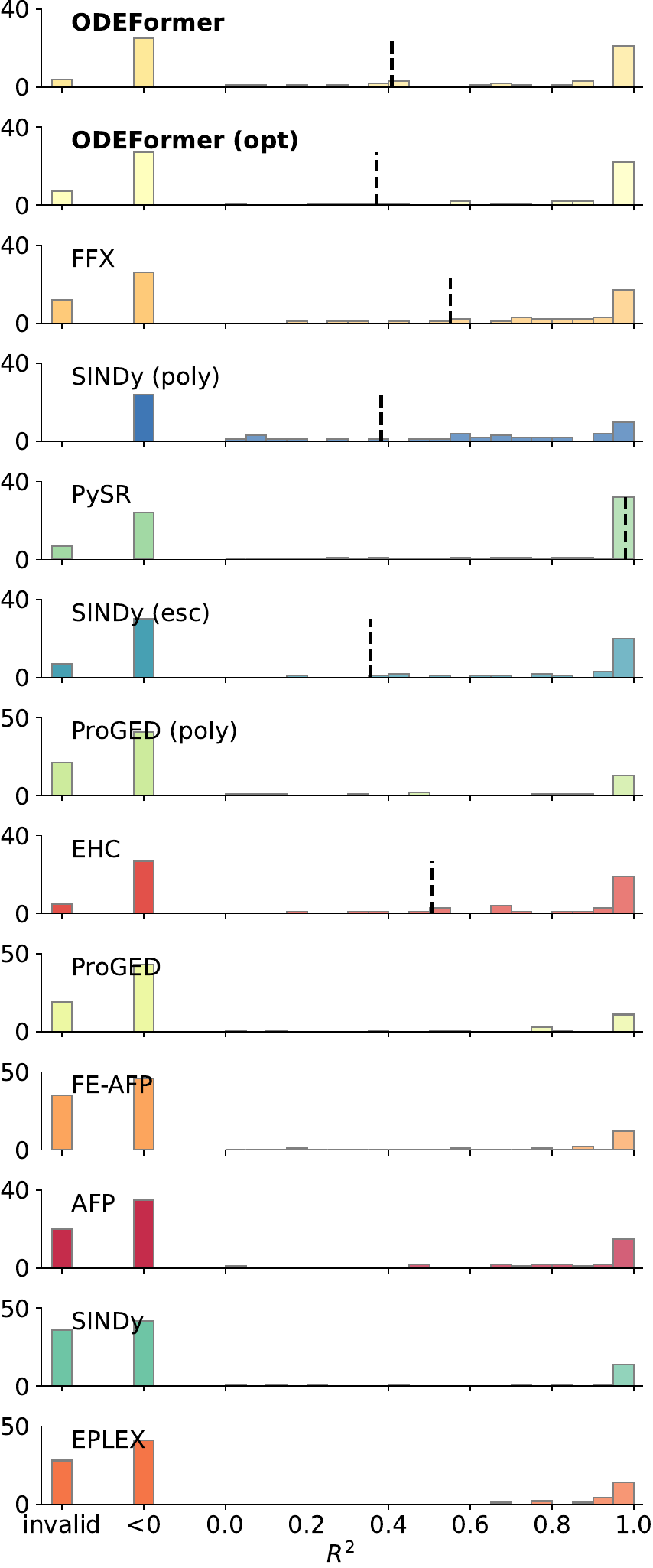}
    \caption[]{$\sigma$=0}
    \end{subfigure}
    \begin{subfigure}[b]{.33\linewidth}
        \includegraphics[width=\linewidth]{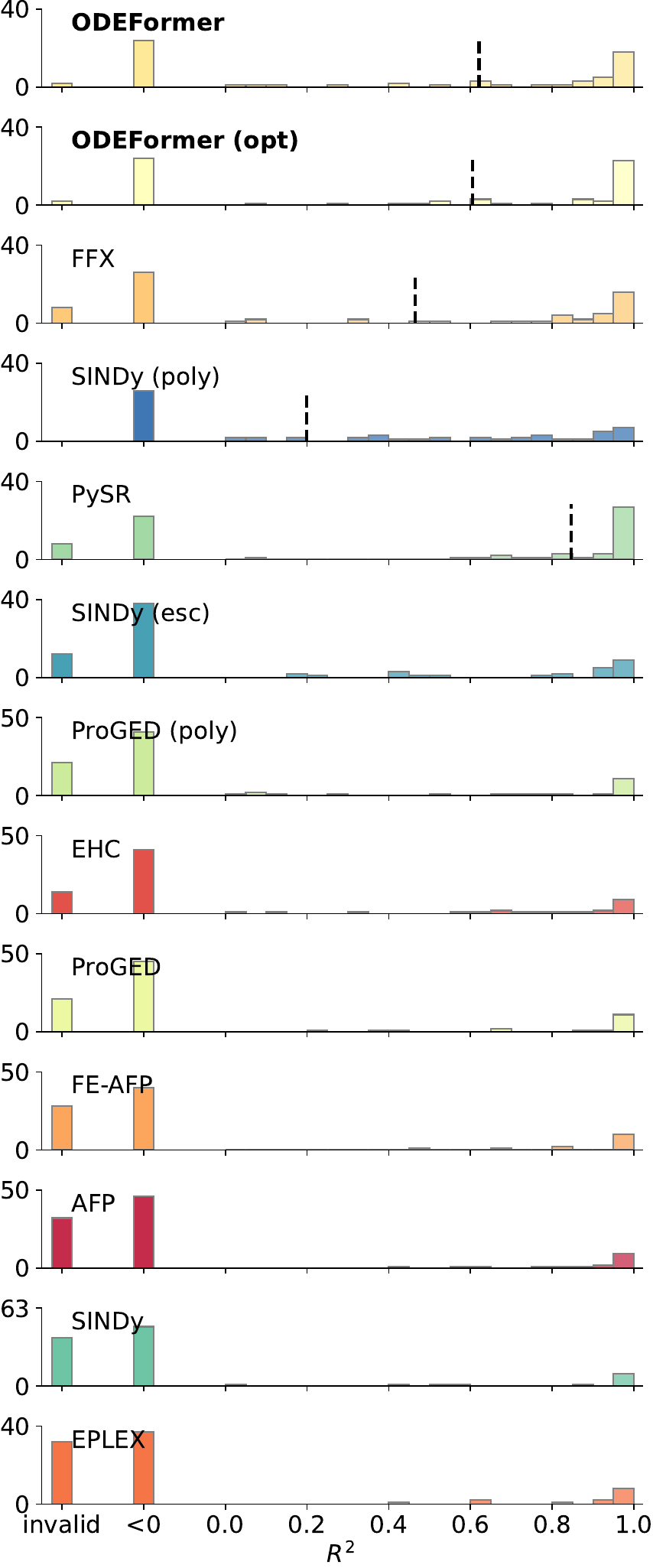}
    \caption[]{$\sigma$=0.01}
    \end{subfigure}
    \begin{subfigure}[b]{.33\linewidth}
        \includegraphics[width=\linewidth]{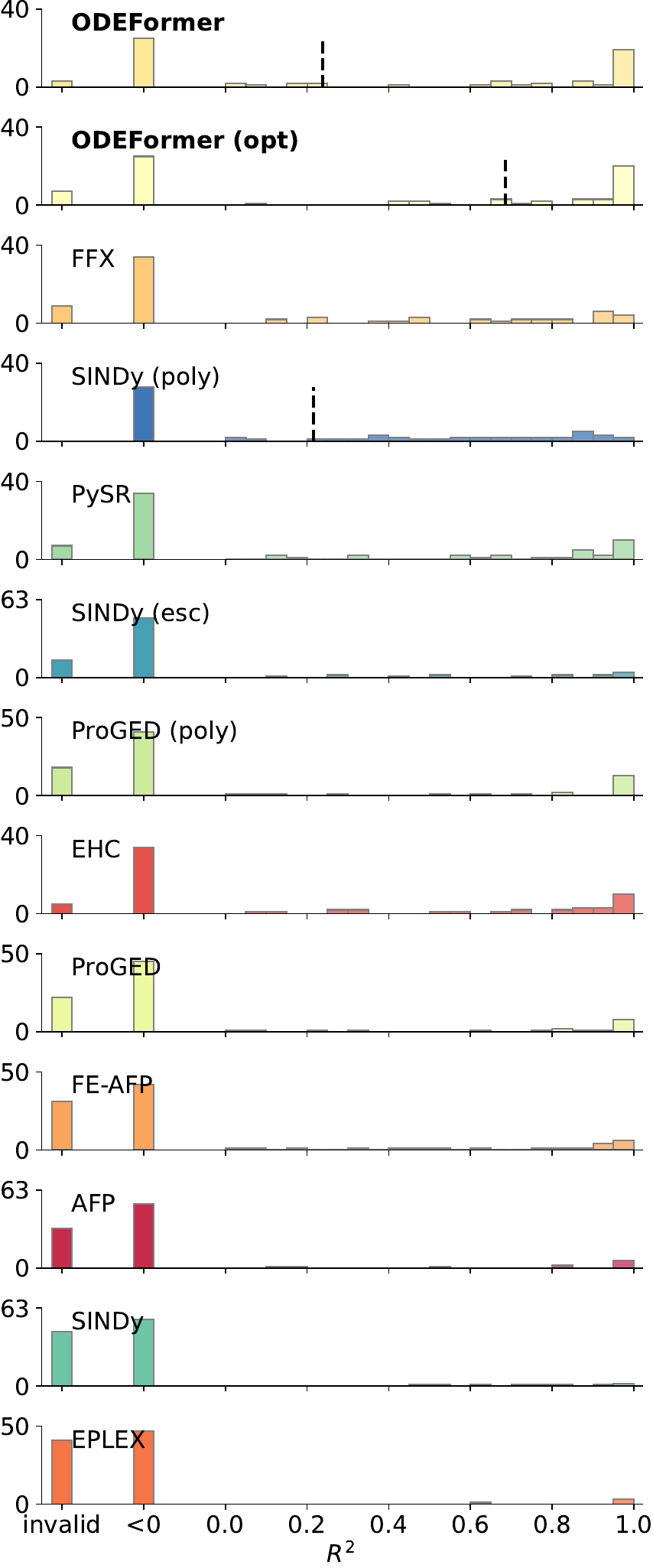}
    \caption[]{$\sigma$=0.05}
    \end{subfigure}
    \caption[]{Histogram of per equation $R^2$ scores for the \textbf{generalization} task on \textbf{ODEBench}. Subfigures correspond to different noise levels. The y-axis represents counts and is scaled per model for better visibility of the distribution of scores. The x-axis annotations ``invalid'' and ``<0'' respectively denote the number of invalid predictions as well as the number of predictions that yielded an $R^2$ score below 0. The red dashed line corresponds to the mean $R^2$ score across equations, the black dashed line corresponds to the median $R^2$ score.}
    \label{fig:histogram-odebench-dense_generalization}
\end{figure}

\begin{figure}[htb]
    \begin{subfigure}[b]{.33\linewidth}
        \includegraphics[width=\linewidth]{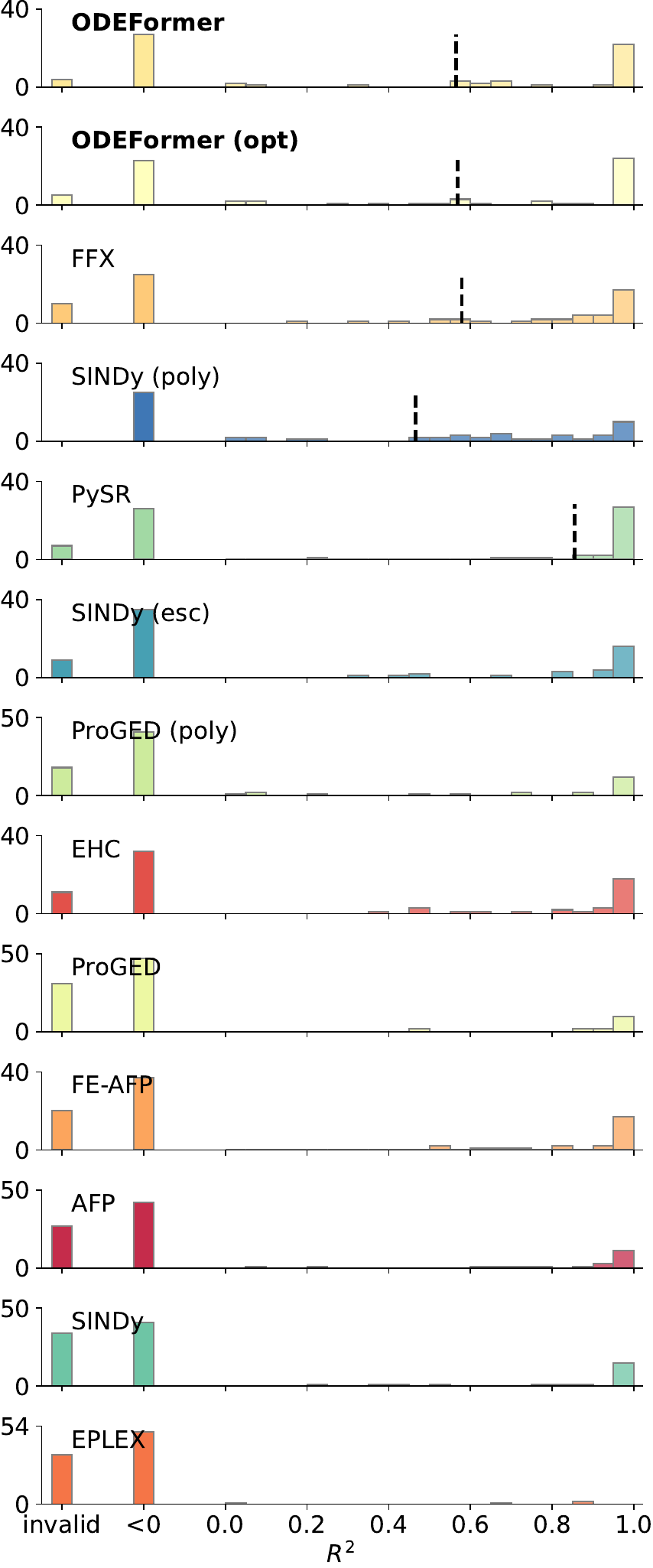}
    \caption[]{$\sigma$=0}
    \end{subfigure}
    \begin{subfigure}[b]{.33\linewidth}
        \includegraphics[width=\linewidth]{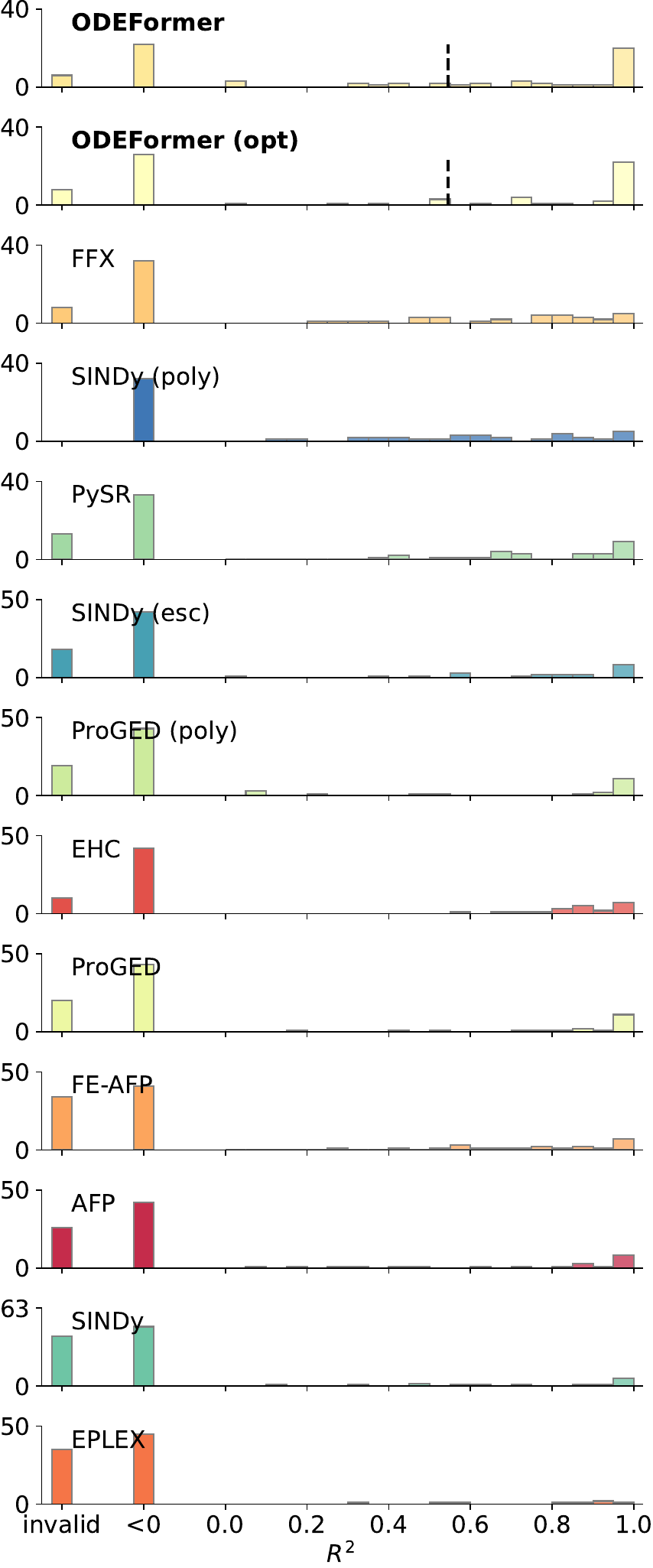}
    \caption[]{$\sigma$=0.01}
    \end{subfigure}
    \begin{subfigure}[b]{.33\linewidth}
        \includegraphics[width=\linewidth]{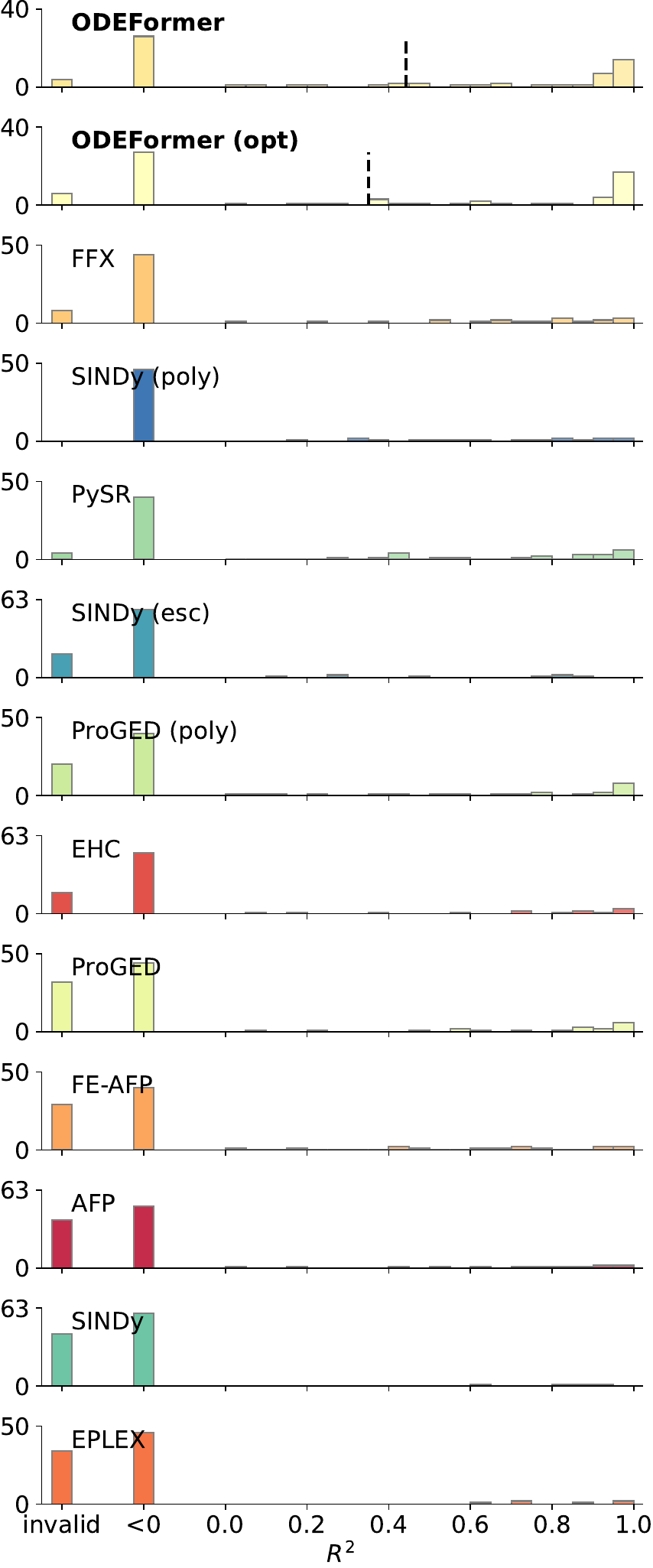}
    \caption[]{$\sigma$=0.05}
    \end{subfigure}
    \caption[]{Histogram of per equation $R^2$ scores for the \textbf{generalization} task on \textbf{ODEBench} where \textbf{50\%} of the trajectory are dropped uniformly at random ($\rho=0.5$). Subfigures correspond to different noise levels. The y-axis represents counts and is scaled per model for better visibility of the distribution of scores. The x-axis annotations ``invalid'' and ``<0'' respectively denote the number of invalid predictions as well as the number of predictions that yielded an $R^2$ score below 0. The red dashed line corresponds to the mean $R^2$ score across equations, the black dashed line corresponds to the median $R^2$ score.}
    \label{fig:histogram-odebench-sparse_generalization}
\end{figure}

In \cref{fig:histogram-strogatz-dense,fig:histogram-strogatz-sparse,fig:histogram-odebench-dense,fig:histogram-odebench-sparse,fig:histogram-odebench-dense_generalization,fig:histogram-odebench-sparse_generalization}, we plot histograms to better visualize how each model performs across different noise conditions, datasets, and evaluation tasks.


\end{document}